\documentclass[english]{article}

\usepackage{natbib}
 \bibpunct[, ]{(}{)}{,}{a}{}{,}%

\usepackage{algorithmicx,algorithm}
\usepackage[noend]{algpseudocode}
\usepackage{graphicx}
\usepackage{amsmath}
\usepackage{multirow}
\usepackage{subcaption}

\usepackage{booktabs}

\usepackage{amsmath}
\usepackage{amssymb}
\usepackage{setspace}
\usepackage[T1]{fontenc}
\usepackage[latin9]{inputenc}
\usepackage{geometry}
\usepackage{babel}
\usepackage{mathrsfs}
\usepackage{mathtools}
\usepackage{amsmath}
\usepackage{amsthm}
\usepackage{amssymb}
\usepackage{authblk}
\usepackage{breakurl}
\usepackage{algorithm}
\usepackage{algpseudocode}
\usepackage{tikz}
\usepackage{enumerate} 
\usepackage{graphicx}  
\usepackage{wrapfig} 
\usepackage{caption} 
\usepackage{hyperref}

\usepackage{natbib}
\bibpunct[, ]{(}{)}{,}{a}{}{,}%

\makeatletter
\theoremstyle{plain}
\newtheorem{thm}{\protect\theoremname}[section]
\theoremstyle{definition}

\theoremstyle{plain}
\newtheorem{assumption}[thm]{\protect\assumptionname}
\theoremstyle{plain}

\theoremstyle{remark}

\theoremstyle{plain}
\newtheorem{cor}[thm]{\protect\corollaryname}
\theoremstyle{plain}
\newtheorem{lem}[thm]{\protect\lemmaname}

\makeatother

\providecommand{\assumptionname}{Assumption}
\providecommand{\corollaryname}{Corollary}
\providecommand{\definitionname}{Definition}
\providecommand{\lemmaname}{Lemma}
\providecommand{\propositionname}{Proposition}
\providecommand{\remarkname}{Remark}
\providecommand{\theoremname}{Theorem}

\begin{document}





\title{Principled Design of Expected Improvement for Continuous Simulation Optimization}


\author[1,3]{Songhao Wang}
\author[1]{Jianglin Xia}
\author[2]{Haowei Wang}
\author[2]{Szu Hui Ng}
\affil[1]{College of Business, Southern University of Science and Technology, Shenzhen, CHINA}
\affil[2]{Department of Industrial Systems Engineering and Management, National University of Singapore, Singapore, SINGAPORE}
\affil[3]{Corresponding author: wangsh2021@sustech.edu.cn}
\date{}

\maketitle

\begin{abstract}
Expected improvement (EI) based on stochastic kriging (SK) surrogate is widely used in continuous simulation optimization, but \emph{unknown heteroscedastic} noise on simulation output makes three basic design choices consequential: the incumbent used to define improvement, the replication level at each sampled point, and the terminal recommendation. Existing methods typically either assume the noise levels are known or choose these components separately through heuristic rules or empirical calibration. We instead derive a high-probability upper bound on simple regret and systematically choose the three components within general rule classes to tighten this bound. We first establish a finite-sample confidence bound for the SK predictor with \emph{unknown and estimated} noise levels as a foundation for the three design choices. We then analyze general posterior-quantile families for both the incumbent and the terminal recommendation. For the incumbent, minimizing the regret bound selects a lower quantile coefficient that matches the coefficient in the SK predictor confidence bound. For the terminal recommendation, minimizing the same bound instead selects the sampled point with the best upper posterior quantile under the same coefficient. These
opposite quantile choices reflect the different roles of search and
implementation in simulation optimization. Intermediate simulations are
exploratory with an optimistic incumbent, while the terminal recommendation is implemented in the real system and is therefore relatively conservative about the remaining posterior uncertainty. 
Under mild conditions satisfied by common kernels, minimizing a tractable kernel-specific bound yields $\Theta(\log T)$ replications. For the squared-exponential kernel, the resulting simple regret is $\widetilde O(T^{-1/2})$, matching the polynomial dependence on the raw simulation budget in the minimax lower bound. For a Mat\'ern kernel, 
the derived rate also matches the corresponding confidence-bound rate under known homoscedastic noise up to logarithmic factors. To our knowledge, this is the first finite-budget
high-probability simple-regret guarantee of this form for a continuous-domain
EI-type simulation optimization method that learns unknown heteroscedastic
variances through replications.
\end{abstract} 





%

\section{Introduction}\label{sec:Intro}

Simulation optimization studies decision problems in which candidate decision points are evaluated by running simulation models. It is widely used when the system of interest is analytically intractable, or too complex to optimize directly. Representative applications include manufacturing and production scheduling, healthcare resource planning, transportation, and supply chain management (see, e.g., \citealt{amaran2016simulation,wang2023simulation,osorio2013simulation}). A typical simulation optimization procedure begins with a decision-maker providing a certain amount of computational resources. An algorithm then allocates these resources among selected decision points to run experiments. Each run produces a random output. Once the budget is exhausted, a final recommended solution is required to be implemented in the real-world system.

Surrogate-based algorithms are a common and efficient strategy for simulation optimization. They fit a surrogate model to the simulation results collected so far and use it to direct subsequent simulations toward inputs that appear promising or remain uncertain. Gaussian process (GP), or kriging, models are especially useful. They provide both a predictor and a measure of predictive uncertainty \citep{williams2006gaussian}. Stochastic kriging (SK) extends this model by separating uncertainty about the response surface from the intrinsic noise in
stochastic simulation outputs \citep{ankenman2010stochastic}. GP-based search has consequently become an important methodology in the simulation literature, including discrete optimization via simulation, Bayesian optimization with correlated sampling, and continuous optimization via simulation (see, e.g., \citealt{sun2014balancing, xie2016bayesian, pearce2022bayesian, wang2020parallel}, \citealt{wang2025gaussian}). Within this broader class, expected improvement (EI) is one of the most widely used sequential selection rules. Introduced in efficient global optimization (EGO) for deterministic black-box functions, EI selects the next design by its expected improvement over a current best value (also called incumbent value). It combines predicted performance and predictive uncertainty in a simple closed-form criterion \citep{jones1998efficient}. Several approaches adapt this idea to stochastic simulation, such as sequential kriging optimization (SKO, \citealp{huang2006global}), expected quantile improvement (EQI, \citealp{picheny2013quantile}), and two-stage stochastic optimization (TSSO, \citealp{quan2013simulation}).

EI approaches have also been used in many areas outside simulation optimization, including engineering design and computer science, typically presented in bandit-style problems \citep{shahriari2015taking}. However, simulation optimization differs from these settings in two important respects. The first difference is the noise structure. Many theoretical analyses in GP bandits assume a known, homoscedastic, or otherwise simple noise model (see, e.g., \citealt{srinivas2010gaussian}). In stochastic simulation, the output variance is often input-dependent and unknown. Queueing simulations provide a classical example. Response variability can change substantially with traffic intensity, and nearly saturated queues are known to produce highly heteroscedastic simulation responses \citep{cheng1999improved}. Similar noise structures arise in manufacturing, epidemic management, and other operational systems \citep{binois2018practical}. In this setting, a single run cannot determine whether an observation reflects the underlying expected response or merely local simulation variability. Replication at the same input is therefore essential for estimating local variance and informing the surrogate model about the reliability of a sample mean. The second difference lies in the performance criterion. Much of the GP-bandit literature studies cumulative regret because each sampled input is treated as a realized decision. Simulation optimization instead treats the sampled inputs as exploratory experiments. As only the design implemented at the end matters, the relevant criterion is simple regret, the optimality gap of the final recommendation. It must therefore certify the quality of that recommendation, not the sequence of sampled points.

These simulation-specific features turn three seemingly routine implementation choices into central design decisions: the incumbent value, the replication allocation, and the terminal recommendation. It is worth emphasizing that these difficulties do not arise in the same form when evaluations are noise-free. Some may also disappear under simpler or fully specified noise structures. In deterministic EI, for example, the incumbent can be taken directly as the best objective value observed so far \citep{jones1998efficient,bull2011convergence}. Replication is unnecessary because repeated evaluations at the same input return the same value, and the terminal recommendation can simply be the best input evaluated during the search. Once simulation outputs exhibit a complex noise structure, however, every historical response is contaminated by potentially heteroscedastic noise. 
The three choices that are natural in deterministic EI therefore become nontrivial and potentially consequential.

Consider first how noise affects the choice of the current best. Using either a raw observation or a sample mean from several replications may select an incumbent point that appears exceptionally good only because of favorable noise, even though its true mean performance is much worse. Such a downward distortion can make genuine improvements appear unlikely and misdirect the subsequent search. Existing studies therefore use more stable information supplied by the GP model, such as the best posterior prediction or a posterior quantile. 
A broad family of such choices can be written as $\xi_n^{(c_n)}
    :=
    \min
    \bigl\{\mu_n(x)-c_n\sigma_n(x)\bigr\}$,
where $x$ represents the decision point, $\mu_n(x)$ and $\sigma_n(x)$ are the SK prediction and predictive standard deviation, respectively, and $c_n\in\mathbb R$ is the quantile coefficient after selecting $n$ decision points. 
The choice $c_n=0$ gives the best-posterior-prediction incumbent used in noisy-EI analyses \citep{wang2014theoretical,gupta2022regret}. SKO sets $c_n=-1$, whereas EQI uses a prespecified $c_n\leq 0$; both choices correspond to upper quantiles \citep{huang2006global,picheny2013quantile,jalali2017comparison}.
Although these choices have intuitive justifications, their coefficients are typically prescribed in advance or treated as tuning parameters based on convention or empirical performance.

Similar discretion appears in the other two design decisions. How to allocate a limited simulation budget across candidate designs is itself a central problem in simulation optimization \citep{chen2000simulation,ankenman2010stochastic,binois2019replication}. In stochastic EI, the number of replications assigned to a design determines the accuracy of its sample mean and local noise-variance estimate. Under a fixed raw simulation budget, it also determines how many distinct designs the algorithm can explore. 
Existing EI procedures use several different mechanisms, including equal allocation, repeated sampling of past points, separate allocation rules such as optimal computing budget allocation (OCBA), or user-specified stopping thresholds \citep{huang2006global,picheny2013quantile,quan2013simulation,pedrielli2020extended,jalali2017comparison}. 
As with the incumbent choice discussed above, quantities such as equal allocation replication sizes, allocation-stage budgets, and stopping thresholds often remain external inputs whose values must be fixed in advance or calibrated empirically. Terminal recommendations are similarly heterogeneous: different methods return the sampled design with the best sample mean, the best GP prediction, or the best posterior quantile \citep{picheny2013quantile,quan2013simulation,jalali2017comparison}.

Taken together, these alternatives allow many combinations of design choices. This flexibility, however, can leave practitioners uncertain about which combination to use or what guarantee it provides. A more systematic and coherent approach is therefore needed to select these components while reducing reliance on heuristic or empirically calibrated settings. Although empirical studies have compared different algorithms numerically \citep{jalali2017comparison}, they do not provide a unified theoretical basis for selecting among them. We address this need from a theoretical perspective. Specifically, we begin with a generic EI procedure in which all three components are left general within broad rule classes. We then derive a high-probability upper bound on the simple regret of this generic procedure. The three design choices are systematically selected to tighten this bound. 
To our knowledge, no existing continuous-domain EI method provides a comparable high-probability simple-regret bound under heteroscedastic noise with unknown input-dependent variances estimated from replications.

In our work, the simple regret bound for the algorithm is built upon a finite-sample confidence bound for the SK predictor. At each selected design, independent replications provide a sample mean and a sample-variance estimate of the unknown noise variance. As a sample variance can underestimate the true variance, directly treating it as known can make the confidence band too narrow. We derive a correction factor for this underestimation and obtain a confidence event that holds jointly over the entire design space and all simulation iterations. The resulting confidence coefficient, denoted by $\Lambda_n$ after $n$ sampled designs, becomes the common calibration scale for the search and recommendation rules.
To our knowledge, this is the first finite-budget SK confidence bound for adaptive designs with unknown heteroscedastic noise estimated from replications. Unlike \citet{chen2024uniform}, which treats estimated variances asymptotically, our bound provides finite-sample coverage and directly supports the EI regret analysis.

Building on this confidence guarantee for the SK predictor, we derive a simple-regret upper bound for a generic EI procedure. The bound is generic because none of the three design components has yet been fixed. It accommodates the global linear-quantile incumbent family described above, an analogous family of terminal rules that use the same linear posterior-quantile form to select among sampled points, and any admissible replication size. We then select the three design components to tighten this certificate. Within the general global linear-quantile incumbent family, minimizing the incumbent-dependent part of the certificate selects $c_n=\Lambda_n$. The resulting lower-quantile incumbent keeps a region in contention whenever the SK confidence bound cannot yet rule out that it contains an optimizer. Within the general family of terminal posterior-quantile rules, the terminal-dependent part is minimized by an upper quantile with coefficient equal to the final confidence width of the SK predictor. This choice therefore accounts conservatively for its remaining uncertainty. Unlike the first two design choices, the replication size depends on the specific covariance kernel used by the GP. After determining the first two parameters, we therefore introduce mild conditions on the kernel that are satisfied by standard kernels, including squared-exponential (SE), Matérn, and polynomial kernels. Under these conditions, we show that the optimal replication size has order $\Theta(\log T)$ given a total budget of $T$ simulation replications. Thus, the simple-regret analysis yields an optimistic incumbent for search, logarithmic replication for learning unknown heteroscedastic variances while preserving spatial exploration, and a conservative sampled-point recommendation for implementation.

Under these calibrated design choices, we obtain the optimized high-probability simple-regret guarantee. For SE kernels, the resulting simple regret is of order $\tilde{O}(T^{-1/2})$, matching the polynomial dependence in the minimax lower bound for noisy GP optimization. For Mat\'ern kernels, the rate is $\tilde{O}(T^{-\frac{2\nu-d}{2(2\nu+d)}})$, where $\tilde{O}$ suppresses logarithmic factors, $\nu$ is the smoothness parameter of the Mat\'ern kernel, and $d$ is the dimension of the design space. It also matches, up to logarithmic factors, existing confidence-based regret bounds under the simpler setting of known homoscedastic noise, demonstrating the strength of our guarantee \citep{gupta2022regret,vakili2021information}. We highlight that our bound is a stronger and more decision-relevant form of guarantee for simulation optimization than Bayesian simple regret \citep{takeno2025regret,wang2025gaussian}, which averages over random functions, noises, and algorithmic randomness. In addition, we find that estimating the unknown heteroscedastic variances through logarithmic replication introduces at most an additional logarithmic factor into the regret bound. It therefore leaves the leading polynomial dependence on $T$ unchanged.

The contributions of this paper are fourfold. First, we derive a finite-sample confidence bound for the SK predictor. This bound is not specific to EI. It can also support other SK-based sequential-design methods. Second, we analyze a general family of global linear-quantile incumbents. This family includes the best posterior prediction and common posterior-quantile choices. Our analysis identifies the quantile level that removes a potentially nonvanishing term in the simple-regret certificate. Thus, the quantile level is determined by theory rather than treated as a tuning parameter. Third, we study replication as a budget-allocation decision. Under mild kernel conditions, we show that $b_T=\Theta(\log T)$ is order-optimal for the resulting regret bound. Fourth, We analyze a general family of terminal posterior quantiles over actually sampled designs. The confidence-matched upper quantile is a bound-minimizing choice for the displayed uniform certificate and yields a finite-budget high-probability simple-regret guarantee. To our knowledge, this is the first such bound for an EI-type simulation optimization method with unknown heteroscedastic noise. 

The remainder of the paper is organized as follows. Sections~\ref{sec:relatedwork} and~\ref{sec:preliminaries} review the related literature and introduce the necessary background on SK and EI. Section~\ref{sec:algorithm} presents the generic EI procedure with general design choices and establishes the confidence bound for the SK predictor. Section~\ref{sec:simple_regret} derives the simple-regret certificate for the generic EI procedure. Section~\ref{sec:designchoices} uses this certificate to select the three design components and establish the resulting convergence rates. Section~\ref{sec:numerical} presents the numerical experiments, and Section~\ref{sec:summary} concludes the paper. The proofs of all the theoretical results are presented in the Appendix.

\section{Related Works} \label{sec:relatedwork}

\citet{jones1998efficient} introduced the EGO framework and made EI its central infill criterion for expensive deterministic black-box optimization. In this noise-free setting, \citet{vazquez2010convergence} established asymptotic convergence properties of EI. \citet{bull2011convergence} later derived simple-regret rates for EI and modified epsilon-greedy EI algorithms. We next review literature in two related areas: EI in simulation optimization and noisy EI theory in GP bandits.

\subsection{EI in simulation optimization}
In stochastic simulation, direct use of EGO is inadequate because simulation outputs are noisy. This motivated several extensions of EI. Sequential kriging optimization (SKO) modifies EI through augmented expected
improvement \citep{huang2006global}. Its incumbent is the SK prediction at an
``effective best'' sampled point identified by minimizing an upper posterior quantile. Its
heteroscedastic implementation studied by \citet{jalali2017comparison}
assigns a fixed replication for selected design and returns the sampled
point with the smallest posterior quantile.
EQI defines improvement directly through posterior quantiles
\citep{picheny2013quantile}. Its incumbent and final recommendation are both
based on the smallest upper posterior quantile among previously
sampled points. It may revisit a sampled
design and add replications until the updated EQI falls below a chosen
fraction of its initial value. The quantile level, replication increment,
and stopping fraction are tuning parameters.
TSSO separates search from allocation \citep{quan2013simulation}. Its
modified EI uses the SK prediction at the design with the smallest sample
mean as its incumbent. A fixed iteration budget is divided between the new design and OCBA allocation
among sampled designs \citep{chen2000simulation}. The final recommendation
has the smallest sample mean. eTSSO retains these incumbent and
recommendation but makes the iteration budget adaptive
\citep{pedrielli2020extended}.

Our work also relates to other GP-based simulation optimization frameworks. GP-based search constructs a sampling distribution for discrete
optimization via simulation and establishes global convergence
\citep{sun2014balancing}. Knowledge-gradient methods incorporate correlated
beliefs and common random numbers \citep{xie2016bayesian,pearce2022bayesian}.
GP random search (GPRS) studies continuous optimization via simulation and proves a convergence
rate under known heteroscedastic noise \citep{wang2025gaussian}. These methods
are not EI procedures, but they show how surrogate modeling, sampling, and
noise estimation can be integrated in other simulation optimization
frameworks.

Replication has also been studied outside the EI literature. Classical
SK experiment design allocates a fixed budget over a fixed
set of sites to improve prediction accuracy
\citep{ankenman2010stochastic}. \citet{binois2019replication} use integrated mean squared prediction error criterion to choose between a new design and a replicate.
Their objective is global surrogate accuracy rather than optimization
regret. GPRS proves convergence and a rate under known heteroscedastic noise
variances, while its unknown-variance implementation is studied only
numerically \citep{wang2025gaussian}. CGLO requires every sampled design to
receive a growing number of replications and establishes asymptotic
convergence under known noise variances \citep{meng2022combined}. 

Prior research in this area has established several types of theoretical results. OCBA provides asymptotic allocation theory for finite alternatives \citep{chen2000simulation}. In a normal
ranking-and-selection model, \citet{ryzhov2016convergence} characterize the
asymptotic allocations of EI and relate them to OCBA. eTSSO and several GP
search methods also establish asymptotic guarantees under their respective
sampling structures. The missing result is a
finite-budget analysis that jointly determines the incumbent, replication
order, and final recommendation when unknown heteroscedastic variances are
estimated from replications. Existing quantile levels, batch sizes, and
allocation thresholds are largely treated as tuning parameters.
\citet{jalali2017comparison} also show that an inappropriate recommendation
rule can offset the gains achieved during search.

\subsection{Noisy EI theory in GP bandits}
As noted earlier, noisy GP-bandit theory considers a related but different setting. It usually treats each observation as one round, assumes a fixed noise model, and
focuses on cumulative regret. Replication allocation and final
identification are therefore not its main objects.

One line of work rescales and inflates the posterior uncertainty used inside EI, forcing more exploration. \citep{wang2014theoretical,gupta2022regret,hu2025adjusted}. The drawback is that the acquisition function is no longer the ordinary EI computed from the fitted GP posterior, and the theoretically required inflation can hurt empirical performance \citep{takeno2025regret}. The second route randomizes the incumbent, for example by using the maximum of a posterior sample path \citep{takeno2025regret}. This avoids deterministic variance inflation, but it introduces additional algorithmic randomness and usually requires approximate posterior sampling in continuous or large input spaces. Moreover, the available guarantee is limited to weaker Bayesian regret, averaged over the GP posterior, observation noise, and algorithmic randomness, rather than a high-probability simple-regret guarantee for a returned design. 

Beyond EI, noisy GP-bandit theory is more developed. GP-UCB uses optimism and
controls high-probability cumulative regret through confidence widths and
maximum information gain \citep{srinivas2010gaussian}. Improved GP-UCB and
GP-TS provide frequentist RKHS guarantees \citep{chowdhury2017kernelized},
while GP-PIMS provides a Bayesian posterior-sampling guarantee
\citep{takeno2024posterior}. For pure exploration, maximum variance reduction
returns the posterior-mean maximizer and achieves an order-optimal
simple-regret bound up to logarithmic factors for common kernels
\citep{vakili2021optimal,cai2021lower}. MINI-META and BTS-RED also repeat
observations, but target cumulative regret, computational cost, or a prescribed
variance threshold \citep{calandriello2022scaling,dai2023batch}.

The distinction is therefore both statistical and decision-oriented. Noisy
EI theory in GP bandits explains how an incumbent or acquisition rule can
control regret under a specified observation model. Our setting additionally
requires the algorithm to learn that model through replication and to return a
single sampled design. To our knowledge, existing bandit results do not use
one finite-budget, high-probability simple-regret bound to select the incumbent,
replication level, and terminal recommendation together. 

\section{Preliminaries}\label{sec:preliminaries}
\subsection{Stochastic Kriging and Expected Improvement}
\label{subsec:sk_ei}

We consider the minimization of an unknown latent mean response
$f:\mathcal X\to\mathbb R$ with the following assumption on design space $\mathcal X$:
\begin{assumption}
\label{ass:design_space}
The design space $\mathcal X$ is a nonempty compact subset of
$\mathbb R^d$.
\end{assumption}
SK models the latent response surface $f$ by a GP, $\mathcal{GP}(\mu_0,k_\theta)$, where $\mu_0$ is a prior mean and $k_\theta$ is a positive definite kernel with hyperparameter $\theta$. As this paper adopts a frequentist perspective under which \(f\) is treated as a fixed function in a Hilbert space (see details later in Section \ref{subsec:rkhs}), we follow the common convention in this line of analysis by using a zero-mean GP as the working surrogate, i.e., $\mu_0=0$.
Popular choices of kernel functions include the SE kernel 
$
    k_{\ell}^{\mathrm{SE}}(x,x')
    =
    \exp\!\left(
    -\frac{1}{2}\rho_\ell^2(x,x')
    \right),
$
and the Mat\'ern-$\nu$ kernel
$
    k_{\ell,\nu}^{\mathrm{Mat}}(x,x')
    =
    \frac{2^{1-\nu}}{\Gamma(\nu)}
    \left(\sqrt{2\nu}\rho_\ell(x,x')\right)^\nu
    K_\nu\!\left(\sqrt{2\nu}\rho_\ell(x,x')\right),
$
where $K_\nu$ is the modified Bessel function of the second kind and $\rho_\ell(x,x')
    =
    \left(
    \sum_{j=1}^d
    \frac{(x_j-x'_j)^2}{\ell_j^2}
    \right)^{1/2}$.
For notational simplicity, the kernel variance scale is normalized to one. A separate scale parameter can be incorporated into $\theta$ if needed.

In stochastic simulation, evaluating a decision point $x$ produces random output $Y(x)=f(x)+\varepsilon(x),$ where $\varepsilon(x)$ is a normal noise following $N (0,v(x))$. The variance function $v(x)$ depends on the input location. Given $n$ decision points $x_1,\ldots,x_n$ with $b_i$ replications at $x_i$, denote the sample mean and sample variance at $x_i$ as $\bar Y_i$ and $S_i^2$, respectively. The SK posterior mean and variance are
\begin{equation}\label{skmean}
     \mu_{n}(x)
    =
    k_{n}(x)^\top
    (K_{n}+R_n)^{-1}\bar Y_{1:n},
\end{equation}
\begin{equation}\label{skvariance}
    \sigma_{n}^2(x)
    =
    k_\theta(x,x)
    -
    k_{n}(x)^\top
    (K_{n}+R_n)^{-1}
    k_{n}(x),
\end{equation}   
where
$\bar Y_{1:n}=(\bar Y_1,\ldots,\bar Y_n)^\top$, $k_{n}(x) = \bigl(k_\theta(x,x_1),\ldots,k_\theta(x,x_n)\bigr)^\top$, the matrix $K_{n} = \bigl(k_\theta(x_i,x_j)\bigr)_{i,j=1}^n$ represents the extrinsic uncertainty about the response surface, and $R_n = \mathrm{diag}(S_1^2/b_1,\ldots,S_n^2/b_n)$ represents the intrinsic uncertainty due to stochastic simulation noise.

Given the posterior distribution, EI function at $x$ is
\[
    \mathrm{EI}_n(x;\xi_n) = \mathbb E_n\bigl[I_n(x;\xi_n)\bigr] = \mathbb E_n\bigl[\bigl(\xi_n-f(x)\bigr)_+],
\]
where $\mathbb E_n[\cdot]$ denotes conditional expectation given results of the $n$ decision points, and $\xi_n$ is the incumbent (current best) value. EI computes the expected improvement of response at $x$ with respect to $\xi_n$. If $\sigma_{n}(x)>0$, this has the closed form
\[
    \mathrm{EI}_n(x;\xi_n)
    =
    \sigma_{n}(x)
    \tau\!\left(
    \frac{\xi_n-\mu_{n}(x)}
         {\sigma_{n}(x)}
    \right),
\]
where
$
    \tau(z)=z\Phi(z)+\phi(z),
$
and $\Phi$ and $\phi$ are the standard normal cumulative distribution function (cdf) and probability density function (pdf). If $\sigma_{n}(x)=0$,
$
    \mathrm{EI}_n(x;\xi_n)
    =
    \bigl(\xi_n-\mu_{n}(x)\bigr)_+ 
$. 

\subsection{Information Gain}
\label{subsec:information_gain}
Information gain quantifies the reduction in uncertainty about the latent response values provided by the observations. Let
$
    A=\{x_1,\ldots,x_m\}\subset \mathcal X
$
be a finite set of decision points, and let
$
    f_A=(f(x_1),\ldots,f(x_m))^\top 
$.
Suppose the corresponding sample-mean observations are
$
    \bar Y_A=f_A+\eta_A,$ where
    $\eta_A\sim N(0,R_A)$, and
$
    R_A=\mathrm{diag}(r_1^2,\ldots,r_m^2)
$
is a diagonal noise matrix. The entries $r_i^2$ may differ across
decision points, reflecting heteroscedastic simulation noise and
possibly different numbers of replications.
Let
$
    K_{A}
    =
    \bigl(k_\theta(x_i,x_j)\bigr)_{i,j=1}^m
$.
The information gain from observing $\bar Y_A$ is the mutual information between $\bar Y_A$ and $f_A$:
\[
    \Gamma(\bar Y_A;f_A)
    =
    H(\bar Y_A)-H(\bar Y_A\mid f_A) =
    \frac{1}{2}
    \log\det
    \left(
    I_m+
    R_A^{-1/2}
    K_{A}
    R_A^{-1/2}
    \right),
\]
since both $f_A$ and $\bar Y_A$ are Gaussian. Information gain is connected to posterior variance. For a sequential ordering $x_1,\ldots,x_m$ with noise variances $r_1^2,\ldots,r_m^2$, the chain rule gives
\[
    \Gamma(\bar Y_A;f_A)
    =
    \frac{1}{2}
    \sum_{i=1}^m
    \log
    \left(
    1+
    \frac{
    \sigma_{i-1}^2(x_i)
    }{
    r_i^2
    }
    \right),
\]
where $\sigma_{i-1}^2(x_i)$ is the GP posterior variance at $x_i$ before observing the $i$th sample mean. This identity is the main reason information gain appears in regret analysis: it provides a way to control the cumulative posterior uncertainty at the points selected by the algorithm.

\subsection{Reproducing Kernel Hilbert Spaces}
\label{subsec:rkhs}

The frequentist analysis of GP-based optimization usually measures
smoothness through the reproducing kernel Hilbert space (RKHS) associated
with the kernel. Let $k:\mathcal X\times\mathcal X\to\mathbb R$ be
a positive definite kernel. The RKHS induced by $k$, denoted by
$\mathcal H_k$, is a Hilbert space of functions
$f:\mathcal X\to\mathbb R$ such that, for every $x\in\mathcal X$,
the evaluation functional $f\mapsto f(x)$ is continuous. Equivalently,
there exists a function $k(\cdot,x)\in\mathcal H_k$ satisfying the
reproducing property
\[
    f(x)
    =
    \langle f,k(\cdot,x)\rangle_{\mathcal H_k},
    \qquad
    \forall f\in\mathcal H_k .
\]
By Cauchy's inequality,
$
    |f(x)|
    \le
    \|f\|_{\mathcal H_k}
    \sqrt{k(x,x)}
$.
When the kernel is normalized so that $k(x,x)\le 1$, this implies
$
    |f(x)|\le \|f\|_{\mathcal H_k}
$.
The RKHS norm quantifies the complexity of the latent response
surface. A smaller RKHS norm means that the function is smoother or more compatible with the kernel. 
Throughout this paper, we make the following assumption on function $f$.
\begin{assumption}
\label{ass:rkhs}
The objective function $f\in\mathcal H_{k_\theta}$ with $\|f\|_{\mathcal H_{k_\theta}}\leq B_f$.
The kernel of the RKHS is is continuous on
$\mathcal X\times\mathcal X$. In addition, the kernel is normalized and there is a fixed
constant $k_{\min}>0$ such that
$k_{\min}\le k_\theta(x,x)\leq 1$ for all $x\in\mathcal X$. 
\end{assumption}
For SE kernels, the associated RKHS contains very
smooth functions; for Mat\'ern kernels, the smoothness is controlled
by the parameter $\nu$. In both cases, the RKHS assumption provides
the regularity condition needed to connect GP posterior uncertainty
with finite-budget regret guarantees.

\section{Generic EI Framework and Finite-Sample SK Confidence}
\label{sec:generic_ei_confidence}
\label{sec:algorithm}
\label{sec:sk_unknown_variance}
We begin with a generic EI procedure in which the incumbent, the replication size, and the final recommendation remain unspecified within broad rule classes. We first state the procedure and identify where these three components enter in Section \ref{subsec:generic_ei}. We then establish a finite-sample confidence bound for the SK predictor used by the procedure in Section \ref{subsec:sk_unknown_variance}. The next section uses this confidence bound to derive a simple-regret certificate for this generic procedure.

Throughout this section, $\mu_n$ and $\sigma_n$ denote the SK posterior mean and standard deviation after $n$ simulation batches. The kernel hyperparameter is assumed fixed and known, and $\theta$ is suppressed from the notation. This is a standard convention in theoretical analyses of Bayesian and simulation optimization
\citep{bull2011convergence,wang2025gaussian}. In our experiments, the hyperparameters are updated using the sequentially collected data.

\subsection{A Generic EI Procedure}
\label{subsec:generic_ei}

In this paper, we consider a simulation optimization problem with a total raw simulation budget $T$. The algorithm sequentially selects decision points and allocates $b_T$ simulation replications to each point. For expositional simplicity, we use a common replication size $b_T$ to every selected point. Nonetheless, the framework and its order-wise guarantees require only that the replication numbers at different points have the same asymptotic order in the total budget $T$, rather than being exactly equal. After the budget is exhausted, it returns a final recommendation $\hat x_T$. With this setting, we use $N_T=\left\lfloor \frac{T}{b_T}\right\rfloor$ to represent the maximum number of distinct points with budget $T$. We further require that the budget must cover at least two distinct designs with at least two replications each. Thus, $2\le b_T\le \lfloor T/2\rfloor$, which gives $2\le N_T\le \lfloor T/2\rfloor$. In this generic algorithm, any value of $b_T$ within the range above is permitted. We next detail the general families of the incumbent value and the recommendation in this procedure.

\subsubsection{Global linear-quantile incumbents}
\label{subsec:quantile_benchmark}

In deterministic EI, the incumbent is usually the best observed value. This
choice is not suitable in stochastic simulation because observations contain
random errors. We instead analyze a family of posterior-quantile incumbents.
Let $\mathbf c=(c_0,\ldots,c_{N_T-1})\in\mathbb R^{N_T}$
be a deterministic coefficient sequence. It may depend on $T$, $\delta$, and
the known model constants, but not on the observed simulation outputs. After
$n$ batches, define
\begin{equation*}
\label{eq:general_quantile_incumbent}
\begin{aligned}
    Q_n^{(c_n)}(x):=\mu_n(x)-c_n\sigma_n(x), \
    g_{n+1}^{\mathbf c}
    :=\arg\min_{x\in\mathcal X}Q_n^{(c_n)}(x), \
    \xi_n^{\mathbf c}
    :=Q_n^{(c_n)}(g_{n+1}^{\mathbf c})
      =\min_{x\in\mathcal X}Q_n^{(c_n)}(x).
\end{aligned}
\end{equation*}
Equivalently, $Q_n^{(c_n)}(x)$ is the posterior quantile at level $\Phi(-c_n)$ under the SK model. The incumbent is chosen as $\xi_n^{\mathbf c}$ and the point selected
by EI is $x_{n+1}^{\mathbf c}
    \in
    \arg\max_{x\in\mathcal X}
    \operatorname{EI}_n(x;\xi_n^{\mathbf c})$.
Continuity of the kernel and compactness of $\mathcal X$ ensure that the
minima and maxima in these definitions are attained. Ties are broken
arbitrarily. Here, $c_n>0$, $c_n=0$, and $c_n<0$ correspond respectively to lower-quantile, posterior-mean, and upper-quantile incumbents, representing optimistic, neutral, and uncertainty-penalizing treatments of posterior uncertainty when deciding the incumbent.

Many incumbent rules in the literature can be written in this general form or are closely related to it. Setting $c_n=0$ gives the widely used global best-posterior-mean incumbent \citep{wang2014theoretical}. Some methods instead optimize the quantile criterion only over previously sampled points. For example, SKO and EQI optimize upper quantile functions with negative $c_n$ coefficients in \eqref{eq:general_quantile_incumbent} over the sampled points. Whether the incumbent is optimized over the full domain or the set of sampled points often reflects computational considerations. The former requires continuous optimization, whereas the latter requires only a finite comparison. A similar distinction arises in a related GP based knowledge-gradient (KG) methods. Although their sampling principle differs from EI, they also depend on a current best value. Due to practical considerations such as computational resources, implementations adopting either the global best posterior mean or the best sampled posterior can both be found in the literature \citep{wu2016parallel,jalali2017comparison}.

We optimize the incumbent over the full domain for two reasons. First, for
commonly used differentiable kernels, the continuous problem can be handled in
practice using multistart gradient-based solvers such as L-BFGS-B, whose
computational cost is often modest relative to expensive simulation runs.
Second, full-domain optimization is aligned with our objective of tightening
the simple-regret certificate. With a larger feasible set, it may lead to better convergence behavior. In addition, in our theoretical analysis, it permits direct comparison with the unknown optimizer and avoids an additional sampled-set approximation term.

\subsubsection{Final recommendation}
\label{subsec:actually_sampled_recommendation}

After the simulation budget is exhausted, the algorithm must return one concrete decision point. To enhance the decision maker's trust in the algorithm's final recommendation, we consider the following three aspects. First, we select a previously evaluated point as the final recommendation as with the simulation literature \citep{jalali2017comparison}. This allows the decision maker to observe the effectiveness of the returned strategy through historical simulation data to enhance their confidence in the recommendation. Second, unlike many classical methods that convert cumulative regret to simple regret by randomly returning a previously sampled point, we select from these points based on their performance, which better justifies our choice. Third, as mentioned earlier, when recommending the current best to the decision maker, we account for the impact of residual uncertainty and deliberately avoid solutions with excessive uncertainty.

Let
$\mathcal X_T^{\mathbf c}=\{x_1^{\mathbf c},\ldots,
x_{N_T}^{\mathbf c}\}$ be the sampled set produced by the coefficient
sequence $\mathbf c$. The returned design is
\begin{equation}
\label{return}
    \hat x_T^{\mathbf c}
    \in
    \arg\min_{z\in\mathcal X_T^{\mathbf c}}
    \left\{
        \mu_{N_T}(z)+d_T\sigma_{N_T}(z)
    \right\},
\end{equation}
where $d_T\in\mathbb R$ is a recommendation coefficient. Similar to the general family of incumbents, the choices $d_T=0$, $d_T>0$, and $d_T<0$ correspond to the smallest sampled
posterior mean, an upper posterior quantile, and a lower posterior quantile,
respectively. We leave its value unspecified here and select it from the simple-regret certificate in Section~\ref{sec:simple_regret}.

The generic EI procedure is summarized in Algorithm~\ref{alg:qei_sk}. 
\begin{algorithm}
\caption{Generic Expected Improvement with Stochastic Kriging}
\label{alg:qei_sk}
\begin{algorithmic}[1]
\Require $T$, $b_T$, $v_0$, $\mathbf c$, $d_T$
\State Set $N_T=\lfloor T/b_T\rfloor$ and $\mathcal D_0=\varnothing$; compute $\kappa_T$ from \eqref{eq:kappa_factor}. 
\For{$n=0,1,\ldots,N_T-1$}
    \State Compute the SK posterior $(\mu_n,\sigma_n)$ from $\mathcal D_n$
    \State Compute the incumbent value: $\xi_n^{\mathbf c}=\min_{x\in\mathcal X}\{\mu_n(x)-c_n\sigma_n(x)\}$
    \State Select the next decision point: $x_{n+1}^{\mathbf c}\in\arg\max_{x\in\mathcal X}\operatorname{EI}_n(x;\xi_n^{\mathbf c})$
    \State Run $b_T$ independent replications at $x_{n+1}^{\mathbf c}$ and compute $\bar Y_{n+1}$ and $S_{n+1}^2$
    \State Set $\widetilde v_{n+1}=\max\{v_0,S_{n+1}^2\}$ and $r_{n+1}^2=\widetilde v_{n+1}/b_T$
    \State Update $\mathcal D_{n+1}=\mathcal D_n\cup\{(x_{n+1}^{\mathbf c},\bar Y_{n+1},r_{n+1}^2)\}$
\EndFor
\State Return $\hat x_T^{\mathbf c}$ according to \eqref{return}
\end{algorithmic}
\end{algorithm}

\subsection{Confidence Bound for SK Prediction}
\label{subsec:sk_unknown_variance}
In simulation settings with heteroscedastic normal noise, the variance function $v(x)$ is unknown and may vary over the design space. We make the following assumption on the noise levels:
\begin{assumption}
\label{ass:noise_variance}
There exist constants $v_\star$ and $V_\star$ such that
\[
    0<v_\star\le v(x)\le V_\star<\infty,
    \qquad x\in\mathcal X.
\]
\end{assumption}
We highlight that this assumption only requres the noise variances to be uniformly bounded. Neither $v_\star$ nor $V_\star$ is assumed to be known to the algorithm.

At each selected decision point, the algorithm collects a batch of $b_T$ independent replications. At the location selected for batch $q$, we can compute the sample mean $\bar Y_q$ and sample variance $S_q^2$ from these replications. Let $\mathcal H_{q-1}$ contain all information available before this batch, including its selected location $x_q$. Conditional on $\mathcal H_{q-1}$, under Gaussian simulation noise,
\[
    \bar Y_q\sim N\!\left(f(x_q),\frac{v(x_q)}{b_T}\right),
    \qquad
    \frac{(b_T-1)S_q^2}{v(x_q)}\sim\chi^2_{b_T-1},
\]
and $\bar Y_q$ and $S_q^2$ are conditionally independent.

In conventional SK implementations, the unknown intrinsic variance $v(x_q)$ is estimated by $S_q^2$, so $S_q^2/b_T$ is used as a plug-in estimate of the noise variance of the sample mean $\bar Y_q$
\citep{ankenman2010stochastic}. We use a regularized estimate $\widetilde v_q:=\max\{v_0,S_q^2\}$.
The plug-in estimate of the noise variance of $\bar Y_q$ is $r_q^2=\frac{\widetilde v_q}{b_T}$. Here, $v_0$ is a positive variance floor independently of the unknown noise bounds. It regularizes the plug-in noise matrix and need not be a lower bound on $v(x)$. Practically, it plays a role similar to a nugget or jitter term to stabilize covariance-matrix inversion when the estimated noise variances are very small.

The confidence analysis must account for possible underestimation of $v(x_q)$ with sample variances. Let $\nu_T:=b_T-1$ and $\alpha_T:=\frac{\delta}{4N_T}$. For $p\in(0,1)$, let $\chi^2_{\nu,p}$ denote the $p$-quantile of a chi-squared random variable with $\nu$ degrees of freedom. Define
\begin{equation}
\label{eq:kappa_factor}
    \kappa_T:=
    \max\left\{
        1,\frac{\nu_T}{\chi^2_{\nu_T,\alpha_T}}
    \right\},
    \qquad
    h_T:=
    \max\left\{
        1,\frac{\chi^2_{\nu_T,1-\alpha_T}}{\nu_T}
    \right\}.
\end{equation}
These two factors are used to control potential under or over estimate of the noise variance. Specifically, $\kappa_T$ accounts for sample-variance estimates that are too small and could otherwise produce overly narrow confidence bands, while $h_T$ controls estimates that are too large and can inflate posterior uncertainty. In addition, define $M:=\max\{v_0,V_\star\}$, and $W_T:=h_TM$, which are used in the performance analysis. 

\begin{lem}
\label{lem:plugin_variance}
With probability at least $1-\delta/2$, the following
inequalities hold for every $q\le N_T$:
\[
    \frac{v(x_q)}{\kappa_T}
    \le S_q^2
    \le h_Tv(x_q),
    \qquad
    v(x_q)\le\kappa_T\widetilde v_q.
\]
In addition, under Assumption \ref{ass:noise_variance}, $ \frac{v_0}{b_T}
    \le r_q^2
    \le\frac{W_T}{b_T}$.
\end{lem}

Although this lemma is valid for every admissible $b_T$, the sizes of $\kappa_T$ and $h_T$ determine how much the under and over estimation of the noise variance influences the SK confidence bound. We therefore first characterize how $\kappa_T$ and $h_T$ depends on the replication level.

\begin{lem}
\label{lem:plugin_correction_order}
Fix $\delta\in(0,1)$, and let $T\to\infty$. For each $T$, let $b_T$ be an
integer satisfying $2
    \le
    b_T
    \le
    \left\lfloor\frac{T}{2}\right\rfloor$.
Then the following statements hold:
\begin{enumerate}
    \item If $b_T=o(\log T)$, then
    $\kappa_T, h_T\longrightarrow\infty$.
    
    \item If $b_T=\Theta(\log T)$, then
    $\kappa_T, h_T=O(1)$.
    
    \item If $\frac{b_T}{\log T}
        \longrightarrow\infty$,
    then $\kappa_T, h_T\longrightarrow1$.
\end{enumerate}
In particular, if $b_T=b\ge2$ is fixed, then $\kappa_T
    =
    \Theta\left(
        T^{2/(b-1)}
    \right)$ and $h_T = \Theta(\log T)$.
\end{lem}

Lemma~\ref{lem:plugin_correction_order} isolates the effect of
replication on the variance-under / overestimation correction. A sublogarithmic
replication level makes these corrections diverge, whereas logarithmic
replication keeps them bounded. A superlogarithmic level can make the
corrections approach one, but it also leaves fewer design evaluations for
spatial exploration. We next show how these factors enters the SK
confidence width. Section~\ref{subsec:repli} then combines this effect with
the number of design evaluations and the information-gain terms in the
simple-regret bound.

With $n$ batch observations, the data used by the surrogate model are
$
    \mathcal D_n
    =
    \{(x_q,\bar Y_q,r_q^2):q=1,\ldots,n\}
$,
where $r_q^2=\widetilde v_q/b_T$. The SK posterior mean and variance are therefore the quantities in equations \eqref{skmean} and \eqref{skvariance}, with $R_n = \operatorname{diag}(r_1^2,\ldots,r_n^2)$. Thus, the surrogate uses the standard SK covariance formula with clipped plug-in estimates of the intrinsic variances. The correction $\kappa_T$ is incorporated into the confidence coefficient for the SK predictor rather than into the plug-in noise matrix. We derive the following confidence bound for the SK predictor. 

\begin{thm}
\label{thm:plugin_sk_confidence}
Under Assumptions \ref{ass:design_space}, \ref{ass:rkhs}, and \ref{ass:noise_variance}, for $n=1,\ldots,N_T$, define
\begin{equation}
\label{eq:sk_maximum_information_gain}
    \gamma_n^{\mathrm{SK}}
    =
    \sup_{1\le m\le n}
    \sup_{a_1,\ldots,a_m\in\mathcal X}
    \frac12
    \log\det
    \left(
        I_m+
        \frac{b_T}{v_0}K_A
    \right),
    \qquad
    \gamma_0^{\mathrm{SK}}=0,
\end{equation}
where $K_A=\bigl(k_\theta(a_i,a_j)\bigr)_{i,j=1}^m$. Let
\begin{equation}
\label{eq:lambda_sequence}
    \Lambda_n
    =
    B_f+
    \sqrt{
        2\kappa_T
        \left(
            \gamma_n^{\mathrm{SK}}
            +
            \log\frac{2}{\delta}
        \right)
    },
    \qquad n=0,\ldots,N_T.
\end{equation}
Then $(\Lambda_n)_{n=0}^{N_T}$ is nondecreasing. With probability at least $1-\delta$, all variance inequalities in Lemma~\ref{lem:plugin_variance} and the following confidence inequalities hold jointly:
\[
    |f(x)-\mu_n(x)|
    \le
    \Lambda_n\sigma_n(x),
    \qquad
    \forall x\in\mathcal X,
    \quad
    \forall n=0,\ldots,N_T.
\]
\end{thm}

The proof of this lemma uses conditional independence of Gaussian sample mean and variance, and a time-uniform self-normalized inequality \citep{kirschner2018information}. This theorem provides the statistical foundation for the subsequent algorithm design. The uncertainty from estimating the unknown variances enters through $\kappa_T$, accounting for variance underestimation in the confidence coefficient $\Lambda_n$. The upper-tail factor $h_T$ controls overestimation of the noise variances and is used in the subsequent simple-regret analysis to bound posterior uncertainty. When the variances are known, we will have $\kappa_T=1$, and the result reduces to the standard time-uniform RKHS confidence bound, up to confidence-allocation constants. If the variance is constant, the bound in Theorem \ref{thm:plugin_sk_confidence} gives its usual homoscedastic form \citep{kirschner2018information}.

\section{A Unified Simple-Regret Certificate}
\label{sec:simple_regret}

Algorithm~\ref{alg:qei_sk} leaves the incumbent sequence, replication size, and final recommendation unspecified within their respective rule classes. We now use the finite-sample confidence guarantee in Theorem~\ref{thm:plugin_sk_confidence} to derive a common high-probability simple-regret certificate before selecting any of these three components. For any $b_T$, $\mathbf c=(c_0,\ldots,c_{N_T-1})$, and $d_T$, define
\[
    \mathcal E_T
    :=
    \left\{
        |f(x)-\mu_n(x)|
        \le
        \Lambda_n\sigma_n(x),\quad
        \forall x\in\mathcal X,\quad
        n=0,\ldots,N_T
    \right\},
\]
\[
    \mathcal V_T:=
    \left\{
        \frac{v(x_q)}{\kappa_T}\le S_q^2\le h_Tv(x_q),
        \quad q=1,\ldots,N_T
    \right\}.
\]
The event $\mathcal V_T$ implies $v(x_q)/\kappa_T\le\widetilde v_q\le W_T$ for every $q$. By Theorem~\ref{thm:plugin_sk_confidence}, $\mathcal G_T:=\mathcal E_T\cap\mathcal V_T$ has probability at least $1-\delta$. All subsequent deterministic inequalities are stated on $\mathcal G_T$.

The derivation proceeds in two stages. We first compare the posterior uncertainties at the incumbent point and EI-selected points and invert the EI function to obtain a regret certificate for each selected point in Section \ref{subsec:sampled_point_certificate}. We then use uniform variance bounds and information gain to identify a well-certified sampled point and compare it with the terminal recommendation. Combining these two stages yields the final simple-regret certificate in Section \ref{subsec:general_simple_regret}.

\subsection{Regret certificate for a sampled point}
\label{subsec:sampled_point_certificate}

For any coefficient sequence $\mathbf c$, as the terminal recommendation is restricted to sampled points, we first derive a regret certificate for each point selected by EI. For each $n<N_T$, define $\rho_n^{\mathbf c}
    :=
    \frac{\sigma_n(x_{n+1}^{\mathbf c})}
         {\sigma_n(g_{n+1}^{\mathbf c})}$.
The denominator is positive because every point has a positive noise proxy.
The following lemma shows that the EI-selected point cannot have less
posterior uncertainty than the incumbent point.
\begin{lem}
\label{lem:tail_cancellation}
For every $n=0,\ldots,N_T-1$, $\rho_n^{\mathbf c}\ge1$. Equivalently, $\sigma_n(g_{n+1}^{\mathbf c})
    \le
    \sigma_n(x_{n+1}^{\mathbf c})$.
\end{lem}
This comparison follows from the global minimization of the linear quantile
and therefore holds for every real coefficient $c_n$ in the general incumbent
family. To convert this comparison into a regret certificate, we next invert
the function $\tau$ appearing in EI. It is a prerequisite for bounding on the gap between each sampled point and the incumbent.

\begin{lem}
\label{lem:q_bound}
For $c\in\mathbb R$ and $r\ge1$, let $q_c(r)$ be the unique real number
satisfying $\tau(-q_c(r))
    =
    \frac{\tau(-c)}{r}$.
Then $q_c(r)\ge c$ and
\[
    q_c(r)
    \le
    \begin{cases}
        \sqrt{c^2+2\log r}, & c\ge0,\\[2pt]
        \sqrt{2\log r}, & c<0.
    \end{cases}
\]
\end{lem}

For each step $n$, write $s_n^{g}:=\sigma_n(g_{n+1}^{\mathbf c})$ and $s_n^{x}:=\sigma_n(x_{n+1}^{\mathbf c})$. Lemma~\ref{lem:tail_cancellation} gives $\rho_n^{\mathbf c}=s_n^{x}/s_n^{g}\ge1$, allowing Lemma~\ref{lem:q_bound} to control the gap between the selected point and the incumbent. For $\lambda>0$, let $z_\lambda$ be the unique solution of $\frac{\phi(z_\lambda)}{\Phi(z_\lambda)}=\lambda$, and set $c_\lambda^\circ:=-z_\lambda$. Define
\begin{equation*}
\label{eq:additional_coefficient}
    A(\lambda,c)
    :=
    \begin{cases}
        0, & c\ge\lambda,\\[2pt]
        \lambda-c, & c_\lambda^\circ<c<\lambda,\\[5pt]
        \displaystyle\frac{\tau(-c)}{\Phi(z_\lambda)},
        & c\le c_\lambda^\circ.
    \end{cases}
\end{equation*}
Based on this definition, the following lemma provides regret certificate for a sampled EI point.
\begin{lem}
\label{lem:ei_point_certificate}
For every $n=0,\ldots,N_T-1$,
\begingroup
\setlength{\abovedisplayskip}{6pt}
\setlength{\belowdisplayskip}{6pt}
\setlength{\abovedisplayshortskip}{3pt}
\setlength{\belowdisplayshortskip}{6pt}
\begin{align}
    f(x_{n+1}^{\mathbf c})-f^\star
    &\le
    \left\{\Lambda_n+q_{c_n}(\rho_n^{\mathbf c})\right\}
    \sigma_n(x_{n+1}^{\mathbf c})
    +(\xi_n^{\mathbf c}-f^\star)_+,
    \label{eq:general_point_decomposition}\\
    (\xi_n^{\mathbf c}-f^\star)_+
    &\le
    (\Lambda_n-c_n)_+\sigma_n(x^\star),
    \label{eq:unobserved_variance_term}\\
    (\xi_n^{\mathbf c}-f^\star)_+
    &\le
    A(\Lambda_n,c_n)\sigma_n(x_{n+1}^{\mathbf c}).
    \label{eq:selected_variance_term}
\end{align}
\endgroup
Consequently,
\begin{equation}
\label{eq:general_point_certificate}
    f(x_{n+1}^{\mathbf c})-f^\star
    \le
    \left\{
        \Lambda_n+
        q_{c_n}(\rho_n^{\mathbf c})+
        A(\Lambda_n,c_n)
    \right\}
    \sigma_n(x_{n+1}^{\mathbf c}).
\end{equation}
\end{lem}

A general incumbent can lie above $f^\star$ even on the SK predictor confidence event. This produces the term $(\xi_n^{\mathbf c}-f^\star)_+$ in \eqref{eq:general_point_decomposition}. Equation~\eqref{eq:unobserved_variance_term} shows the direct effect of the incumbent coefficient, but it depends on the posterior variance at the unknown optimizer. The information-gain argument used in GP-based optimization literature instead controls posterior variances at sampled points. Therefore, \eqref{eq:selected_variance_term} transfers this uncertainty term to the EI-selected point at the cost of the factor $A(\Lambda_n,c_n)$. The proof in Appendix F further indicates that no smaller coefficient than $ A(\Lambda_n,c_n)$ exists under the definition of the incumbent and the SK predictor confidence bound in Theorem~\ref{thm:plugin_sk_confidence}. It remains to control the posterior standard deviations at the selected points and connect a well-certified sampled point to the terminal recommendation.

\subsection{From sampled points to final simple regret}
\label{subsec:general_simple_regret}

The pointwise certificate in \eqref{eq:general_point_certificate} still depends on the data-dependent variance ratio $\rho_n^{\mathbf c}$. We first replace this ratio with a uniform bound and then control the accumulated posterior uncertainties at the selected points.

\begin{lem}
\label{lem:variance_ratio}
For every $x\in\mathcal X$ and $n=0,\ldots,N_T$,
\[
    \frac{k_{\min}}{1+nb_T/v_0}
    \le\sigma_n^2(x)\le1,
\ \
    1\le\rho_n^{\mathbf c}
    \le\sqrt{\frac{1+nb_T/v_0}{k_{\min}}}.
\]
\end{lem}
Define
\[
    \ell_n:=
    \log\left(\frac{1+nb_T/v_0}{k_{\min}}\right)
\
\text{and} \
    \bar q_n(c)
    :=
    \begin{cases}
        \sqrt{c^2+\ell_n}, & c\ge0,\\[2pt]
        \sqrt{\ell_n}, & c<0.
    \end{cases}
\]
Lemmas~\ref{lem:q_bound} and \ref{lem:variance_ratio} imply $q_{c_n}(\rho_n^{\mathbf c})\le\bar q_n(c_n)$, removing the data-dependent ratio from the pointwise certificate. It remains to control the posterior standard deviations at the selected points, which is achieved through information gain.
\begin{lem}
\label{lem:variance_sum}
For the points selected under $\mathbf c$,
\[
    \sum_{n=0}^{N_T-1}
    \sigma_n^2(x_{n+1}^{\mathbf c})
    \le
    \frac{2\gamma_{N_T}^{\mathrm{SK}}}
    {\log(1+b_T/W_T)}.
\]
\end{lem}
As in standard BO analyses, Lemma~\ref{lem:variance_sum} controls the cumulative posterior uncertainty at the points when they are selected. Together with the pointwise certificate, it ensures that at least one sampled point has both a small posterior standard deviation and a small regret certificate. To collect the coefficient-dependent terms, define
\begin{equation*}
\label{eq:general_prefactor}
    D_T(\mathbf c)
    :=
    \max_{0\le n<N_T}
    \left\{
        \Lambda_n+
        \bar q_n(c_n)+
        A(\Lambda_n,c_n)
    \right\}.
\end{equation*}
We also define the uncertainty scales used to compare this well-certified sampled point with the terminal recommendation:
\begin{equation*}
\label{eq:terminal_uncertainty_scales}
\begin{aligned}
    \eta_T
    :=
    \sqrt{
        \frac{2\gamma_{N_T}^{\mathrm{SK}}}
        {N_T\log(1+b_T/W_T)}
    }, \ \ 
    \omega_T
    :=
    \sqrt{\frac{W_T}{b_T+W_T}},\ \ \text{and} \ \
    \zeta_T
    :=
    \min\{\eta_T,\omega_T\}.
\end{aligned}
\end{equation*}

Lemma~\ref{lem:ei_point_certificate} and
Lemma~\ref{lem:variance_sum} together guarantee a sampled point whose simple
regret is at most $D_T(\mathbf c)\eta_T$. After all observations have been
incorporated, this point has posterior standard deviation at most
$\zeta_T=\min\{\eta_T,\omega_T\}$, while every sampled point, including the
terminal recommendation, has posterior standard deviation at most
$\omega_T$. Because the terminal recommendation minimizes its linear
posterior quantile over the sampled points, it can be compared with this
well-certified point. Applying the terminal SK predictor confidence bound
then gives the following result.

\begin{thm}
\label{thm:global_quantile_family_certificate}

Let $b_T$ be any integer satisfying $2\le b_T\le\left\lfloor\frac{T}{2}\right\rfloor$. Under Assumptions 1 $\sim$ 3, for each fixed deterministic incumbent coefficient sequence $\mathbf c$ and $d_T$, with probability at least $1-\delta$,
\begin{equation}
\label{eq:joint_quantile_certificate}
\begin{aligned}
    f(\hat x_T^{\mathbf c,d_T})-f^\star
    \le\;
    D_T(\mathbf c)\eta_T
    +
    (\Lambda_{N_T}+d_T)_+\zeta_T
    +
    (\Lambda_{N_T}-d_T)_+\omega_T.
\end{aligned}
\end{equation}
\end{thm}
Theorem~\ref{thm:global_quantile_family_certificate} links the three design components through one finite-budget simple-regret certificate. The first term is generated by the search process and contains the incumbent-dependent factor $D_T(\mathbf c)$. The last two terms measure the transfer from a well-certified sampled point to the terminal recommendation and contain the terminal coefficient $d_T$. The replication size affects all three terms through $N_T$, $\kappa_T$, $\Lambda_n$, $\gamma_{N_T}^{\mathrm{SK}}$, and the uncertainty scales $\eta_T$, $\zeta_T$, and $\omega_T$. Because none of the three components has yet been calibrated, the result applies to the generic EI procedure. The next section selects their parameters to tighten this certificate.


\section{Bound-Guided Design Choices and Regret Rates}
\label{sec:designchoices}
This section uses Theorem~\ref{thm:global_quantile_family_certificate} as a common criterion
for selecting the three design components. The first two subsections
calibrate the terminal and incumbent coefficients from their respective
parts of the certificate. Unlike these two coefficients, the replication
size depends on the covariance kernel. After fixing the first two components, we therefore select the
replication size by minimizing a tractable upper bound under mild conditions for kernels.
The final subsection derives the resulting rates for specific kernel classes.

\subsection{Choice of the terminal coefficient}
\label{subsec:terminal_coefficient_choice}

For fixed $\mathbf c$ and $b_T$, define $G_T(d)
    :=
    (\Lambda_{N_T}+d)_+\zeta_T
    +
    (\Lambda_{N_T}-d)_+\omega_T$.
Because $\zeta_T\le\omega_T$,
\[
G_T(d)-G_T(\Lambda_{N_T})
=
\begin{cases}
    (\Lambda_{N_T}-d)\omega_T-2\Lambda_{N_T}\zeta_T,
        & d\le-\Lambda_{N_T},\\[2mm]
    (\Lambda_{N_T}-d)(\omega_T-\zeta_T),
        &-\Lambda_{N_T}\le d\le\Lambda_{N_T},\\[2mm]
    (d-\Lambda_{N_T})\zeta_T,
        &d\ge\Lambda_{N_T}.
\end{cases}
\]
All three expressions are nonnegative. In the first case,
$\Lambda_{N_T}-d\ge2\Lambda_{N_T}$, so the difference is at least
$2\Lambda_{N_T}(\omega_T-\zeta_T)$. Therefore, $d_T=\Lambda_{N_T}$
is a global bound-minimizing choice for the displayed uniform certificate in \eqref{eq:joint_quantile_certificate}.
It yields the sampled design with the smallest posterior upper confidence
value. Henceforth, we select $d_T=\Lambda_{N_T}$ to return the final recommendation written as
$\hat x_T^{\mathbf c}
    :=
    \hat x_T^{\mathbf c,\Lambda_{N_T}}$.

\subsection{Choice of the incumbent coefficient}
\label{subsec:coefficient_choice}
Having calibrated the terminal coefficient, we next select the incumbent coefficient sequence $\mathbf c$. Under $d_T=\Lambda_{N_T}$ and the feasible range $2\le b_T\le\lfloor T/2\rfloor$, 
the simple regret has the form:
\begin{equation*}
\label{eq:general_simple_regret}
    f(\hat x_T^{\mathbf c})-f^\star
    \le
    2\left\{
        2\Lambda_{N_T}+D_T(\mathbf c)
    \right\}
    \sqrt{
        \frac{
            b_T\gamma_{N_T}^{\mathrm{SK}}
        }{
            T\log(1+b_T/W_T)
        }
    }.
\end{equation*}
With this form, we observe that to minimize the bound for simple regret, we need to select $\mathbf c$ to minimize $D_T(\mathbf c)$. This result is summarized in the following lemma.
\begin{lem}
\label{lem:incumbent_coefficient_optimality}
For every $n=0,\ldots,N_T-1$ and $c\in\mathbb R$, we have $\bar q_n(c)+A(\Lambda_n,c)
    \ge
    \bar q_n(\Lambda_n)$.
\end{lem}
Recall that $D_T(\mathbf c) = \max_{0\le n<N_T}
    \left\{
        \Lambda_n+
        \bar q_n(c_n)+
        A(\Lambda_n,c_n)
    \right\}$ and $A(\Lambda_n,\Lambda_n)=0$. Consequently, the choice $c_n= \Lambda_n$ minimizes every term $\Lambda_n+
        \bar q_n(c_n)+A(\Lambda_n,c_n)$ in the definition of $D_T(\mathbf c)$. Hence, the sequence $\mathbf c=(\Lambda_0,\ldots,\Lambda_{N_T-1})$
minimizes $D_T(\mathbf c)$ over $\mathbb R^{N_T}$.
For this choice, we omit the superscript $\mathbf c$ from the selected points, sampled set, and final recommendation.

\begin{cor}
\label{thm:observed_simple_regret}
Set $c_n=\Lambda_n$ for $n=0,\ldots,N_T-1$ and
$d_T=\Lambda_{N_T}$. Write
$\hat x_T:=\hat x_T^{\mathbf c,\Lambda_{N_T}}$. Under the assumptions of
Theorem~\ref{thm:global_quantile_family_certificate}, with probability at
least $1-\delta$,
\begin{equation*}
\label{eq:calibrated_simple_regret}
    f(\hat x_T)-f^\star
    \le
    8\sqrt{
        \frac{
            \left[
                \Lambda_{N_T}^2+
                \log\left(
                    \frac{1+T/v_0}{k_{\min}}
                \right)
            \right]
            b_T\gamma_{N_T}^{\mathrm{SK}}
        }{
            T\log(1+b_T/W_T)
        }
    }.
\end{equation*}
\end{cor}

Thus, the proposed algorithm uses $c_n=\Lambda_n$ during search and $d_T=\Lambda_{N_T}$ at termination. We highlight that the calibrated incumbent and terminal rules treat uncertainty in opposite ways. During the search, EI uses the best lower posterior quantile as its incumbent, whereas the terminal rule returns the sampled point with the smallest upper posterior quantile. This contrast reflects the different roles of intermediate simulation experiments and the final implemented decision. As intermediate evaluations are exploratory and are not themselves deployed in the real system, the incumbent can treat remaining uncertainty optimistically and keep potentially good regions under consideration. The final recommendation, by contrast, is intended for real-world implementation and therefore treats remaining uncertainty conservatively.

\subsection{Kernel-Dependent Choice of the Replication Size}
\label{subsec:repli}
Unlike the incumbent and terminal coefficients, the replication size depends on the covariance kernel used by the SK predictor. Increasing the replication size improves the estimation of the local noise variances and reduces the noise variance of each sample mean. However, under a fixed raw simulation budget, it also reduces the number of EI iterations. The effect of the latter reduction is governed by the kernel-specific information gain. To make this tradeoff explicit, consider any candidate integer $b
    \in
    \mathcal B_T
    :=
    \left\{
        2,\ldots,\left\lfloor\frac{T}{2}\right\rfloor
    \right\}$. Define
    $N_T(b)
    :=
    \left\lfloor\frac{T}{b}\right\rfloor$, $\nu_b:=b-1$,
    $\alpha_T(b)
    :=
    \frac{\delta}{4N_T(b)}$,
    $\kappa_T(b)
    :=
    \max\left\{
        1,\,
        \frac{\nu_b}
        {\chi^2_{\nu_b,\,\alpha_T(b)}}
    \right\}$, $h_T(b):=\max\left\{
    1,\frac{\chi^2_{\nu_b,1-\alpha_T(b)}}{\nu_b}
\right\}$, and $W_T(b):=Mh_T(b)$.
For the same candidate $b$, denote $\gamma_T(b)$ and $\Lambda_T(b)$ as the information gain and the confidence coefficient for the SK predictor with $b_T= b$, $n=N_T(b)$ and $\kappa_T=\kappa_T(b)$ in equations \eqref{eq:sk_maximum_information_gain} and \eqref{eq:lambda_sequence}.
Thus, increasing $b$ reduces the number of EI evaluations $N_T(b)$, but it also makes
the sample-variance correction $\kappa_T(b)$ smaller and reduces the noise variance
of each sample mean to order $1/b$ in $\gamma_T(b)$.

After setting $c_n=\Lambda_n$ and $d_T=\Lambda_{N_T}$ using the quantities associated with the same candidate $b$, Corollary~\ref{thm:observed_simple_regret} gives the following certificate:
\begin{equation}
\label{eq:b_dependent_regret_bound}
\begin{aligned}
    \mathcal R_T(b)
    &:=
    8
    \sqrt{
        \frac{
            \left[
                \Lambda_T^2(b)
                +
                \log\left(\frac{1+T/v_0}{k_{\min}}\right)
            \right]
            b\gamma_T(b)
        }{
            T\log\left(1+b/W_T(b)\right)
        }
    }.
\end{aligned}
\end{equation}
For every predetermined $b\in\mathcal B_T$, Corollary~1 gives $f(\hat x_T)-f^\star
    \le
    \mathcal R_T(b)$
with probability at least $1-\delta$. The exact maximum information gain in \eqref{eq:b_dependent_regret_bound} is generally unavailable in closed form. However, there exist upper bounds in the literature for the information gain. Define $U_T(b):=\frac{bN_T(b)}{v_0}$. For a kernel $k$, the
information-gain bound can be denoted as $G_k(U_T(b))$ \citep{vakili2021information}. Such kernel-specific information-gain bounds are mainly derived by using the smoothness of the kernel or the decay of its eigenvalues. Their main purpose is to describe how information gain grows as the number of observations increases. Although these bounds remain valid when only a small number of
observations are available, they may then be loose. To obtain a fair comparison among different choices of $b_T$ and $N_T$, we derive the following bound for $\gamma_T(b)$ that can be tighter with smaller $N_T$.

\begin{lem}
\label{lem:finite_batch_information_gain}
Under Assumption \ref{ass:rkhs}, for any candidate replication
number $b$, 
\begin{equation*}
\label{eq:finite_batch_information_bound}
    \gamma_T(b)
    \le
    \frac{N_T(b)}{2}
    \log\left(
        1+\frac{b}{v_0}
    \right).
\end{equation*}
\end{lem}
Unlike the kernel-specific bound, Lemma~\ref{lem:finite_batch_information_gain} only uses the number of batch observations and the normalization of the kernel. The result can be interpreted as follows: each batch mean contributes at most $\frac12
    \log\left(
        1+\frac{b}{v_0}
    \right)$
to the information gain, and there are at most $N_T(b)$ such observations.
This bound becomes useful when $b$ is large and the number of
batches is small. We then combine the two bounds and define
\begin{equation*}
\label{eq:kernel_information_envelope}
\begin{aligned}
    \overline\gamma_T^{\,k}(b)
    :=
    \min\Bigg\{
        G_k\bigl(U_T(b)\bigr),\,
        \frac{N_T(b)}{2}
        \log\left(
            1+\frac{b}{v_0}
        \right)
    \Bigg\}.
\end{aligned}
\end{equation*}

After determining the incumbent and terminal coefficients, we introduce the following mild conditions on the kernel-specific information-gain bound. 
\begin{assumption}
\label{assu:kernel}
Let $g_T
    :=
    G_k\left(\frac{T}{v_0}\right)$. The kernel-specific bound $G_k$ satisfies the following
conditions:
\begin{enumerate}
    \item $G_k$ is nondecreasing, and there is a constant $C_k<\infty$ such that $G_k(2u)
        \le
        C_kG_k(u)$ for all sufficiently large $u$;
    \item $\log T
        =
        O(g_T)$;
    \item
    $g_T
        =
        o\left(
            \sqrt{
                T\log\log T
            }
        \right)$.
\end{enumerate}
\end{assumption}
The first condition requires the spectral information-gain bound to change by at most a constant factor when its argument changes by a constant factor. The second condition
ensures that the information-gain term is at least of logarithmic order. The
third condition requires it to grow more slowly than
$\sqrt{T\log\log T}$. These conditions are satisfied
by most standard kernel classes. They include finite-rank kernels, such as
linear and fixed-degree polynomial kernels; kernels with exponentially
decaying eigenvalues, including the SE kernel; and kernels
with sufficiently fast polynomial eigenvalue decay. In the last class, the
required growth condition holds when the polynomial decay exponent is greater
than two, which includes Mat\'ern kernels with $\nu>d/2$. Finite-rank and
exponential-eigendecay kernels have logarithmic or polylogarithmic information
gain, while sufficiently fast polynomial eigendecay gives a sub-square-root
information-gain order. These properties ensure that the three conditions hold. Detailed verification for
these kernel classes is provided in Appendix M.

We next choose the order of the replication size by minimizing a tractable kernel-specific upper bound. For each candidate $b$, replace the exact information gain with its upper bound $\overline\gamma_T^{\,k}(b)$. Then, the optimal replication number preferred by the displayed bound is
\[
    b_T^{\mathrm{bd}}
    \in
    \arg\min_{b\in\mathcal B_T}
    \overline{\mathcal R}_T^{\,k}(b)
\]
where $\overline{\mathcal R}_T^{\,k}(b)$ is the upper bound of $\mathcal R_T(b)$ with
$\gamma_T(b)$ replaced by its upper bound $\overline\gamma_T^{\,k}(b)$ in \eqref{eq:b_dependent_regret_bound}. 
The optimal order of $b_T$ is presented in the following theorem. 
\begin{thm}
\label{thm:general_kernel_replication}
For $\delta\in(0,1)$, under Assumptions \ref{ass:design_space}, \ref{ass:rkhs}, \ref{ass:noise_variance}, and \ref{assu:kernel}, as $T\to\infty$, $b_T^{\mathrm{bd}}
    =
    \Theta\left(
        \log\frac{T}{\delta}
    \right)$.
\end{thm}
From this theorem and its proof, we find that replication smaller than logarithmic order produces a large correction $\kappa$, whereas replication larger than logarithmic order sacrifices spatial exploration without improving the leading order of the bound. Together with the calibrated $\mathbf c$ and $d_T$, these choices form a coherent algorithmic design: an optimistic incumbent for exploration, logarithmic replication for variance learning, and a conservative recommendation for implementation.

Under the rule
$b_T= \Theta(\log(T/\delta))$, Corollary~\ref{thm:observed_simple_regret} gives
\begin{equation}
\label{eq:general_kernel_rate_after_replication}
    f(\hat x_T)-f^\star
    =
    O\left(
        g_T
        \sqrt{
            \frac{
                \log(T/\delta)
            }{
                T\log\log(T/\delta)
            }
        }
    \right).
\end{equation}
This bound is consistent with related results in the literature. For variance-rescaled noisy EI, an $O(\gamma_T\sqrt{T})$ cumulative-regret bound has been established (see discussions in \citealp{takeno2025regret}). If one of the sampled points is returned uniformly at random, its conditional expected regret is therefore $O(\gamma_T/\sqrt{T})$, matching the polynomial order of our bound. Notably, these results assume known homoscedastic noise, whereas our guarantee accommodates unknown heteroscedastic variances learned through replications, further demonstrating the strength of our result. The next subsection makes the resulting guarantee explicit by deriving simple-regret rates for different kernels.

\subsection{Kernel-specific rates and comparison with lower bounds}
\label{subsec:kernel_rates}

We now substitute the corresponding kernel-specific information-gain bounds
into \eqref{eq:general_kernel_rate_after_replication}. Constants depending on
$B_f$, $v_{\min}$, $\bar v$, and the kernel parameters are suppressed. The
notation $\widetilde O$ suppresses logarithmic factors.

\paragraph{SE kernels.}
According to \cite{vakili2021information}, the information gain bound is $g_T
    =
    O\left(
        (\log T)^{d+1}
    \right)$.
Therefore,
\begin{equation}
\label{eq:se_kernel_rate}
    f(\hat x_T)-f^\star
    =
    O\left(
        \frac{
            (\log T)^{d+3/2}
        }{
            \sqrt{T\log\log T}
        }
    \right).
\end{equation}
This polynomial order can be compared with the minimax lower bound under SE kernel. Even when
the Gaussian noise variance $v_0>0$ is known and constant,
\citet[Theorem~1]{cai2021lower} show that an algorithm needs
\[
    T
    =
    \Omega\left(
        \frac{v_0}{\varepsilon^2}
        \left(
            \log\frac{B_f}{\varepsilon}
        \right)^{d/2}
        \log\frac1\delta
    \right)
\]
observations to attain simple regret at most $\varepsilon$ with probability
at least $1-\delta$ over the SE RKHS ball. For fixed
$B_f$, $v_0$, and $\delta$, the unavoidable polynomial dependence is
$T^{-1/2}$. The lower bound permits adaptive sampling and an arbitrary
returned point, and therefore also applies to the present setting. Hence,
\eqref{eq:se_kernel_rate} is minimax optimal in its polynomial dependence on
the raw simulation budget, up to logarithmic factors.

\paragraph{Mat\'ern kernels.}
For a Mat\'ern kernel with $\nu>d/2$,
the information gain bound is $g_T
    =
    O\left(
        T^{\frac{d}{2\nu+d}}
        (\log T)^{\frac{2\nu}{2\nu+d}}
    \right)$.  
Substitution into
\eqref{eq:general_kernel_rate_after_replication} yields
\begin{equation*}
\label{eq:matern_kernel_rate}
    f(\hat x_T)-f^\star
    =
    \widetilde O\left(
        T^{-\frac{2\nu-d}{2(2\nu+d)}}
    \right).
\end{equation*}

As mentioned above, related bounds are available under known homoscedastic noise (no replication is required), so comparing our result with them helps understand the effect of learning heteroscedastic noise through replications. The comparison shows that the price of learning unknown heteroscedastic noise enters through logarithmic factors. For instance, with known homoscedastic noise, the same
information-gain argument gives
\[
    f(\hat x_T)-f^\star
    =
    O\left(
        \frac{
            (\log T)^{d+1}
        }{
            \sqrt T
        }
    \right),
\]
for SE kernels. Thus, under the present analysis, estimating the unknown local variances
changes the SE bound by the logarithmic factor $\sqrt{
        \frac{\log T}{\log\log T}
    }$,
but does not change its polynomial dependence on the raw simulation budget. Similar results can be found for the Mat\'ern kernels.

\section{Numerical Experiments}
\label{sec:numerical}

We conduct two complementary numerical studies. The first uses generated objectives that lie exactly in the assumed RKHS, with known kernel hyperparameters, to examine the finite-budget convergence behavior and the three design choices identified by the regret analysis. The second compares our regret certificate guided EI (CertEI) with SKO, EQI, and TSSO, first on the same RKHS objectives and then on four standard test functions. The RKHS study isolates the mechanisms analyzed by the theory, whereas the standard test functions examine the robustness of the resulting algorithm beyond the matched-kernel setting. 

\subsection{Controlled RKHS Experiments}
\label{subsec:rkhs_experiments}
We first use controlled RKHS objectives to examine two questions: whether the finite-budget behavior of CertEI is consistent with the upper-bound rates derived in Section~\ref{subsec:kernel_rates}, and how its three design choices affect its empirical performance.

\subsubsection{Controlled setting.}
For each dimension $d\in\{2,4\}$ and each of the Mat\'ern-$5/2$ and SE kernels, we construct one objective function $f(x)$ from the corresponding RKHS on decision space $[0,1]^d$ in the following form:
\[
    f(x)=\sum_{j=1}^{100}w_j k(x,z_j).
\]
A simulation replication at $x$ follows $Y(x)=f(x)+\tau(x)Z$, where $Z\sim N(0,1)$ independently and $\tau(x)$ varies with the objective value. Specifically, following the construction in \cite{jalali2017comparison}, we let the noise standard deviation increase linearly with the objective value. Denote $R_f=f_{\max}-f^\star$ as the response range and define $u(x)
    =
    \operatorname{clip}
    \left(
        \frac{f(x)-f^\star}{R_f},\,0,\,1
    \right)$.
We then set $\tau(x)=R_f\bigl[0.75+2.25u(x)\bigr]$. Consequently, the noise standard deviation equals $0.75R_f$ at a global minimizer and increases to $3R_f$ at the largest objective value. 
The response GP uses the kernel and hyperparameters employed to construct the objective, and these hyperparameters are held fixed throughout each run. Following related theoretical studies (see, e.g., \citealt{wang2025gaussian}), the known hyperparameters are used to align the experiment closely with the conditions of theoretical analysis and isolate the convergence behavior of the algorithm. Thus, the response-GP hyperparameters are known, but the input-dependent simulation variance must still be learned from replications. 
For each combination of test function and budget, we perform 100 independent runs.
Performance is measured by the simple regret $\mathrm{GAP}(T)=f(\widehat x_T)-f^\star$. The complete objective construction and related experimental details are reported in Appendix O.

\subsubsection{Empirical evidence of the regret bound}
\begin{figure}[h]
    \centering
    \captionsetup[subfigure]{font=scriptsize}
    \begin{subfigure}{0.48\textwidth}
        \centering
        \includegraphics[width=\linewidth]
        {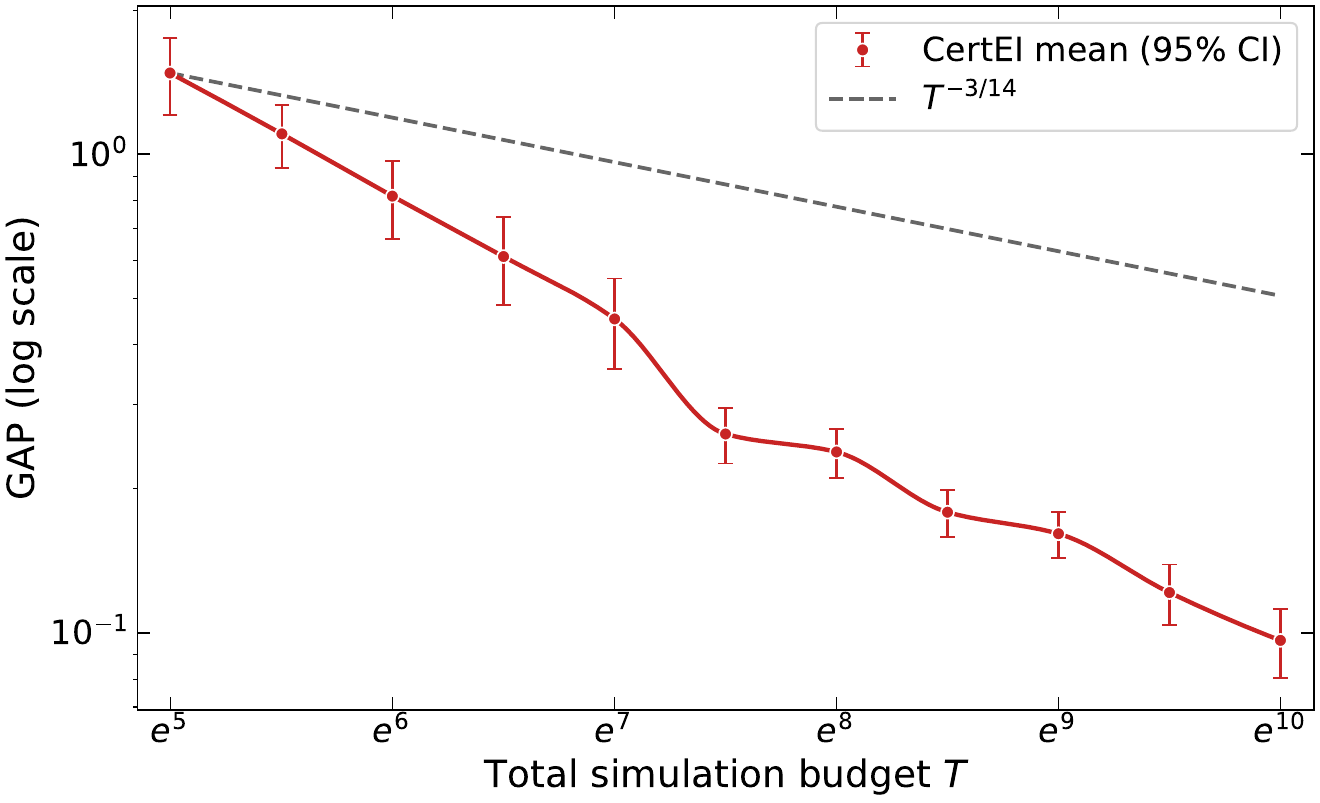}
        \caption{2d Mat\'ern-$5/2$}
        \label{fig:rate-m52-2d}
    \end{subfigure}
    \hfill
    \begin{subfigure}{0.48\textwidth}
        \centering
        \includegraphics[width=\linewidth]
        {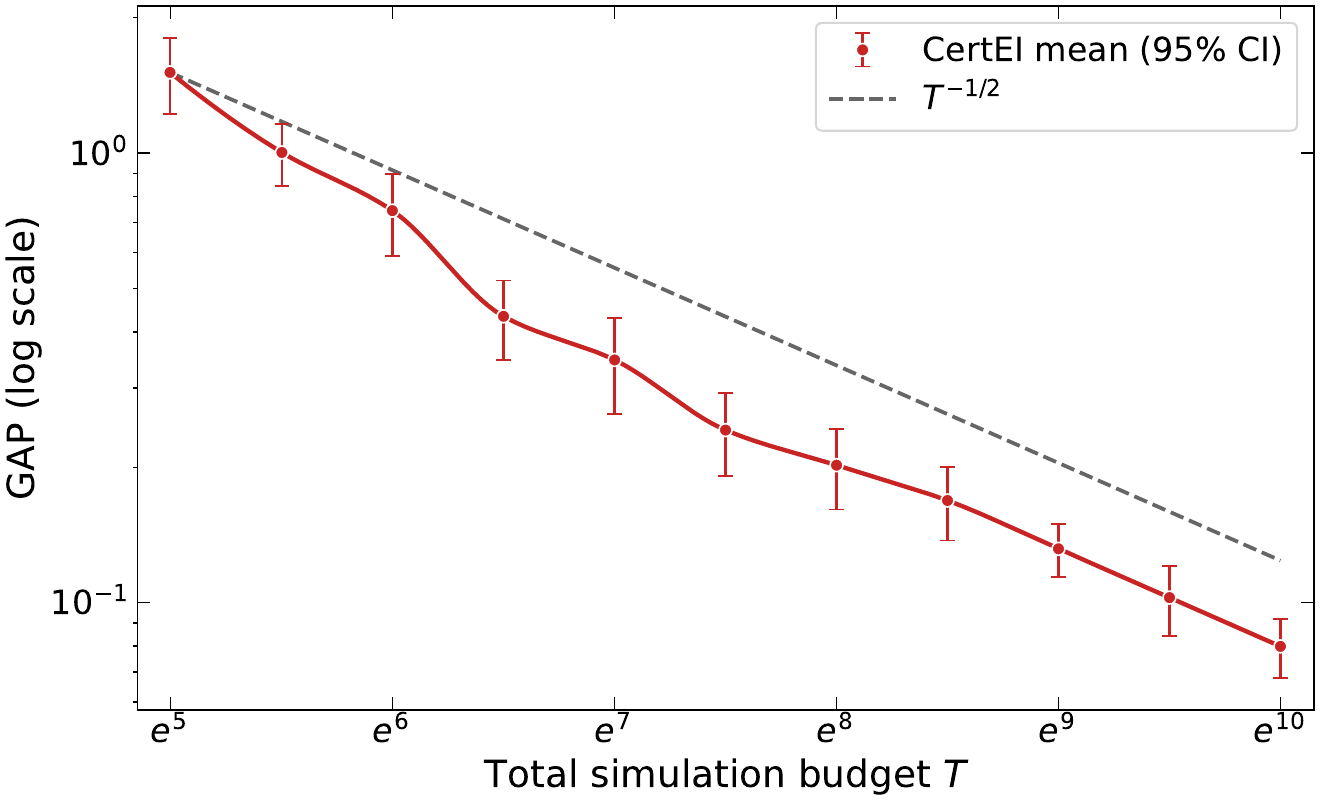}
        \caption{2d squared exponential}
        \label{fig:rate-se-2d}
    \end{subfigure}

    \medskip

    \begin{subfigure}{0.48\textwidth}
        \centering
        \includegraphics[width=\linewidth]
        {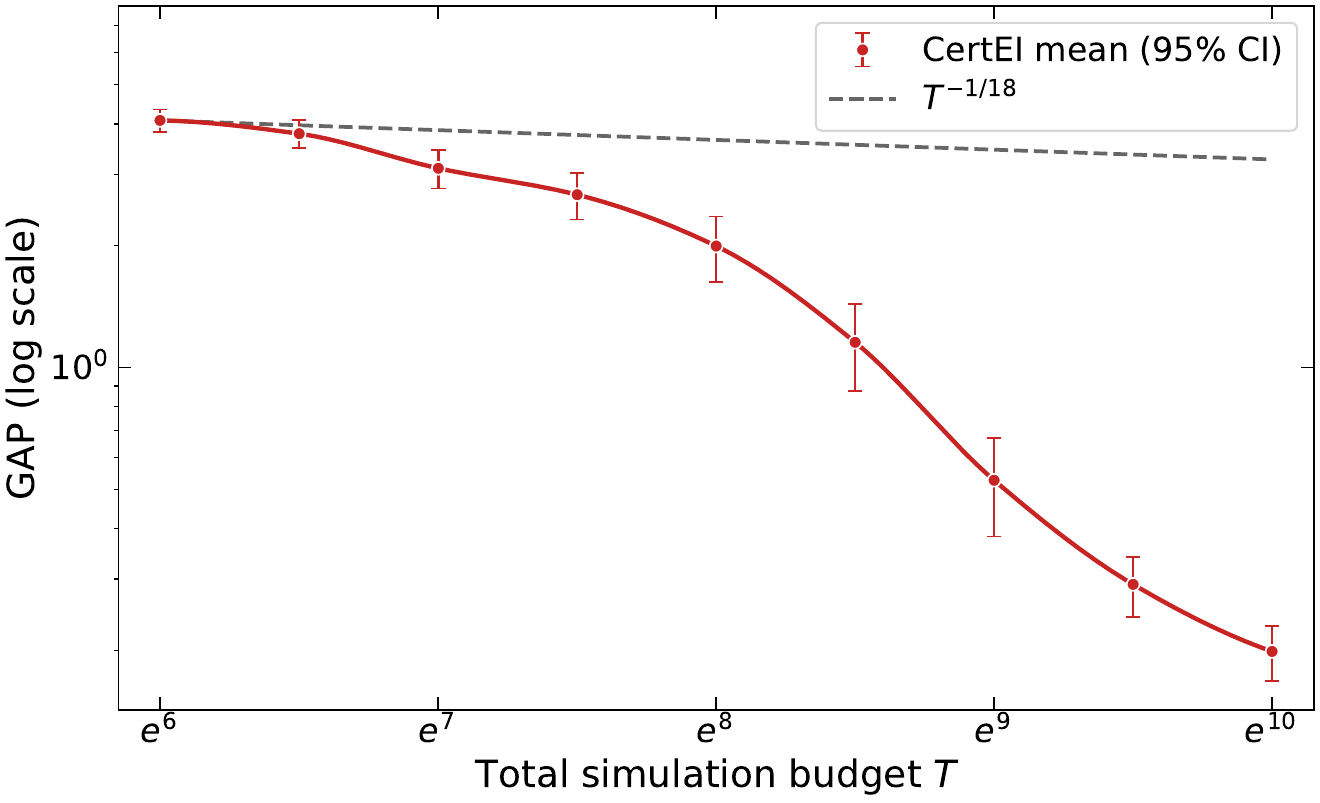}
        \caption{4d Mat\'ern-$5/2$}
        \label{fig:rate-m52-4d}
    \end{subfigure}
    \hfill
    \begin{subfigure}{0.48\textwidth}
        \centering
        \includegraphics[width=\linewidth]
        {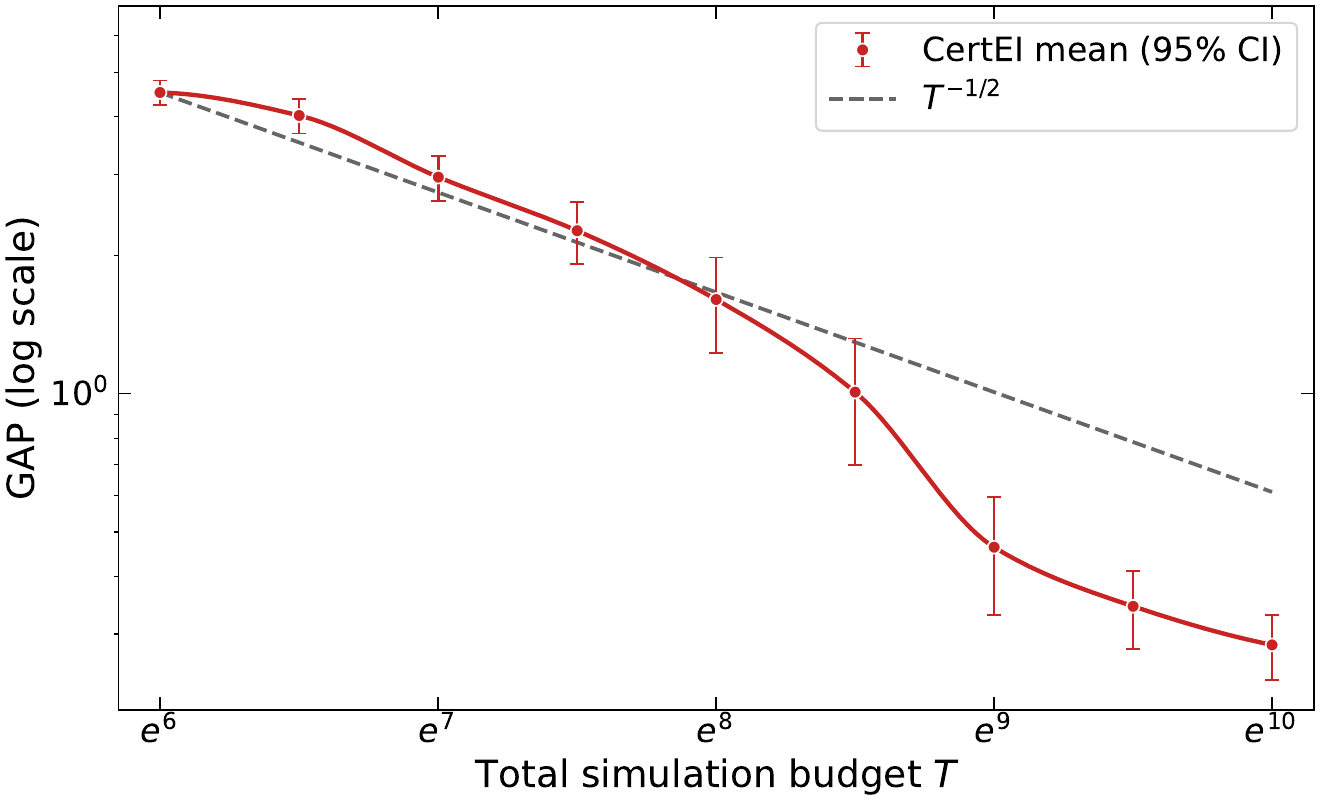}
        \caption{4d squared exponential}
        \label{fig:rate-se-4d}
    \end{subfigure}

    \caption{Average simple regret of CertEI on the RKHS objectives.}
    \label{fig:rkhs-rate}
\end{figure}

Figure~\ref{fig:rkhs-rate} compares the average simple regret of CertEI over 100 macro-replications with the polynomial parts of the corresponding theoretical rates. 
The dashed lines represent the regret bound, with orders of $T^{-3/14}$ and $T^{-1/2}$ in two dimensions, and $T^{-1/18}$ and $T^{-1/2}$ in four dimensions, for the Mat\'ern-$5/2$ and SE kernels, respectively. Each dashed line is normalized to agree with CertEI at the first displayed budget. We can observe that the mean simple regret decreases throughout the displayed budget range in all four settings. In addition, the empirical values remain below the corresponding reference curves. For example, on the four-dimensional Mat\'ern-$5/2$ objective, the mean GAP decreases from $2.5861$ at $T=e^8$ to $0.1017$ at $T=e^{10}$. This finite-budget behavior is consistent with the theoretical orders. As these orders are upper bounds, empirical regret often decreases faster than the stated theoretical order, as observed both in our experiments and in related studies \citep{wang2025gaussian}.

\subsubsection{Ablation of the design choices}
\begin{figure}
    \centering
    \captionsetup[subfigure]{font=scriptsize}
    \begin{subfigure}{0.32\textwidth}
        \centering
        \includegraphics[width=\linewidth]
        {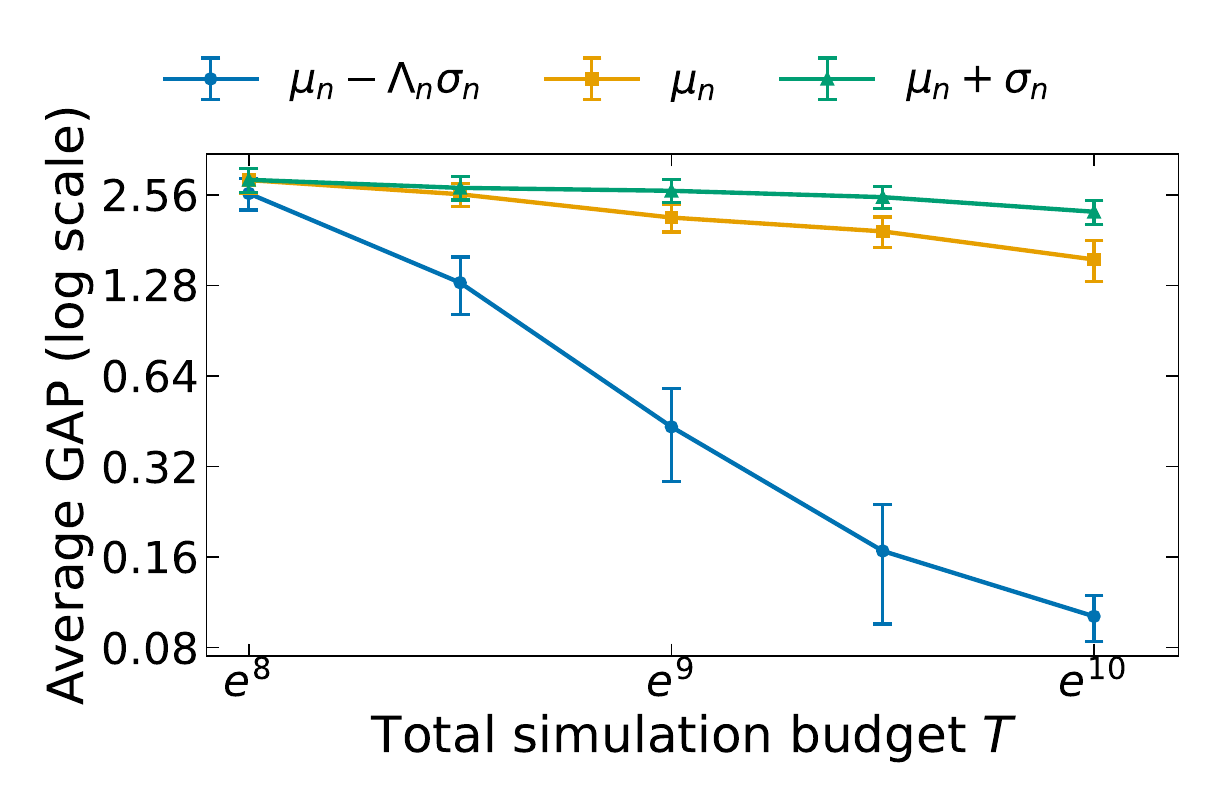}
        \caption{EI incumbent}
        \label{fig:ablation-incumbent}
    \end{subfigure}
    \hfill
    \begin{subfigure}{0.32\textwidth}
        \centering
        \includegraphics[width=\linewidth]
        {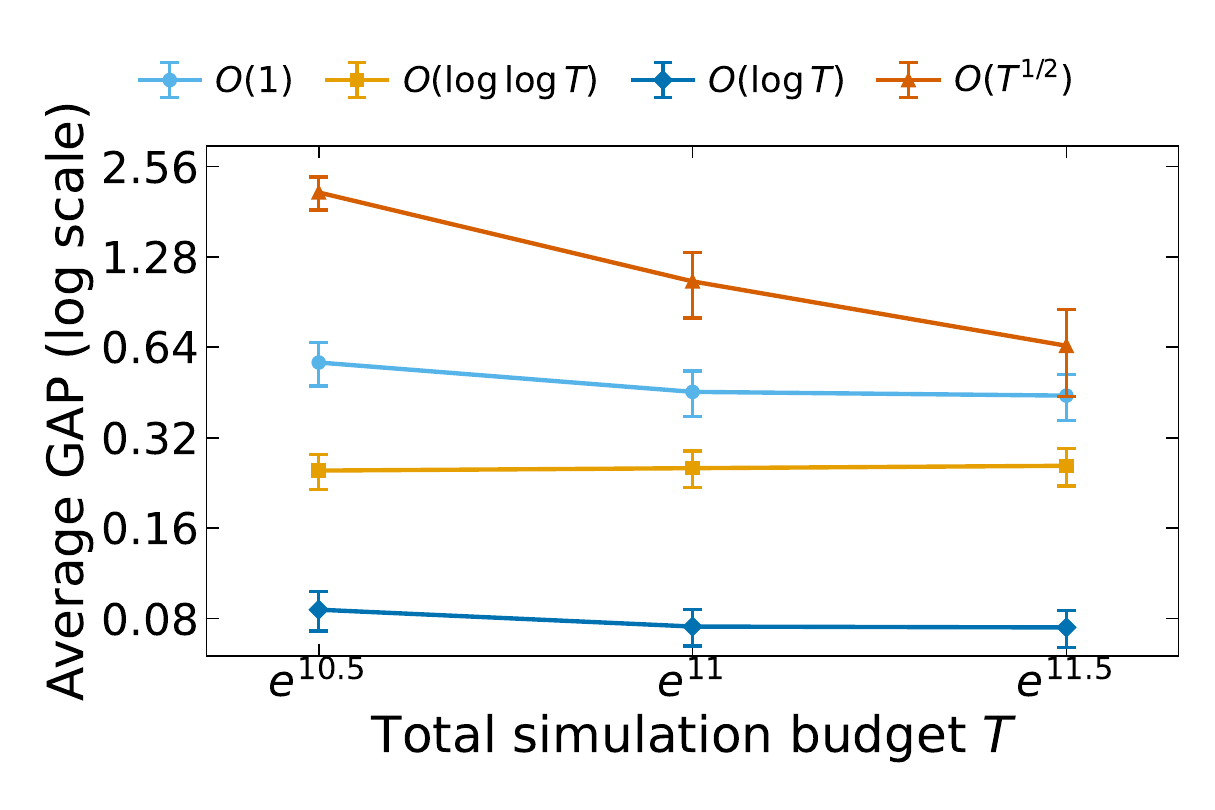}
        \caption{Replication order}
        \label{fig:ablation-replication}
    \end{subfigure}
    \hfill
    \begin{subfigure}{0.32\textwidth}
        \centering
        \includegraphics[width=\linewidth]
        {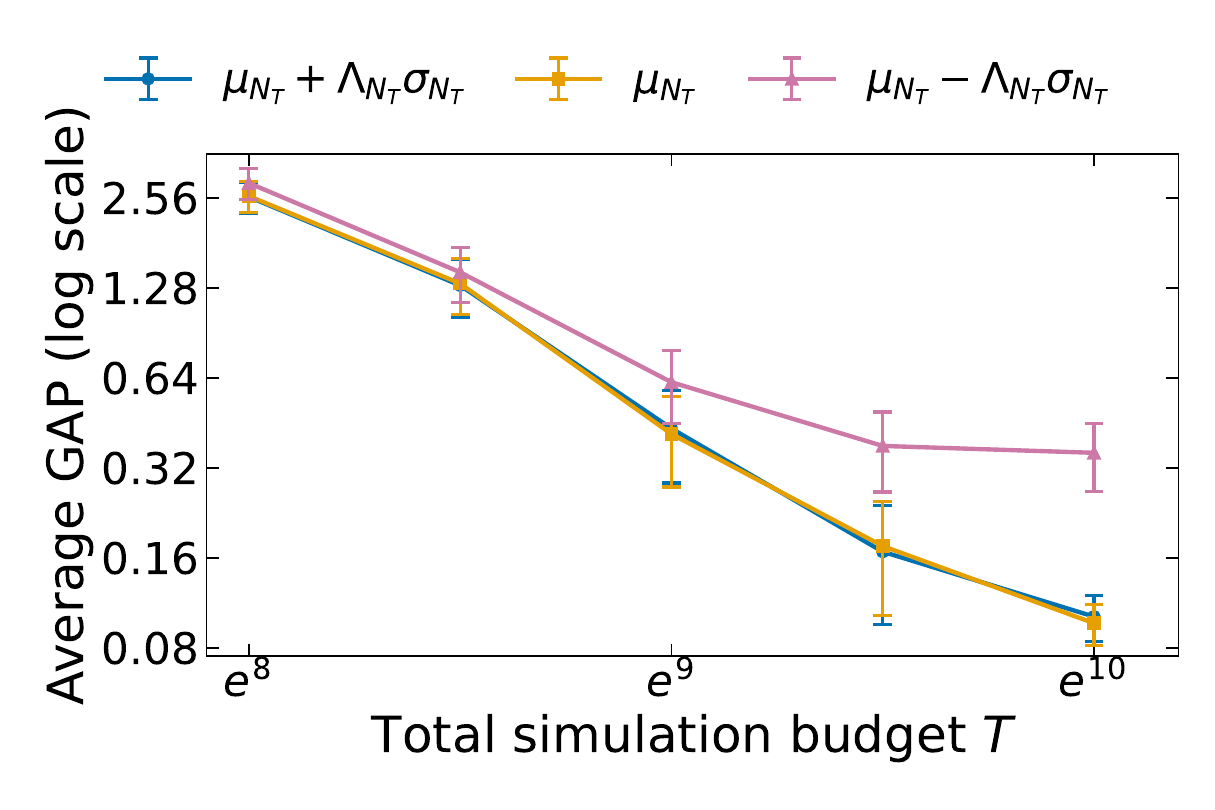}
        \caption{Final recommendation}
        \label{fig:ablation-final}
    \end{subfigure}

    \caption{Controlled ablations on the four-dimensional
    Mat\'ern-$5/2$ objective. }
    \label{fig:ablation-m52-4d}
\end{figure}

We next conduct ablation experiments to evaluate the effect of each design choice. To isolate the effect, we consider three sets of experiments, varying only one design choice at a time while holding all other experimental details fixed.
For the EI incumbent, we compare $\min_x\{\mu_n(x)-\Lambda_n\sigma_n(x)\}$, $\min_x\mu_n(x)$, and $\min_x\{\mu_n(x)+\sigma_n(x)\}$.
For the replication size, we compare the orders
$O(1)$, $O(\log\log T)$, $O(\log T)$, and $O(T^{1/2})$.
For the final recommendation, we apply the three terminal scores $\mu_{N_T}(x)-\Lambda_{N_T}\sigma_{N_T}(x)$, and $\mu_{N_T}(x)$, $\mu_{N_T}(x)+\Lambda_{N_T}\sigma_{N_T}(x)$
to the set of sampled designs. The comparison levels are chosen to reflect both common practice and
systematic deviations from our theoretically motivated choices. The
alternative incumbent and final-recommendation rules are commonly used
in the literature. For replication, $O(1)$ corresponds to the fixed
equal-replication rule used in prior studies, while $O(\log\log T)$
and $O(T^{1/2})$ are smaller and larger, respectively, than our
$O(\log T)$ choice. They represent different tradeoffs between
more decision points and more precise estimates at each
point. To keep the presentation compact,
Figure~\ref{fig:ablation-m52-4d} reports only the average GAP of the
four-dimensional Mat\'ern-$5/2$ instance. Results for the
two-dimensional Mat\'ern-$5/2$, two-dimensional
SE, and four-dimensional SE
objectives are reported in Appendix O.

In Figure \ref{fig:ablation-m52-4d} (a), the calibrated lower-confidence incumbent consistently produces the smallest average GAP. 
Under small simulation budgets, all three rules perform similarly. As the budget increases, however, the other two rules converge more slowly, with the upper-quantile incumbent exhibiting the slowest convergence. This indicates that they do not explore the decision space sufficiently to locate the region containing the global minimizer, which is consistent with our theoretical analysis.
In Figure \ref{fig:ablation-m52-4d} (b), the calibrated logarithmic replication schedule attains the smallest mean GAP. 
The two small replications choices preserve more search opportunities but provide noisier local estimates, while the square-root choice can leave too few adaptive decisions. Similar to our theoretical analysis, they all attain lower convergence rates.
In Figure \ref{fig:ablation-m52-4d} (c), the proposed final rule and the posterior-mean rule are nearly indistinguishable on this test, both much better than the lower quantile criterion. Thus, the proposed final rule avoids the deterioration caused by
rewarding residual uncertainty. We also report an additional
four-dimensional Mat\'ern-$5/2$ experiment showing the advantage of
our rule over selecting the smallest posterior mean in Appendix O.4.1. Under larger noise, the same budget leaves greater residual uncertainty in the
candidate solutions, making the difference between the two rules more
pronounced. The results show that explicitly penalizing this
uncertainty can produce a more reliable final recommendation.

The preceding one-factor-at-a-time ablations isolate the effect of each design choice. In Appendix O.5, we present a fully crossed experiment over all levels of the three design choices considered in the preceding ablations. This yields $3\times4\times3=36$ algorithmic configurations, which we evaluate on the RKHS objectives. As a complement to the one-factor-at-a-time ablations, this experiment examines whether the selected design choices remain effective when used jointly. The results are consistent with the individual ablations. Across all tested settings, the proposed configuration consistently ranks among the top two of the 36 configurations.

\subsection{Comparison with Existing Algorithms}
\label{subsec:algorithm-comparison}

We next compare CertEI with commonly used EI algorithms for stochastic simulations, namely SKO\footnote{The original SKO was designed for functions with homoscedastic noises. We used its adapted version from \cite{jalali2017comparison}.}, EQI, and TSSO. We first compare them under the RKHS objectives in the last subsection with an ideal setting of known GP hyperparameter. We then use four synthetic test functions to evaluate the algorithms when the GP hyperparameters must be learned during optimization. Each test is evaluated using 100 macro-replications.

The four synthetic functions are six-hump Camelback, rescaled Branin, Hills, and Hartmann-6. The Camelback and Branin functions are low-dimensional multimodal problems with several well-separated minima. The Hills function provides a more densely multimodal landscape with many competing local optima, while Hartmann-6 examines performance on a higher-dimensional multimodal problem. Together, these functions cover optimization landscapes with different dimensions and degrees of multimodality. As in the preceding RKHS experiments, we add Gaussian noise to these functions following the construction of \cite{jalali2017comparison}. The complete settings of these functions are provided in Appendix O.6.

Figure \ref {fig:comparison-analytical} reports the comparison results. 
\begin{figure}[htbp]
    \centering
    \captionsetup[subfigure]{font=scriptsize}
    \begin{subfigure}[t]{0.31\textwidth}
        \centering
        \includegraphics[width=\linewidth]
        {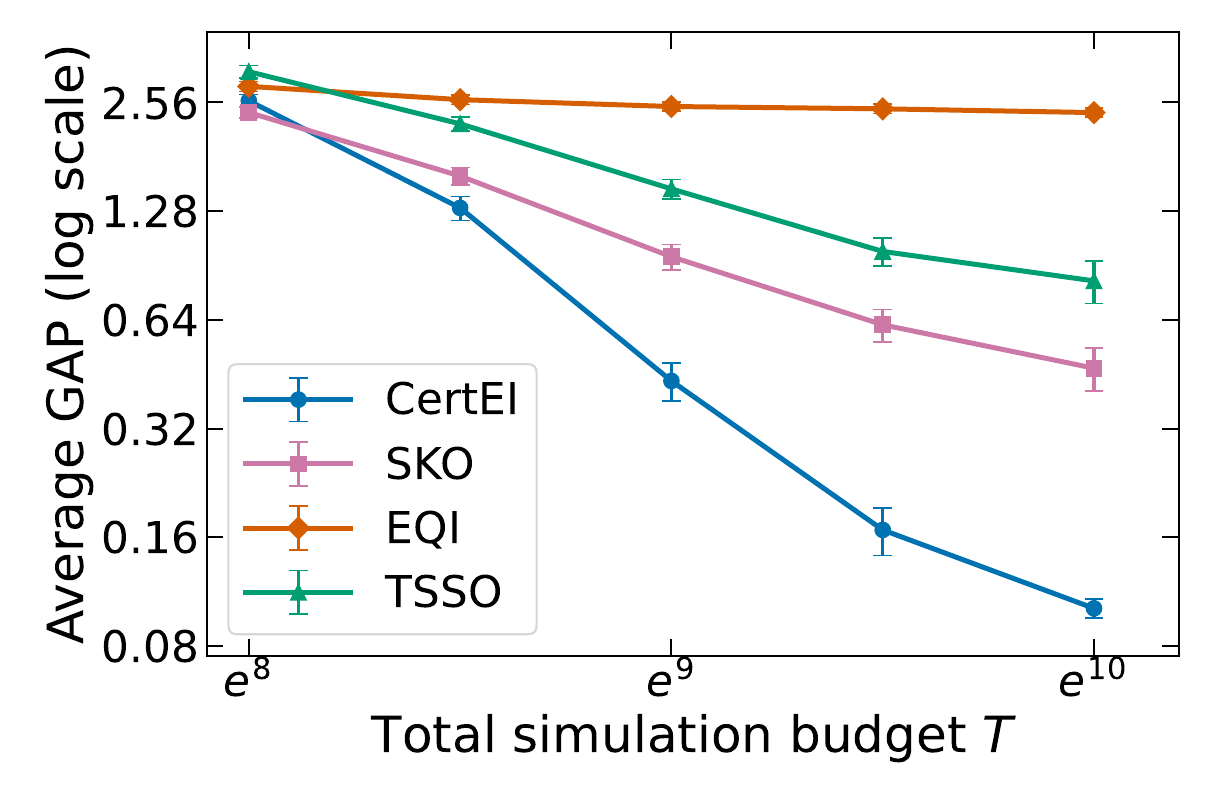}
        \caption{4d Mat\'ern-$5/2$ objective}
    \end{subfigure}
    \hfill
    \begin{subfigure}[t]{0.31\textwidth}
        \centering
        \includegraphics[width=\linewidth]
        {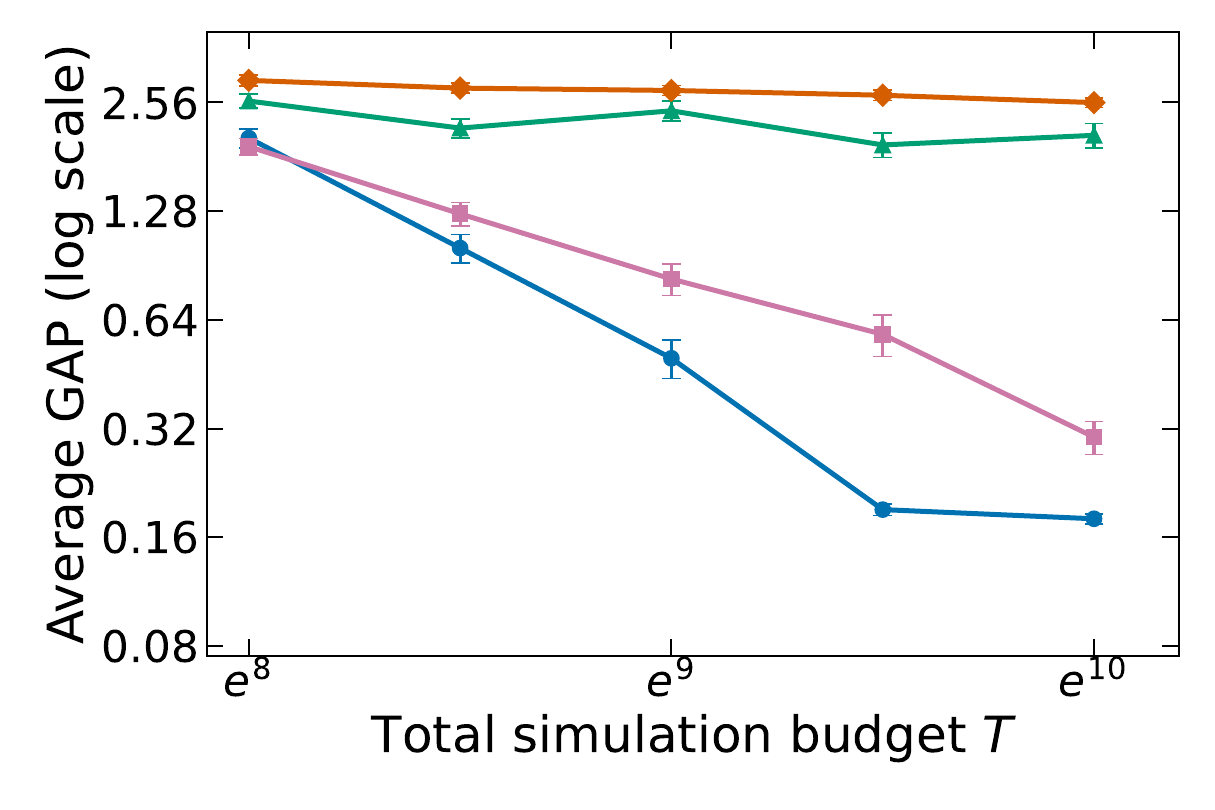}
        \caption{4d SE objective}
    \end{subfigure}
    \hfill
    \begin{subfigure}[t]{0.31\textwidth}
        \centering
        \includegraphics[width=\linewidth]
        {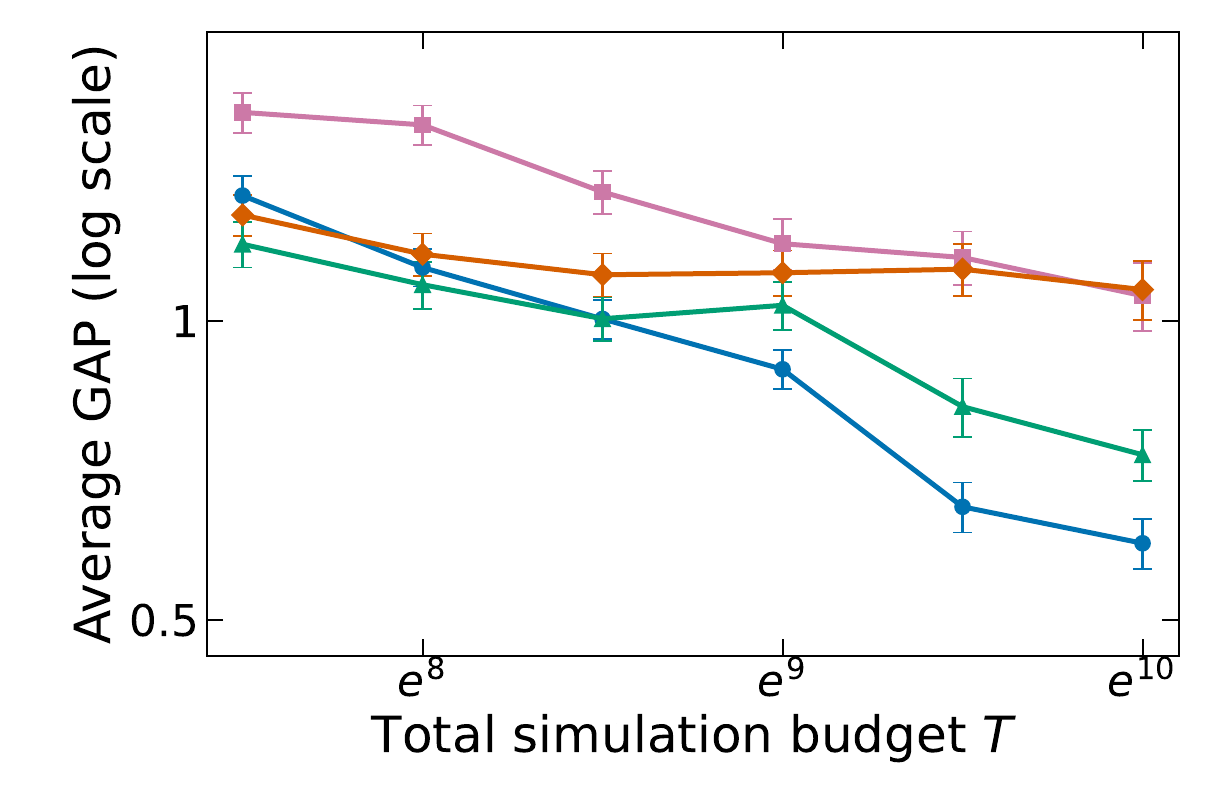}
        \caption{2d six-hump Camelback}
    \end{subfigure}

     \par\medskip
     
    \begin{subfigure}[t]{0.31\textwidth}
        \centering
        \includegraphics[width=\linewidth]
        {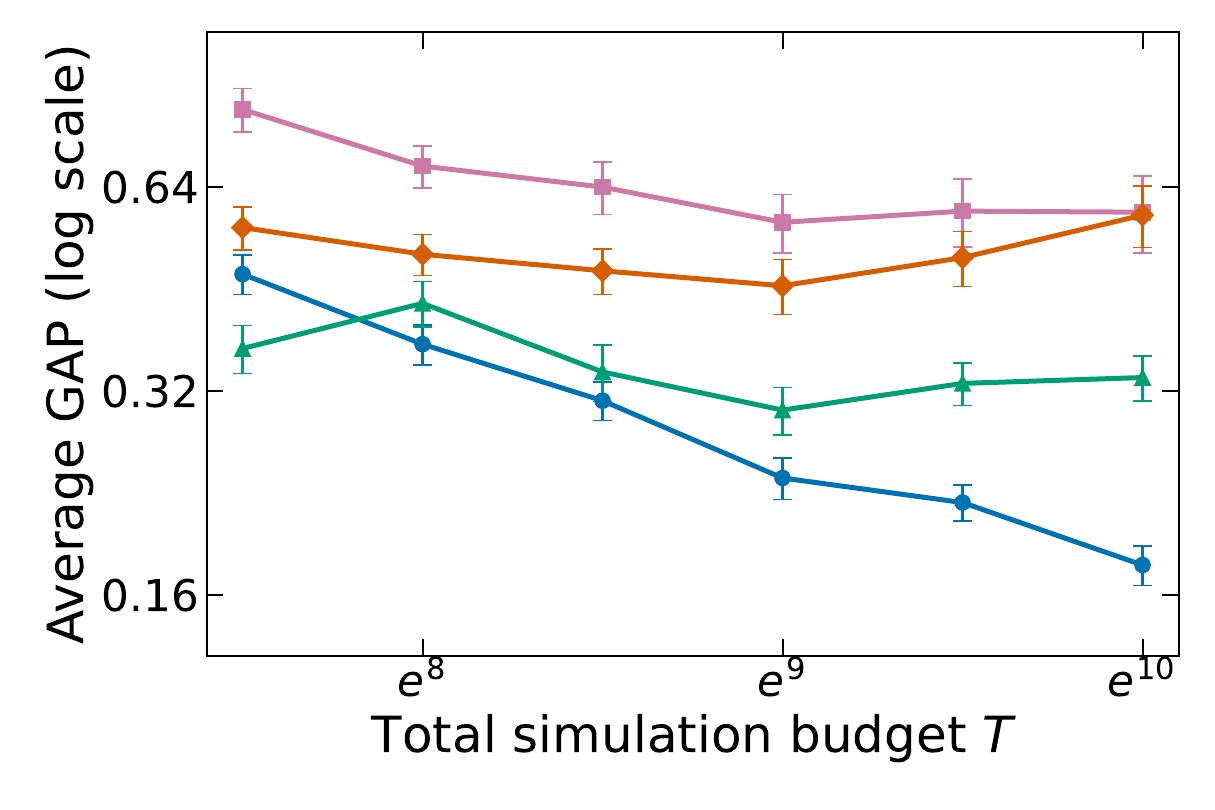}
        \caption{2d rescaled Branin}
    \end{subfigure}
    \hfill
    \begin{subfigure}[t]{0.31\textwidth}
        \centering
        \includegraphics[width=\linewidth]
        {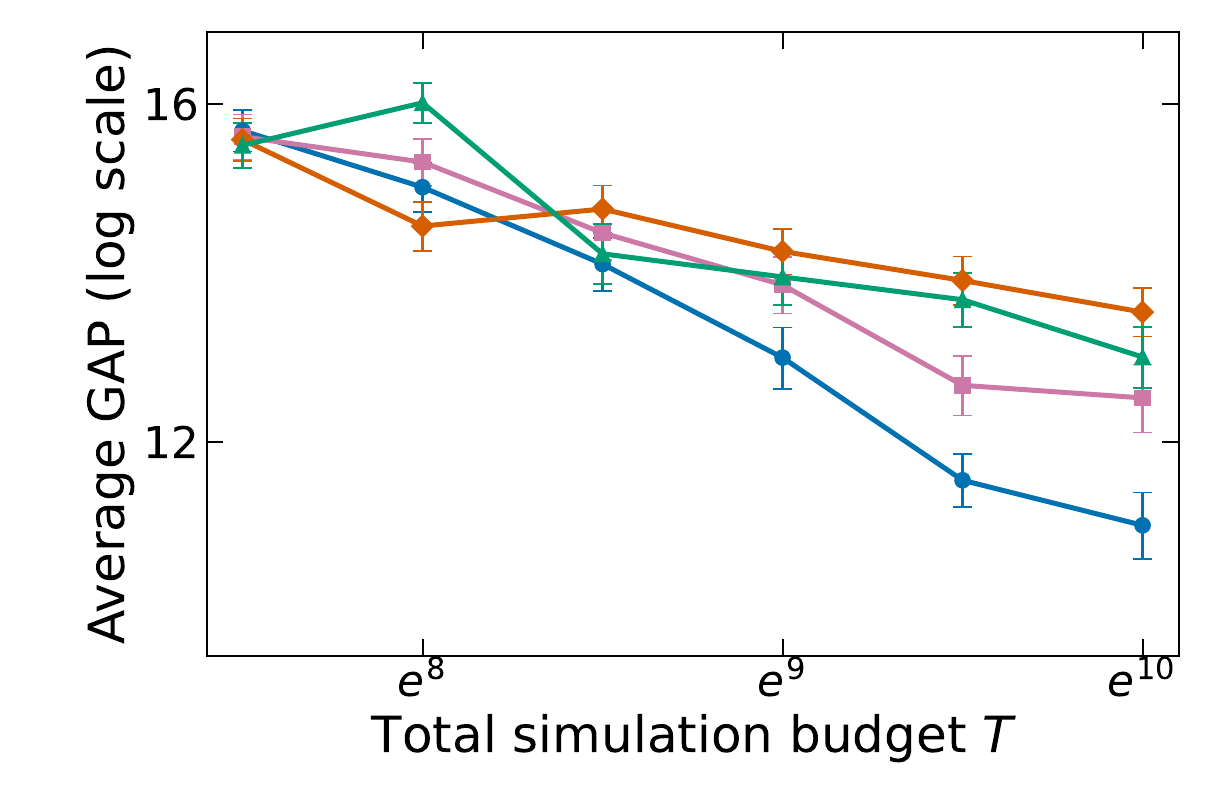}
        \caption{2d Hills}
    \end{subfigure}
    \hfill
    \begin{subfigure}[t]{0.31\textwidth}
        \centering
        \includegraphics[width=\linewidth]
        {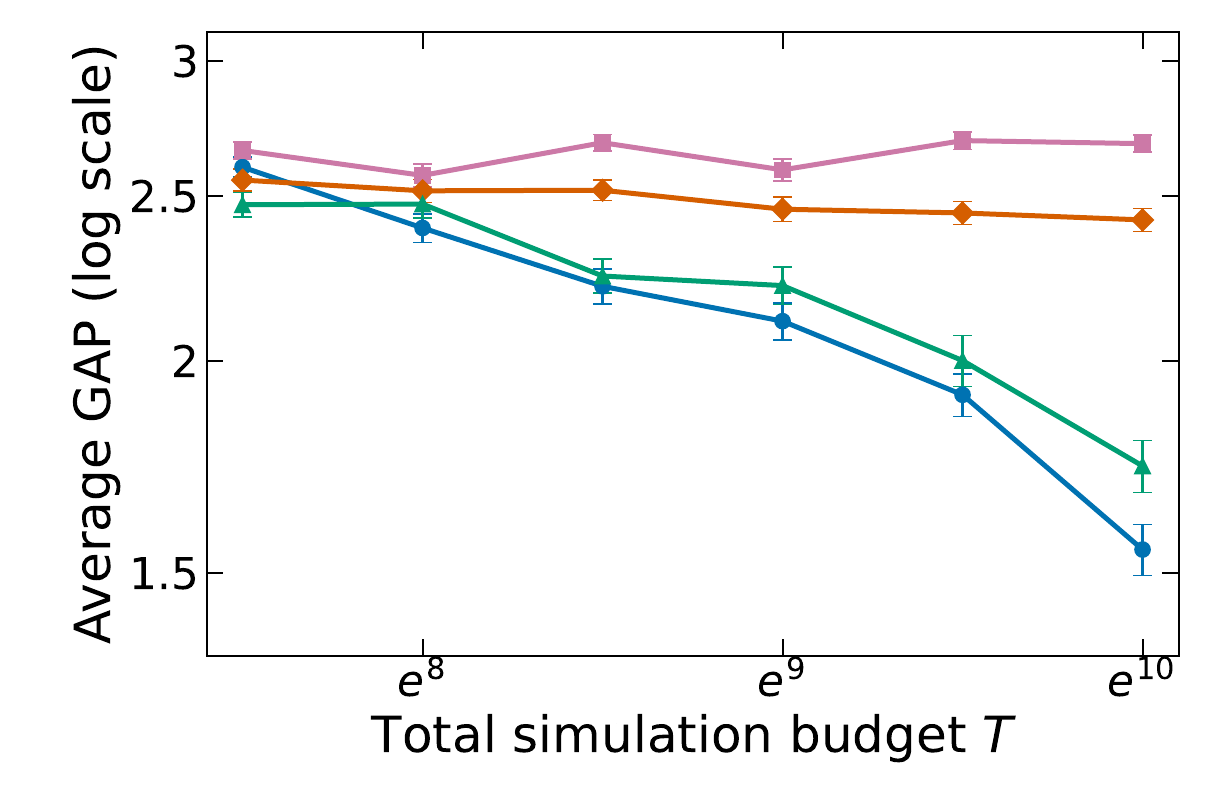}
        \caption{6d Hartmann-6}
    \end{subfigure}

    \caption{Average GAP of the four algorithms}
    \label{fig:comparison-analytical}
\end{figure}

We can observe that at the smaller budgets, the relative performance of the four algorithms varies across the six problems, and CertEI is not always the best-performing method. As the budget increases, however, CertEI exhibits the most consistent reduction in average GAP. From approximately $T=e^{8.5}$ through $T=e^{10}$, it attains the smallest average GAP on all six problems. In particular, CertEI continues to improve on the four analytical functions after some competing methods begin to flatten, and it achieves the smallest average GAP on every problem at $T=e^{10}$. The advantage is most pronounced on Camelback and rescaled Branin, while the Hills problem remains difficult for all four methods and exhibits substantial variation across macro-replications. CertEI also maintains its improvement on the higher-dimensional Mat\'ern-$5/2$, SE, and Hartmann-6 problems, indicating that the observed performance is not limited to low-dimensional test functions. 

Overall, these experiments provide complementary evidence for both the convergence behavior and the design choices underlying CertEI. On the fixed RKHS objectives, the empirical GAP decreases consistently with, and often faster than, the rates suggested by the theoretical upper bounds. The one-factor-at-a-time and fully crossed ablations further show that the proposed incumbent, replication order, and final recommendation form a robust combination. In comparison with existing EI algorithms, CertEI exhibits the most consistent improvement as the budget increases and attains the smallest average GAP on all six test problems. Taken together, the results support the proposed algorithmic design and the competitive finite-budget performance of CertEI.

\section{Summary}
\label{sec:summary}
This paper studies how to design an EI algorithm for continuous simulation optimization when the simulation noise is heteroscedastic and unknown. Designing such an algorithm requires specifying three basic components: the incumbent, the replication size, and the final recommendation. Existing theoretical and empirical studies generally either assume that the heteroscedastic noise variances are known or determine these three components separately using heuristic rules. In contrast, we take a theory-guided approach: we derive a high-probability simple-regret bound and use it as a common criterion to systematically determine all three design choices. 
This bound is built on a finite-sample confidence inequality for the SK predictor when the noise variances are estimated from replications. Within the rule classes considered, to tighten the regret bound, the resulting method uses a lower-quantile incumbent to preserve exploration, $\Theta(\log T)$ replications to balance variance estimation and spatial search, and a conservative upper-quantile recommendation over sampled designs. The resulting simple regret is $\widetilde O(T^{-1/2})$ for SE kernels, matching the polynomial dependence in the minimax lower bound, while the Mat\'ern rate matches its known-homoscedastic confidence-bound counterpart up to logarithmic factors. The numerical results and ablation studies are consistent with the theoretical convergence behavior and support the three design choices. Comparisons with existing common EI algorithms further show the competitive empirical performance of our calibrated design choices.


\appendix
 \section{Proof of Lemma \ref{lem:plugin_variance}}

 \begin{proof}{Proof of Lemma \ref{lem:plugin_variance}.}
Conditional on the history $\mathcal H_{q-1}$, $Z_q
    :=
    \frac{\nu_TS_q^2}{v(x_q)}
    \sim
    \chi^2_{\nu_T}$,
where
    $\nu_T=b_T-1$.
By the definition of the chi-squared quantile, $\mathbb P\!\left(
        Z_q<\chi^2_{\nu_T,\alpha_T}
        \,\middle|\,
        \mathcal H_{q-1}
    \right)
    \le
    \alpha_T$ and $\mathbb P\!\left(
        Z_q>\chi^2_{\nu_T,1-\alpha_T}
        \,\middle|\,
        \mathcal H_{q-1}
    \right)
    \le
    \alpha_T$. 
In addition, according to the definition of $\kappa_T$ and $h_T$, we have $1/\kappa_T\leq \chi^2_{\nu_T,\alpha_T}/\nu_T$ and $h_T\geq \chi^2_{\nu_T,1-\alpha_T}/\nu_T$. Thus, we have 
$$ \mathbb P\!\left(  S_q^2<\frac{v(x_q)}{\kappa_T}
    \right) \le  \mathbb P\!\left(  S_q^2<\frac{v(x_q)\chi^2_{\nu_T,\alpha_T}}{\nu_T}
    \right)
    \le
    \alpha_T, \  \mathbb P\!\left(  S_q^2>v(x_q)h_T
    \right) \le  \mathbb P\!\left(  S_q^2>\frac{v(x_q)\chi^2_{\nu_T,1-\alpha_T}}{\nu_T}
    \right)
    \le
    \alpha_T.$$
Therefore, except on an event with conditional probability at most
$2\alpha_T$, we have $\frac{v(x_q)}{\kappa_T}\le S_q^2
    \le h_Tv(x_q)$.
Taking the union bound over these events on all $N_T$ batches gives, with probability at least $1-\delta/2$,
\[
    \frac{v(x_q)}{\kappa_T}
    \le S_q^2\le h_Tv(x_q),
    \qquad \forall q\le N_T.
\]
On this event, $v(x_q)\le\kappa_TS_q^2
    \le\kappa_T\widetilde v_q$,
and
\[
    v_0\le\widetilde v_q
    =\max\{v_0,S_q^2\}
    \le\max\{v_0,h_TV_\star\}
    \le h_T\max\{v_0,V_\star\}
    =W_T.
\]
Dividing by $b_T$ proves the last ineuqality in Lemma \ref{lem:plugin_variance}.
\end{proof}

\section{Proof of Lemma \ref{lem:plugin_correction_order}}
\begin{proof}{Proof of Lemma \ref{lem:plugin_correction_order}.}
For a sequence of candidate replication numbers $b=b_T$, write $\nu_T
    :=
    b_T-1$, and
    $L_T
    :=
    \log\frac{4N_T}{\delta}$.
Then $\alpha_T
    =
    e^{-L_T}$.
We first prove the three statements for $\kappa_T$ separately.

\paragraph{Case 1: $b_T=o(\log T)$.}

We first establish a lower bound on the lower tail of a chi-squared random
variable. Let $Q_\nu
    =
    \sum_{j=1}^{\nu}Z_j^2
    \sim
    \chi_\nu^2$,
where $Z_1,\ldots,Z_\nu$ are independent standard normal random variables.
There exists a constant $c_0>0$ such that
\begin{equation}
\label{eq:gaussian_small_ball}
    \mathbb P(|Z_1|\le a)
    \ge
    c_0 a,
    \qquad
    0<a\le1.
\end{equation}
We can select this $c_0$ as follows. As the standard normal density is continuous and strictly positive on
$[-1,1]$, we have
\[
\begin{aligned}
    \mathbb P(|Z_1|\le a)
    =
    \int_{-a}^{a}\phi(z)\,dz
    \ge
    2a
    \min_{|z|\le1}\phi(z),
\end{aligned}
\]
and \eqref{eq:gaussian_small_ball} follows by taking $c_0
    :=
    2\phi(1)>0$.

If $0<a\le1$, then $\{|Z_j|\le a,\ j=1,\ldots,\nu\}
    \subseteq
    \{Q_\nu\le\nu a^2\}$.
Hence, by independence,
\begin{equation}
\label{eq:chi_square_small_ball}
\begin{aligned}
    \mathbb P(Q_\nu\le\nu a^2)
    \ge
    \prod_{j=1}^{\nu}
    \mathbb P(|Z_j|\le a)
    \ge
    (c_0a)^\nu.
\end{aligned}
\end{equation}
Apply this result with $\nu=\nu_T$ and define $a_T
    :=
    c_0^{-1}
    \exp\left(
        -\frac{L_T}{\nu_T}
    \right)$.
Since $b_T=o(\log T)$ and $\delta$ is fixed, $N_T
    =
    \frac{T}{b_T}$, and 
    $L_T
    =
    \log\frac{4N_T}{\delta}
    =
    \Theta(\log T)$.
Moreover, $1
    \le
    \nu_T
    =
    b_T-1
    =
    o(\log T)$,
and hence, $\frac{L_T}{\nu_T}
    \longrightarrow
    \infty$.
Consequently, $a_T\le1$ for all sufficiently large $T$.

Substitution of $a_T$ into
\eqref{eq:chi_square_small_ball} gives $\mathbb P\left(
        Q_{\nu_T}
        \le
        \nu_Ta_T^2
    \right)
    \ge
    (c_0a_T)^{\nu_T}
    =
    \exp(-L_T)
    =
    \alpha_T$.
By the definition of the lower $\alpha_T$-quantile, $\chi^2_{\nu_T,\alpha_T}
    \le
    \nu_Ta_T^2$.
Therefore,
\begin{equation}
\label{eq:kappa_small_b_lower}
\begin{aligned}
    \kappa_T
    \ge
    a_T^{-2}
    =
    c_0^2
    \exp\left(
        \frac{2L_T}{\nu_T}
    \right).
\end{aligned}
\end{equation}
Since $L_T/\nu_T\to\infty$, $\kappa_T\longrightarrow\infty$.

\paragraph{Case 2: $b_T=\Theta(\log T)$.}

In this case, $N_T
    =
    \Theta\left(
        \frac{T}{\log T}
    \right)$,
so $L_T
    =
    \Theta(\log T)$.
Since $\nu_T
    =
    b_T-1
    =
    \Theta(\log T)$,
there exists a constant $c_1>0$ such that $\frac{\nu_T}{L_T}
    \ge
    c_1$
for all sufficiently large $T$.

We use a standard Chernoff bound for the lower tail of a chi-squared random
variable. Let $Q_\nu\sim\chi_\nu^2$.
For any $s\in(0,1)$, according to \cite{ghosh2021exponential}, we have 
\begin{equation}
\label{eq:chi_square_chernoff}
    \mathbb P(Q_\nu\le s\nu)
    \le
    \exp\left\{
        -\nu I(s)
    \right\},
\end{equation}
where $I(s)
    :=
    \frac{s-1-\log s}{2}$.
Because $I(s)\longrightarrow\infty$ as $s\downarrow0$,
we can choose a fixed $s_0\in(0,1)$ such that $c_1 I(s_0)>1$. It follows that, $\nu_T I(s_0)
    >
    L_T$
for all sufficiently large $T$. Hence,
\[
\begin{aligned}
    \mathbb P\left(
        Q_{\nu_T}
        \le
        s_0\nu_T
    \right)
    \le
    \exp\left\{
        -\nu_T I(s_0)
    \right\}
    <
    e^{-L_T}
    =
    \alpha_T.
\end{aligned}
\]
Therefore, $\chi^2_{\nu_T,\alpha_T}
    \ge
    s_0\nu_T$.
It follows that $\kappa_T
    \le
    \max\left\{
        1,\frac{1}{s_0}
    \right\}
    =
    \frac{1}{s_0}$.
Thus, $\kappa_T=O(1)$.

\paragraph{Case 3: $b_T/\log T\to\infty$.}

Since $N_T\le T$, $L_T
    =
    \log\frac{4N_T}{\delta}
    =
    O(\log T)$.
On the other hand, $\nu_T
    =
    b_T-1$
satisfies $\frac{\nu_T}{\log T}
    \longrightarrow
    \infty$.
Hence, $\frac{L_T}{\nu_T}
    \longrightarrow
    0$.
The Laurent--Massart lower-tail inequality gives, for every $x>0$,
\[
    \mathbb P\left(
        Q_{\nu_T}
        \le
        \nu_T
        -
        2\sqrt{\nu_Tx}
    \right)
    \le
    e^{-x}.
\]
Taking $x=L_T$
gives
\[
    \mathbb P\left(
        Q_{\nu_T}
        \le
        \nu_T
        -
        2\sqrt{\nu_TL_T}
    \right)
    \le
    e^{-L_T}
    =
    \alpha_T.
\]
Therefore, $\chi^2_{\nu_T,\alpha_T}
    \ge
    \nu_T
    -
    2\sqrt{\nu_TL_T}$.
For all sufficiently large $T$, the right-hand side is positive, and hence
\[
\begin{aligned}
    1
    \le
    \kappa_T
    \le
    \frac{
        \nu_T
    }{
        \nu_T
        -
        2\sqrt{\nu_TL_T}
    }
    =
    \frac{
        1
    }{
        1-
        2\sqrt{L_T/\nu_T}
    }.
\end{aligned}
\]
Since $\frac{L_T}{\nu_T}
    \longrightarrow0$,
the right-hand side converges to one. Therefore, $\kappa_T\longrightarrow1$.

\paragraph{Fixed replication number.}

Finally, suppose that $b_T=b\ge2$ is fixed, and write $\nu:=b-1$.
The chi-squared density is
\[
    f_\nu(x)
    =
    \frac{
        x^{\nu/2-1}e^{-x/2}
    }{
        2^{\nu/2}\Gamma(\nu/2)
    },
    \qquad x>0.
\]
As $x\downarrow0$, $e^{-x/2}=1+o(1)$,
and hence
\[
\begin{aligned}
    \mathbb P(Q_\nu\le x)
    =
    \int_0^x f_\nu(y)\,dy
    =
    \frac{
        x^{\nu/2}
    }{
        2^{\nu/2}
        \Gamma(\nu/2+1)
    }
    \{1+o(1)\}.
\end{aligned}
\]
Therefore, as $\alpha\downarrow0$, $\chi^2_{\nu,\alpha}
    =
    \Theta\left(
        \alpha^{2/\nu}
    \right)$.

For fixed $b$, $N_T
    =
    \left\lfloor\frac{T}{b}\right\rfloor
    =
    \Theta(T)$,
so $\alpha_T
    =
    \frac{\delta}{4N_T}
    =
    \Theta(T^{-1})$.
Consequently,
\[
\begin{aligned}
    \kappa_T
    =
    \Theta\left(
        \alpha_T^{-2/(b-1)}
    \right)
    =
    \Theta\left(
        T^{2/(b-1)}
    \right).
\end{aligned}
\]

We next establish the corresponding results for $h_T$. For notation simplicity,  define $q_T:=\chi^2_{\nu_T,1-\alpha_T}$, and $r_T:=\frac{L_T}{\nu_T}$. The Chernoff bound in \cite{ghosh2021exponential} gives, when $s>1$
\[
    \mathbb P(Q_\nu\ge s\nu)
    \le
    \exp\{-\nu I(s)\}.
\]

Suppose first that $b_T=o(\log T)$, so $r_T\to\infty$.
Fix $\varepsilon\in(0,1)$ and set
$s_T=(2+\varepsilon)r_T$.
Since $I(s)\sim s/2$ as $s\to\infty$,
\[
    \frac{\nu_T I(s_T)}{L_T}
    \longrightarrow
    1+\frac{\varepsilon}{2}>1.
\]
Thus, for all sufficiently large $T$, $s_T\nu_T=(2+\varepsilon)r_T\nu_T=(2+\varepsilon)L_T$. The Chernoff bound above gives $\mathbb P\!\left(Q_{\nu_T}\ge(2+\varepsilon)L_T\right) \le e^{-\nu_T I(s_T)}<e^{-L_T}=\alpha_T$. Hence, $q_T\le(2+\varepsilon)L_T$ for all sufficiently
large $T$. In addition, $x_T=(2-\varepsilon)L_T$, we have
\[
    \mathbb P(Q_{\nu_T}>x_T)\ge
    \mathbb P(Q_1>x_T)
    \ge
    \int_{x_T}^{x_T+1}
        \frac{e^{-u/2}}{\sqrt{2\pi u}}\,du
    \ge
    \frac{e^{-(x_T+1)/2}}
         {\sqrt{2\pi(x_T+1)}}
    >
    e^{-L_T}
    =
    \alpha_T
\]
for all sufficiently large $T$.
Hence, $q_T\ge(2-\varepsilon)L_T$.
As $\varepsilon$ is arbitrary, $q_T\sim2L_T$, and therefore
\[
    h_T
    \sim
    \frac{2L_T}{\nu_T}=
    \Theta(\frac{\log T}{b_T-1}) \longrightarrow \infty.
\]
In particular, for fixed $b_T=b\ge2$,
$h_T = \Theta(\log T)$.

If $b_T=\Theta(\log T)$, then $r_T$ is bounded as both $L_t$ and $\nu_T$ are $\Theta(T)$.
Choose a fixed $s>1$ such that $I(s)>r_T$ eventually. This leads to $\mathbb P(Q_\nu\ge s\nu_T)
    \le
    \exp\{-\nu_T I(s)\} \leq \exp\{-\nu_T r_T\} = \alpha_T$. Hence, $q_T\le s\nu_T$, so
$1\le h_T\le s$ and $h_T=\Theta(1)$.

Finally, if $b_T/\log T\to\infty$, as shown in Case 3, $r_T\to0$. For every fixed $\varepsilon>0$,
$I(1+\varepsilon)>r_T$ eventually. Setting $s = 1+\epsilon$ and following the similar analysis above, we have $1\le h_T\le1+\varepsilon$. Consequently, $h_T\to1$. This completes the proof. 
\end{proof}

\section{Proof of Theorem \ref{thm:plugin_sk_confidence}}
\label{app:plugin_sk_confidence}
 \begin{proof}{Proof of Theorem \ref{thm:plugin_sk_confidence}.}
Let $\mathcal V_T$ be the event defined in the main text. The proof of Lemma~\ref{lem:plugin_variance} gives $\mathbb P(\mathcal V_T^c)\le\delta/2$. 
We first define an auxiliary noise scale that is valid on every sample path, including paths outside $\mathcal V_T$.

Before observing the simulation results at the $q$th point, the previous points have been observed and $x_q$ has been selected. For the analysis, we construct the filtration by first revealing the sample variance $S_q^2$ and then the sample mean $\bar Y_q$. Let $\mathcal H_{q-1/2}$ denote the information available after
this first reveal. Define $\varepsilon_q
    :={}
    \bar Y_q-f(x_q)$, $d_q^2
    :={}
    \frac{v(x_q)}{b_T}$, $r_q^2
    :={}
    \frac{\widetilde v_q}{b_T}$, and
    $r_q^{\circ}
    :={}
    \max\left\{
        r_q,
        \frac{d_q}{\sqrt{\kappa_T}} \right\}$.
The location $x_q$, the sample variance, and these scales are measurable with
respect to $\mathcal H_{q-1/2}$. Under Gaussian sampling, the sample mean and
sample variance within a batch are independent. Thus,
conditional on $\mathcal H_{q-1/2}$, $\varepsilon_q\sim N(0,d_q^2)$. Moreover, the definition of $r_q^\circ$ gives $d_q^2\le\kappa_T(r_q^\circ)^2$. Therefore, for every $t\in\mathbb R$,
\[
\begin{aligned}
    \mathbb E\!\left[
        \exp(t\varepsilon_q)
        \,\middle|\,
        \mathcal H_{q-1/2}
    \right]
    =
    \exp\left(\frac{t^2d_q^2}{2}\right)
    \le
    \exp\left(
        \frac{t^2\kappa_T(r_q^{\circ})^2}{2}
    \right).
\end{aligned}
\]
Hence, $\varepsilon_q$ is conditionally
$\sqrt{\kappa_T}r_q^\circ$-sub-Gaussian.

Let $\mu_n^{\circ}$ and $\sigma_n^{\circ}$ be the SK prediction and predictive variance
formed with $R_n^{\circ}
    :=
    \operatorname{diag}
    ((r_1^{\circ})^2,\ldots,(r_n^{\circ})^2)$,
and let $\Gamma_n^{\circ}
    :=
    \frac12
    \log\det
    \left(
        I_n+
        (R_n^{\circ})^{-1/2}
        K_n
        (R_n^{\circ})^{-1/2}
    \right)$.
We now apply the RKHS confidence inequality of
\citet[Lemma~7]{kirschner2018information} to the enlarged filtration.
In the notation of that lemma, set $\rho_q=\sqrt{\kappa_T}\,r_q^\circ$, $\Sigma_n=\kappa_T R_n^\circ$, $\lambda=\kappa_T^{-1}$, and $\delta_0=\frac{\delta}{2}$.
Then $\lambda\Sigma_n=R_n^\circ$, so the estimator in that lemma is
$\mu_n^\circ$. Its confidence-width function equals
$\sqrt{\kappa_T}\,\sigma_n^\circ$, and
\[
    \log\left(
        \frac{1}{\delta_0}
        \frac{\det(\lambda\Sigma_n+K_n)^{1/2}}
             {\det(\lambda\Sigma_n)^{1/2}}
    \right)
    =
    \Gamma_n^\circ+\log\frac{2}{\delta}.
\]
Therefore, with probability at least $1-\delta/2$, simultaneously for
all $n\ge1$ and all $x\in\mathcal X$,
\begin{equation}
\label{eq:oracle_confidence}
    |f(x)-\mu_n^{\circ}(x)|
    \le
    \left[
        B_f+
        \sqrt{
            2\kappa_T
            \left(
                \Gamma_n^{\circ}+
                \log\frac{2}{\delta}
            \right)
        }
    \right]
    \sigma_n^{\circ}(x).
\end{equation}
Denote the event that this bound holds as $\mathcal E_T^{\circ}$.

By construction,
\[
    (r_q^\circ)^2\ge r_q^2
    =\frac{\max\{v_0,S_q^2\}}{b_T}
    \ge\frac{v_0}{b_T}.
\]
Therefore, $(R_n^{\circ})^{-1}
    \preceq
    \frac{b_T}{v_0}I_n$.
Using the identity $\det(I+AB)=\det(I+BA)$, we can rewrite
\[
    \Gamma_n^{\circ}
    =
    \frac12
    \log\det
    \left(
        I_n+
        K_n^{1/2}(R_n^{\circ})^{-1}K_n^{1/2}
    \right).
\]
The Loewner-order bound on $(R_n^{\circ})^{-1}$ gives $K_n^{1/2}(R_n^{\circ})^{-1}K_n^{1/2}
    \preceq
    \frac{b_T}{v_0}K_n$.
Since the map $A\mapsto \log\det(I+A)$ is monotone on the positive
semidefinite cone, $\Gamma_n^{\circ}
    \le
    \frac12
    \log\det
    \left(
        I_n+
        \frac{b_T}{v_0}K_n
    \right)$.
By the definition of the replicated-noise information-gain envelope $\Gamma_n^{\circ}
    \le
    \gamma_n^{\mathrm{SK}}$, $\forall n\le N_T$.
Substituting this bound into \eqref{eq:oracle_confidence} gives the coefficient
$\Lambda_n$ in Theorem \ref{thm:plugin_sk_confidence}.

On $\mathcal V_T$, Lemma \ref{lem:plugin_variance} gives
$d_q^2\le\kappa_T r_q^2$ for every observed batch. Therefore,
$r_q^{\circ}=r_q$, and the auxiliary estimator is exactly the clipped plug-in
SK predictor: $(\mu_n^{\circ},\sigma_n^{\circ})
    =
    (\mu_n,\sigma_n)$, $n=1,\ldots,N_T$.
At $n=0$, the reproducing property and the kernel normalization give
\[
    |f(x)|
    \le
    \|f\|_{\mathcal H_{k_\theta}}
    \sqrt{k_\theta(x,x)}
    \le
    B_f\sigma_0(x)
    \le
    \Lambda_0\sigma_0(x).
\]
All variance and confidence inequalities hold jointly on
$\mathcal V_T\cap\mathcal E_T^\circ$.
Their total failure probability is at most
\[
    \mathbb P\left\{
        (\mathcal V_T
        \cap\mathcal E_T^\circ)^c
    \right\}
    \le\frac{\delta}{4}
       +\frac{\delta}{4}
       +\frac{\delta}{2}
    =\delta.
\]
The sequence $\gamma_n^{\mathrm{SK}}$ is nondecreasing by definition, so
$\Lambda_n$ is also nondecreasing.
\end{proof}

\section{Proof of Lemma \ref{lem:tail_cancellation}}
\begin{proof}{Proof of Lemma \ref{lem:tail_cancellation}.}
Fix $n<N_T$ and abbreviate $c=c_n$, $\xi=\xi_n^{\mathbf c}$, $g=g_{n+1}^{\mathbf c}$, and $x=x_{n+1}^{\mathbf c}$.
For every $y\in\mathcal X$, global quantile minimality gives $\xi
    \le
    \mu_n(y)-c\sigma_n(y)$.
Hence, $\frac{\xi-\mu_n(y)}{\sigma_n(y)}
    \le
    -c$,
and the monotonicity of $\tau$ gives $\operatorname{EI}_n(y;\xi)
    \le
    \sigma_n(y)\tau(-c)$.
At $g$, equality holds: $\operatorname{EI}_n(g;\xi)
    =
    \sigma_n(g)\tau(-c)$.
Since $x$ maximizes EI,
\[
    \sigma_n(x)\tau(-c)
    \ge
    \operatorname{EI}_n(x;\xi)
    \ge
    \operatorname{EI}_n(g;\xi)
    =
    \sigma_n(g)\tau(-c).
\]
The function $\tau$ is strictly positive, so the common factor cancels and
$\sigma_n(x)\ge\sigma_n(g)$. 
\end{proof}

\section{Proof of Lemma \ref{lem:q_bound}}
 \begin{proof}{Proof of Lemma \ref{lem:q_bound}.}
The function $\tau$ is continuous and strictly increasing because
$\tau'(z)=\Phi(z)>0$, and its range is $(0,\infty)$. Hence the equation
defining $q_c(r)$ has a unique solution.
Since $r\ge1$, $\tau(-q_c(r))
    \le
    \tau(-c)$,
which implies $q_c(r)\ge c$.

For $u\ge0$, define $H(u):=e^{u^2/2}\tau(-u)$. Differentiation of $H$ gives
\[
    H'(u)
    =
    e^{u^2/2}
    \left\{
        u\phi(u)-(1+u^2)\Phi(-u)
    \right\}.
\]
Since $(1+u^2)\Phi(-u)-u\phi(u)
    =
    \int_u^\infty (z-u)^2\phi(z)\,dz
    \ge 0$,
$H'(u)\le0$. Suppose first that $c\ge0$. Since $q_c(r)\ge c$ and $H$ is nonincreasing,
\[
    r
    =
    \frac{\tau(-c)}{\tau(-q_c(r))}
    \ge
    \exp\left(
        \frac{q_c^2(r)-c^2}{2}
    \right).
\]
Taking logarithms proves $q_c(r)
    \le
    \sqrt{c^2+2\log r}$.

Now suppose that $c<0$. If $q_c(r)\le0$, then
$q_c(r)\le\sqrt{2\log r}$ is immediate. If $q_c(r)>0$, monotonicity of
$\tau$ and the preceding tail-ratio bound with lower endpoint zero give
\[
    r
    =
    \frac{\tau(-c)}{\tau(-q_c(r))}
    \ge
    \frac{\tau(0)}{\tau(-q_c(r))}
    \ge
    \exp\left(\frac{q_c^2(r)}{2}\right).
\]
This proves the second case. 
\end{proof}

 \section{Proof of Lemma \ref{lem:ei_point_certificate}}
 \begin{proof}{Proof of Lemma \ref{lem:ei_point_certificate}.}
Fix $n<N_T$ and use the abbreviations from the proof of Lemma \ref{lem:tail_cancellation}. Let $s_x:=\sigma_n(x)$, $s_g:=\sigma_n(g)$, $\rho:=\frac{s_x}{s_g}$.
By Lemma \ref{lem:tail_cancellation}, $\rho\ge1$. EI optimality and the equality at the quantile
minimizer give
\[
\begin{aligned}
    s_x
    \tau\left(
        \frac{\xi-\mu_n(x)}{s_x}
    \right)
    =
    \operatorname{EI}_n(x;\xi)
    \ge
    \operatorname{EI}_n(g;\xi)
    =
    s_g\tau(-c)
    =
    s_x\frac{\tau(-c)}{\rho}.
\end{aligned}
\]
By the definition of $q_c(\rho)$ and the monotonicity of $\tau$, we have $\frac{\xi-\mu_n(x)}{s_x}
    \ge
    -q_c(\rho)$,
or equivalently, $\mu_n(x)-\xi
    \le
    q_c(\rho)s_x$. On $\mathcal E_T$, $f(x)
    \le
    \mu_n(x)+\Lambda_n s_x$.
Therefore,
\[
\begin{aligned}
    f(x)-f^\star
    \le
    f(x)-\xi+(\xi-f^\star)_+
    \le
    \left\{
        \Lambda_n+q_c(\rho)
    \right\}s_x
    +
    (\xi-f^\star)_+.
\end{aligned}
\]
This proves the first inequality in Lemma \ref{lem:ei_point_certificate}.

The definition of $\xi$ and the confidence bound at $x^\star$ give
\[
\begin{aligned}
    \xi
    \le
    \mu_n(x^\star)-c\sigma_n(x^\star)
    \le
    f^\star+
    (\Lambda_n-c)\sigma_n(x^\star).
\end{aligned}
\]
Taking positive parts proves the second inequality in Lemma \ref{lem:ei_point_certificate}.


It remains to replace the variance at $x^\star$ by $s_x$. We first derive the
scalar comparison needed for this step. For notation simplicity, let $\lambda=\Lambda_n$ and we first consider the scenario when $c<\lambda$.
At any point $y$ with $\sigma_n(y)>0$, define $u
    :=
    \frac{\xi-\mu_n(y)}{\sigma_n(y)}$, $t
    :=
    \frac{\xi-f(y)}{\sigma_n(y)}$.
Global quantile minimality gives $u\le-c$, while the confidence inequality
gives $|u-t|\le\lambda$.

We next show that \begin{equation}
\label{eq:appendix_transfer}
    \operatorname{EI}_n(y;\xi)
    \ge
    K_{\lambda,c}
    (\xi-f(y))_+,
\end{equation}
where $K_{\lambda,c}
    :=
    \inf_{-\lambda<u\le-c}
    \frac{\tau(u)}{u+\lambda}$.
If $t\le0$, then $(\xi-f(y))_+=0$, and the claimed inequality follows
from the nonnegativity of EI. Suppose now that $t>0$. Then
$t\le u+\lambda$, and hence $-\lambda<u\le-c$. By the definition of
$K_{\lambda,c}$, $\tau(u)
    \ge
    K_{\lambda,c}(u+\lambda)
    \ge
    K_{\lambda,c}t$.
Multiplying by $\sigma_n(y)$ gives
\[
\begin{aligned}
    \operatorname{EI}_n(y;\xi)
    =
    \sigma_n(y)\tau(u)
    \ge
    K_{\lambda,c}\sigma_n(y)t
    =
    K_{\lambda,c}(\xi-f(y))_+.
\end{aligned}
\]

To calculate this infimum in $K_{\lambda,c}$, let $R_\lambda(u)
    :=
    \frac{\tau(u)}{u+\lambda}$.
Since $\tau'(u)=\Phi(u)$, we have $R_\lambda'(u)
    =
    \frac{
        \lambda\Phi(u)-\phi(u)
    }{
        (u+\lambda)^2
    }$.
The ratio $\phi(u)/\Phi(u)$ is strictly decreasing from $+\infty$ to zero.
Hence, there is a unique point $z_\lambda$ satisfying $\frac{\phi(z_\lambda)}{\Phi(z_\lambda)}
    =
    \lambda$.
Moreover, 
\[
\begin{aligned}
    (z_\lambda+\lambda)\Phi(z_\lambda)
    =
    z_\lambda\Phi(z_\lambda)
    +
    \lambda\Phi(z_\lambda)
    =
    z_\lambda\Phi(z_\lambda)
    +
    \phi(z_\lambda)
    =
    \tau(z_\lambda).
\end{aligned}
\]
Since $\Phi(z_\lambda)>0$ and $\tau(z_\lambda)>0$, it follows that $z_\lambda+\lambda>0$, and hence $z_\lambda>-\lambda$. Finally,
\[
    R_\lambda'(u)
    =
    \frac{\Phi(u)}{(u+\lambda)^2}
    \left(
        \lambda-\frac{\phi(u)}{\Phi(u)}
    \right).
\]
Therefore, $R_\lambda$ is decreasing on $(-\lambda,z_\lambda)$ and
increasing on $(z_\lambda,\infty)$. If
$-c\ge z_\lambda$, the interval $(-\lambda,-c]$ contains $z_\lambda$, and $K_{\lambda,c}
    =
    R_\lambda(z_\lambda)
    =
    \Phi(z_\lambda)$.
If $-z_\lambda<c<\lambda$, the interval ends before $z_\lambda$, and $K_{\lambda,c}
    =
    R_\lambda(-c)
    =
    \frac{\tau(-c)}{\lambda-c}$.
The two expressions agree at $c=-z_\lambda$.

Apply \eqref{eq:appendix_transfer} at $y=x^\star$ with
$\lambda=\Lambda_n$. EI optimality gives
\[
\begin{aligned}
    K_{\Lambda_n,c}(\xi-f^\star)_+
    \le
    \operatorname{EI}_n(x^\star;\xi)
    \le
    \operatorname{EI}_n(x;\xi).
\end{aligned}
\]
Global quantile minimality also gives $\operatorname{EI}_n(x;\xi)
    \le
    s_x\tau(-c)$.
Therefore,
\[
    (\xi-f^\star)_+
    \le
    \frac{\tau(-c)}{K_{\Lambda_n,c}}s_x
    =
    A(\Lambda_n,c)s_x.
\]

We next consider the scenario when $c\ge\Lambda_n$, the second inequality in Lemma \ref{lem:ei_point_certificate} already gives
$(\xi-f^\star)_+=0$, which is the first case in the definition of $A$.
This proves the third inequality in Lemma \ref{lem:ei_point_certificate}. Substituting it into the first
inequality proves the final certificate in that lemma. The calculation of
$K_{\lambda,c}$ also shows that no smaller coefficient than $ A(\Lambda_n,c)$ exists under
the two scalar conditions $u\le-c$ and $|u-t|\le\lambda$.

\end{proof}

\section{Proof of Lemma \ref{lem:variance_ratio}}
\begin{proof}{Proof of Lemma \ref{lem:variance_ratio}.}
The upper bound $\sigma_n^2(x)\le1$ follows from
$k_\theta(x,x)\le1$ and the fact that conditioning cannot increase posterior
variance.

For the lower bound, write $k_n(x)
    =
    \bigl(
        k_\theta(x,x_1),\ldots,k_\theta(x,x_n)
    \bigr)^\top$.
The joint prior covariance of $f(x)$ and
$(f(x_1),\ldots,f(x_n))$ is positive semidefinite:
\[
	\begin{pmatrix}
		k_\theta(x,x) & k_n(x)^\top\\
		k_n(x) & K_n
	\end{pmatrix}\succeq 0.
	\]
Since $k_\theta(x,x)\ge k_{\min}>0$, its Schur complement gives $K_n-k_n(x)k_n(x)^\top/k_\theta(x,x)\succeq0$. Also, $R_n
    \succeq
    \frac{v_0}{b_T}I_n$. 
Let $r_0^2:=v_{0}/b_T$. Then
\[
    K_n+R_n
    \succeq
    k_n(x)k_n(x)^\top/k_\theta(x,x)+r_0^2I_n.
\]
Taking inverses reverses the Loewner order. With $u=k_n(x)$,
\[
\begin{aligned}
    u^\top(K_n+R_n)^{-1}u
    \le
    u^\top(uu^\top/k_\theta(x,x)+r_0^2I_n)^{-1}u
    =
    \frac{k_\theta(x,x)\|u\|^2}{k_\theta(x,x)r_0^2+\|u\|^2}.
\end{aligned}
\]
Kernel normalization and Cauchy's inequality imply
$|k_\theta(x,x_i)|^2\le k_\theta(x,x)k_\theta(x_i,x_i)\le k_\theta(x,x)$, so $\|u\|^2\le nk_\theta(x,x)$. Therefore,
\[
\begin{aligned}
    \sigma_n^2(x)=
    k_\theta(x,x)-u^\top(K_n+R_n)^{-1}u
    \ge
    \frac{k_\theta(x,x)^2r_0^2}{k_\theta(x,x)r_0^2+\|u\|^2}\ge \frac{k_\theta(x,x)r_0^2}{r_0^2+n}
    \ge
    \frac{k_{\min}}{1+nb_T/v_0}.
\end{aligned}
\]
For $n=0$, both bounds follow from $\sigma_0^2(x)=k_\theta(x,x)$.

Combining the upper bound at $x_{n+1}^{\mathbf c}$ with the lower bound at
$g_{n+1}^{\mathbf c}$ gives $\rho_n^{\mathbf c}
    \le
    \sqrt{\frac{1+nb_T/v_0}{k_{\min}}}$.
The lower bound $\rho_n^{\mathbf c}\ge1$ is Lemma \ref{lem:tail_cancellation}. Lemma \ref{lem:q_bound} then gives
$q_{c_n}(\rho_n^{\mathbf c})\le\bar q_n(c_n)$ in both cases
$c_n\ge0$ and $c_n<0$. 
\end{proof}

\section{Proof of Lemma \ref{lem:variance_sum}}
\begin{proof}{Proof of Lemma \ref{lem:variance_sum}.}
For brevity, write $s_n
    :=
    \sigma_n(x_{n+1}^{\mathbf c})$.
The determinant chain rule for the realized clipped plug-in SK predictor gives
\[
    \Gamma_{N_T}(X_{N_T},R_{N_T})
    =
    \frac12
    \sum_{n=0}^{N_T-1}
    \log\left(
        1+\frac{s_n^2}{r_{n+1}^2}
    \right).
\]
On $\mathcal V_T$, $r_{n+1}^2\le W_T/b_T$. Hence,
\[
    \log\left(
        1+\frac{s_n^2}{r_{n+1}^2}
    \right)
    \ge
    \log\left(
        1+\frac{b_T}{W_T}s_n^2
    \right).
\]
Since $0\le s_n^2\le1$, concavity gives
\[
    \log\left(
        1+\frac{b_T}{W_T}s_n^2
    \right)
    \ge
    s_n^2
    \log\left(
        1+\frac{b_T}{W_T}
    \right).
\]
It follows that
\[
    \Gamma_{N_T}(X_{N_T},R_{N_T})
    \ge
    \frac12
    \log\left(
        1+\frac{b_T}{W_T}
    \right)
    \sum_{n=0}^{N_T-1}s_n^2.
\]

Also, $R_{N_T}\succeq v_0/b_T I_{N_T}$. The log-determinant comparison
used in the proof of Theorem \ref{thm:plugin_sk_confidence} therefore gives $\Gamma_{N_T}(X_{N_T},R_{N_T})
    \le
    \gamma_{N_T}^{\mathrm{SK}}$.
Combining the last two displays proves
\[
    \sum_{n=0}^{N_T-1}s_n^2
    \le
    \frac{
        2\gamma_{N_T}^{\mathrm{SK}}
    }{
        \log\left(1+b_T/W_T\right)
    }.
\]
\end{proof}

\section{Proof of Theorem \ref{thm:global_quantile_family_certificate}}

\begin{proof}{Proof of Theorem \ref{thm:global_quantile_family_certificate}.}
We work on $\mathcal G_T$. For brevity, we write $s_n
    :=
    \sigma_n(x_{n+1}^{\mathbf c}), N:=N_T$ and $\lambda_T:=\Lambda_N$.
Lemma \ref{lem:variance_sum} gives $\sum_{n=0}^{N-1}s_n^2
    \le
    \frac{
        2\gamma_{N}^{\mathrm{SK}}
    }{
        \log\left(1+b_T/W_T\right)
    }$.
Choose $j\in\{0,\ldots,N-1\}$ such that $s_j$ is no larger than the root-mean-square of these values. Since $N\ge1$ and 
$\lfloor x\rfloor\ge x/2$ for every $x\ge1$, applied with
$x=T/b_T$ we have, $N_T
    =
    \left\lfloor\frac{T}{b_T}\right\rfloor
    \ge
    \frac{T}{2b_T}$.
It follows that
\begin{equation}
\label{eq:appendix_small_variance}
    s_j
    \le \eta_T \le
    2
    \sqrt{
        \frac{
            b_T\gamma_{N}^{\mathrm{SK}}
        }{
            T\log\left(1+b_T/W_T\right)
        }
    }.
\end{equation}
By Lemmas \ref{lem:ei_point_certificate} and \ref{lem:variance_ratio} and the definition of $D_T(\mathbf c)$,
\begin{equation}
\label{eq:appendix_selected_point_regret}
    f(x_{j+1}^{\mathbf c})-f^\star
    \le
    D_T(\mathbf c)s_j
    \le
    D_T(\mathbf c)\eta_T.
\end{equation}

We next control the terminal posterior variance at every sampled location. When the $q$th point is added, the one-step posterior variance update at its own location gives
\[
\begin{aligned}
    \sigma_q^2(x_q^{\mathbf c})
    =
    \sigma_{q-1}^2(x_q^{\mathbf c})
    -
    \frac{
        \sigma_{q-1}^4(x_q^{\mathbf c})
    }{
        \sigma_{q-1}^2(x_q^{\mathbf c})+r_q^2
    }
    =
    \frac{
        \sigma_{q-1}^2(x_q^{\mathbf c})r_q^2
    }{
        \sigma_{q-1}^2(x_q^{\mathbf c})+r_q^2
    }.
\end{aligned}
\]
Lemma~\ref{lem:variance_ratio} gives $\sigma_{q-1}^2(x_q^{\mathbf c})\le1$. Since $\mathcal G_T\subseteq\mathcal V_T$,
$r_q^2\le W_T/b_T$, and therefore $\sigma_q^2(x_q^{\mathbf c})
    \le\frac{W_T}{b_T+W_T}
    =\omega_T^2$.

For the fixed kernel and the realized sequence of appended positive noise
scales, posterior variance is nonincreasing as further observations are
added. Hence,
\begin{equation}
\label{eq:appendix_terminal_sampled_variance}
    \sigma_N(x_q^{\mathbf c})
    \le
    \omega_T,
    \qquad q=1,\ldots,N.
\end{equation}
For $x_{j+1}^{\mathbf c}$ selected in \eqref{eq:appendix_small_variance}, posterior-variance monotonicity also gives $\sigma_N(x_{j+1}^{\mathbf c})
    \le
    \sigma_j(x_{j+1}^{\mathbf c})
    =
    s_j
    \le
    \eta_T$.
Combining the last inequality with
\eqref{eq:appendix_terminal_sampled_variance} yields
\begin{equation}
\label{eq:appendix_terminal_comparator_variance}
    \sigma_N(x_{j+1}^{\mathbf c})
    \le
    \zeta_T.
\end{equation}

Fix any $d_T\in\mathbb R$ and abbreviate $\hat x
    :=
    \hat x_T^{\mathbf c,d_T}$.
Since $\hat x$ minimizes
$\mu_N(z)+d_T\sigma_N(z)$ over the sampled set and
$x_{j+1}^{\mathbf c}$ belongs to that set, the confidence inequalities on
$\mathcal G_T\subseteq\mathcal E_T$ give
\begin{align*}
    f(\hat x)-f^\star
    &\le
    \mu_N(\hat x)
    +
    \lambda_T\sigma_N(\hat x)
    -
    f^\star\\
    &=
    \left\{
        \mu_N(\hat x)
        +
        d_T\sigma_N(\hat x)
    \right\}
    +
    (\lambda_T-d_T)\sigma_N(\hat x)
    -
    f^\star\\
    &\le
    \mu_N(x_{j+1}^{\mathbf c})
    +
    d_T\sigma_N(x_{j+1}^{\mathbf c})
    +
    (\lambda_T-d_T)\sigma_N(\hat x)
    -
    f^\star\\
    &\le
    f(x_{j+1}^{\mathbf c})-f^\star
    +
    (\lambda_T+d_T)
    \sigma_N(x_{j+1}^{\mathbf c})
    +
    (\lambda_T-d_T)
    \sigma_N(\hat x)\\
    &\le
    f(x_{j+1}^{\mathbf c})-f^\star
    +
    (\lambda_T+d_T)_+
    \sigma_N(x_{j+1}^{\mathbf c})
    +
    (\lambda_T-d_T)_+
    \sigma_N(\hat x).
\end{align*}
Because $\hat x$ is sampled, \eqref{eq:appendix_terminal_sampled_variance} gives
$\sigma_N(\hat x)\le\omega_T$. Together with
\eqref{eq:appendix_selected_point_regret} and
\eqref{eq:appendix_terminal_comparator_variance}, this proves
\begin{equation}
\label{eq:appendix_joint_terminal_certificate}
\begin{aligned}
    f(\hat x_T^{\mathbf c,d_T})-f^\star
    \le{}
    D_T(\mathbf c)\eta_T
    +
    (\lambda_T+d_T)_+\zeta_T
    +
    (\lambda_T-d_T)_+\omega_T.
\end{aligned}
\end{equation}

When $d_T=\lambda_T=\Lambda_{N_T}$,
\[
\begin{aligned}
    f(\hat x_T^{\mathbf c,\Lambda_{N_T}})-f^\star
    \le
    D_T(\mathbf c)\eta_T
    +
    2\Lambda_{N_T}\zeta_T
    \le
    \left\{
        D_T(\mathbf c)+2\Lambda_{N_T}
    \right\}\eta_T.
\end{aligned}
\]
Using \eqref{eq:appendix_small_variance}, we obtain
\[
    f(\hat x_T^{\mathbf c,\Lambda_{N_T}})-f^\star
    \le
    2
    \left\{
        2\Lambda_{N_T}+D_T(\mathbf c)
    \right\}
    \sqrt{
        \frac{
            b_T\gamma_{N_T}^{\mathrm{SK}}
        }{
            T\log(1+b_T/W_T)
        }
    }.
\]
Since $\mathbb P(\mathcal G_T)\ge1-\delta$, all stated conclusions follow.
\end{proof}

\section{Proof of Corollary \ref{thm:observed_simple_regret}}

\begin{proof}{Proof of Corollary \ref{thm:observed_simple_regret}.}
Set $c_n=\Lambda_n$ and $d_T=\Lambda_{N_T}$ for every $n<N_T$. Then $A(\Lambda_n,c_n)=0$
and, because $\Lambda_n$ and $\ell_n$ are nondecreasing,
\[
\begin{aligned}
    D_T(\mathbf c)
    =
    \max_{n<N_T}
    \left\{
        \Lambda_n+
        \sqrt{\Lambda_n^2+\ell_n}
    \right\}
    \le
    \Lambda_{N_T}
    +
    \sqrt{
        \Lambda_{N_T}^2+
        \log\left(
            \frac{1+T/v_0}{k_{\min}}
        \right)
    }.
\end{aligned}
\]
Let
\[
    J_T
    :=
    \sqrt{
        \Lambda_{N_T}^2+
        \log\left(
            \frac{1+T/v_0}{k_{\min}}
        \right)
    }.
\]
Since $\Lambda_{N_T}\le J_T$,
\[
    2\left\{
        2\Lambda_{N_T}+D_T(\mathbf c)
    \right\}
    \le
    2(3\Lambda_{N_T}+J_T)
    \le
    8J_T.
\]
Substitution into Theorem \ref{thm:global_quantile_family_certificate} gives the claimed bound.
\end{proof}

\section{Proof of Lemma \ref{lem:incumbent_coefficient_optimality}}

\begin{proof}{Proof of Lemma \ref{lem:incumbent_coefficient_optimality}.}
For ease of exposition, we write $\lambda:=\Lambda_n>0$ in this proof. We first show that, $\forall c\in\mathbb R$, $A(\lambda,c)\ge(\lambda-c)_+$. This holds with equality when $c\ge\lambda$ or
$c_\lambda^\circ<c<\lambda$ by the definition of $A$.
Suppose that $c\le c_\lambda^\circ=-z_\lambda$.
Since $\tau'(u)=\Phi(u)$ and $\tau''(u)=\phi(u)>0$,
convexity of $\tau$ function and the definition of $z_\lambda$ give
\[
    \tau(-c)
    \ge
    \tau(z_\lambda)
    +\Phi(z_\lambda)(-c-z_\lambda)
    =
    (\lambda-c)\Phi(z_\lambda),
\]
where the equality uses
$\phi(z_\lambda)=\lambda\Phi(z_\lambda)$ and hence
$\tau(z_\lambda)=(z_\lambda+\lambda)\Phi(z_\lambda)$.
The latter identity also implies
$z_\lambda+\lambda>0$, so $c_\lambda^\circ<\lambda$.
Dividing by $\Phi(z_\lambda)>0$ yields
\[
    A(\lambda,c)
    =
    \frac{\tau(-c)}{\Phi(z_\lambda)}
    \ge\lambda-c
    =(\lambda-c)_+.
\]

Next, the definition of $\bar q_n$ can be written as $\bar q_n(c)
    =
    \sqrt{(c_+)^2+\ell_n}
    =
    \left\|(c_+,\sqrt{\ell_n})\right\|_2$,
where $\ell_n\ge0$.
Thus $\bar q_n$ is nondecreasing.
The triangle inequality gives, for any
$c_1,c_2\in\mathbb R$,
\[
    |\bar q_n(c_2)-\bar q_n(c_1)|
    \le
    |(c_2)_+-(c_1)_+|
    \le |c_2-c_1|.
\]
Therefore, $\bar q_n$ is $1$-Lipschitz.

Thus, If $c<\lambda$, we have $\bar q_n(\lambda)-\bar q_n(c)\le\lambda-c$. Combining the bound for $A$, we have
\[
    \bar q_n(c)+A(\lambda,c)
    \ge
    \bar q_n(c)+\lambda-c
    \ge
    \bar q_n(\lambda).
\]
If $c\ge\lambda$, then $A(\lambda,c)=0$, and
monotonicity gives the same inequality.
\end{proof}

\section{Proofs for the Replication-Number Analysis}
\label{app:replication_number_analysis}

The replication analysis below uses the calibrated consequence of Theorem~\ref{thm:global_quantile_family_certificate} obtained by setting $d_T=\Lambda_{N_T}$. This gives exactly the bound in Corollary~\ref{thm:observed_simple_regret} and hence the same replication objective.
We first define some notations used in the replication-number
analysis:
\[
    F_T(b)
    :=
    \frac{N_T(b)}{2}
    \log\left(
        1+\frac{b}{v_0}
    \right).
\]
\[
    \overline\Lambda_T^{\,k}(b)
    :=
    B_f+
    \sqrt{
        2\kappa_T(b)
        \left(
            \overline\gamma_T^{\,k}(b)
            +
            \log\frac{2}{\delta}
        \right).
    }
\]

\subsection{Proof of the Finite-Batch Information-Gain Bound}
\label{app:finite_batch_information_gain}

\begin{proof}{Proof of
Lemma~\ref{lem:finite_batch_information_gain}.}
First fix $1\le m\le N_T(b)$ and any collection of points $A=(a_1,\ldots,a_m)\in\mathcal X^m$.
Let $\lambda_1,\ldots,\lambda_m$ be the nonnegative eigenvalues of the positive semidefinite matrix $K_A$.
Then
\[
\begin{aligned}
    \frac12
    \log\det\left(
        I_m+\frac{b}{v_0}K_A
    \right)
    &=
    \frac12
    \sum_{i=1}^m
    \log\left(
        1+\frac{b}{v_0}\lambda_i
    \right).
\end{aligned}
\]
The function $z
    \longmapsto
    \log\left(
        1+\frac{b}{v_0}z
    \right)$
is concave on $[0,\infty)$. Jensen's inequality therefore gives
\[
\begin{aligned}
    \frac12
    \sum_{i=1}^m
    \log\left(
        1+\frac{b}{v_0}\lambda_i
    \right)
    &\le
    \frac{m}{2}
    \log\left(
        1+
        \frac{b}{v_0}
        \frac{1}{m}
        \sum_{i=1}^m\lambda_i
    \right).
\end{aligned}
\]
The sum of the eigenvalues equals the trace:
\[
\begin{aligned}
    \sum_{i=1}^m\lambda_i
    =
    \operatorname{tr}(K_A)
    =
    \sum_{i=1}^m
    k_\theta(a_i,a_i).
\end{aligned}
\]
Since the kernel is normalized so that $k_\theta(x,x)\le1$
for every $x\in\mathcal X$, $\operatorname{tr}(K_A)
    \le
    m$.
Consequently,
\[
\begin{aligned}
    \frac12
    \log\det\left(
        I_m+\frac{b}{v_0}K_A
    \right)
    \le
    \frac{m}{2}
    \log\left(
        1+\frac{b}{v_0}
    \right)
    \le
    \frac{N_T(b)}{2}
    \log\left(
        1+\frac{b}{v_0}
    \right).
\end{aligned}
\]
The last expression does not depend on the points in $A$ or on
$m\le N_T(b)$. Taking the supremum over all admissible $m$ and $A$ proves
\[
    \gamma_T(b)
    \le
    \frac{N_T(b)}{2}
    \log\left(
        1+\frac{b}{v_0}
    \right).
\]
\end{proof}

\subsection{Proof of the General-Kernel Replication Theorem}
\label{app:general_kernel_replication}

\begin{proof}{Proof of
Theorem~\ref{thm:general_kernel_replication}.}
Fix $\delta\in(0,1)$ and write $g_T
    :=
    G_k\left(
        \frac{T}{v_0}
    \right)$.
For every $b\in\mathcal B_T$, $T-b
    <
    bN_T(b)
    \le
    T$.
Since $b\le T/2$, $\frac{T}{2}
    \le
    bN_T(b)
    \le
    T$.
Therefore, $\frac{T}{2v_0}
    \le
    U_T(b)
    \le
    \frac{T}{v_0}$.
As $G_k$ is nondecreasing, $G_k\left(
        \frac{T}{2v_0}
    \right)
    \le
    G_k\bigl(U_T(b)\bigr)
    \le
    g_T$.
The condition $G_k(2u)
    \le
    C_kG_k(u)$
gives, for all sufficiently large $T$, $g_T
    \le
    C_k
    G_k\left(
        \frac{T}{2v_0}
    \right)$.
Hence,
\begin{equation}
\label{eq:appendix_uniform_spectral_bound}
    G_k\bigl(U_T(b)\bigr)
    =
    \Theta(g_T)
\end{equation}
uniformly over $b\in\mathcal B_T$.

We first evaluate the bound at a logarithmic replication number. Fix any
constant $c>0$ and define $b_0
    :=
    \left\lceil
        c\log T
    \right\rceil$.
For all sufficiently large $T$, $b_0\in\mathcal B_T$.
Moreover, $N_T(b_0)
    =
    \Theta\left(
        \frac{T}{\log T}
    \right)$,
and
\begin{equation*}
\label{eq:appendix_finite_bound_at_log_b}
\begin{aligned}
    F_T(b_0)
    =
    \frac{N_T(b_0)}{2}
    \log\left(
        1+\frac{b_0}{v_0}
    \right)
    =
    \Theta\left(
        \frac{
            T\log\log T
        }{
            \log T
        }
    \right).
\end{aligned}
\end{equation*}
Moreover, denote $L_b$ as $\log(4N_T(b)/\delta)$. By the Laurent--Massart upper-tail inequality \citep[eq.~(4.3)]{laurent2000adaptive} and the definition of $h_T(b)$, we have
\[
    1\le h_T(b)\le
    1+2\sqrt{\frac{L_b}{\nu_b}}
    +2\frac{L_b}{\nu_b}.
\]
Thus, $h_T(b_0)=O(1)$, $W_T(b_0)=\Theta(1)$, and $\log(1+b_0/W_T(b_0))
    =\Theta(\log\log T)$.

By assumption, $g_T
    =
    o\left(
        \sqrt{
            T\log\log T
        }
    \right)$.
Since
\[
    \sqrt{
        T\log\log T
    }
    =
    o\left(
        \frac{
            T\log\log T
        }{
            \log T
        }
    \right),
\]
we have $g_T
    =
    o\left(
        F_T(b_0)
    \right)$.
Together with
\eqref{eq:appendix_uniform_spectral_bound}, this gives
\begin{equation*}
\label{eq:appendix_log_candidate_information}
    \overline\gamma_T^{\,k}(b_0)
    =
    G_k\bigl(U_T(b_0)\bigr)
    =
    \Theta(g_T).
\end{equation*}

Furthermore, $\log\frac{
        3N_T(b_0)
    }{
        \delta
    }
    =
    \Theta(\log T)$.
Since $b_0
    =
    \Theta(\log T)$,
Lemma~\ref{lem:plugin_correction_order} gives $\kappa_T(b_0)
    =
    O(1)$.
The assumption $\log T
    =
    O(g_T)$
then implies
\[
    \left(
        \overline\Lambda_T^{\,k}(b_0)
    \right)^2
    +
    \log\left(
        \frac{1+T/v_0}{k_{\min}}
    \right)
    =
    O(g_T).
\]
Therefore,
\begin{equation*}
\label{eq:appendix_log_candidate_regret}
\begin{aligned}
    \left(
        \overline{\mathcal R}_T^{\,k}(b_0)
    \right)^2
    &=
    O\left(
        \frac{
            g_T^2\log T
        }{
            T\log\log T
        }
    \right).
\end{aligned}
\end{equation*}
Since $g_T^2
    =
    o\left(
        T\log\log T
    \right)$,
we obtain
\begin{equation}
\label{eq:appendix_log_candidate_smaller_than_log}
    \left(
        \overline{\mathcal R}_T^{\,k}(b_0)
    \right)^2
    =
    o(\log T).
\end{equation}

We next show that a minimizer cannot lie in the range where the finite-batch
bound is active, i,e., $F_T(b) \leq G_k(U_T(b))$ and thus $\overline\gamma_T^{\,k}(b)
    =
    F_T(b)$. Suppose that, for some $b\in\mathcal B_T$, $\overline\gamma_T^{\,k}(b)
    =
    F_T(b)$.
Then
\[
\begin{aligned}
    \frac{
        b\overline\gamma_T^{\,k}(b)
    }{
        T\log(1+b/ W_T(b))
    }
    &=
    \frac{
        bN_T(b)
    }{
        2T
    }
    \frac{
        \log(1+b/v_0)
    }{
        \log(1+b/ W_T(b))
    }.
\end{aligned}
\]
Because $bN_T(b)
    \ge
    \frac{T}{2}$
and $W_T(b)\ge M\ge v_0$,
we have
\[
    \frac{
        b\overline\gamma_T^{\,k}(b)
    }{
        T\log(1+b/ W_T(b))
    }
    \ge
    \frac14.
\]
Also,
\[
    \left(
        \overline\Lambda_T^{\,k}(b)
    \right)^2
    +
    \log\left(
        \frac{1+T/v_0}{k_{\min}}
    \right)
    \ge
    \log\left(
        \frac{1+T/v_0}{k_{\min}}
    \right).
\]
It follows that
\begin{equation}
\label{eq:appendix_finite_batch_regret_lower}
\begin{aligned}
    \left(
        \overline{\mathcal R}_T^{\,k}(b)
    \right)^2
    \ge
    16
    \log\left(
        \frac{1+T/v_0}{k_{\min}}
    \right)
    =
    \Omega(\log T).
\end{aligned}
\end{equation}
Comparison of
\eqref{eq:appendix_log_candidate_smaller_than_log} and
\eqref{eq:appendix_finite_batch_regret_lower} shows that, for all
sufficiently large $T$, no minimizer can lie in the range where the
finite-batch bound is active.

It remains to minimize the bound over replication numbers for which $\overline\gamma_T^{\,k}(b)
    =
    G_k\bigl(U_T(b)\bigr)$.
By
\eqref{eq:appendix_uniform_spectral_bound}, $\overline\gamma_T^{\,k}(b)
    =
    \Theta(g_T)$ uniformly over this range.
Since $\kappa_T(b)
    \ge
    1$,
   $\log T
    =
    O(g_T)$,
we have
\begin{equation*}
\label{eq:appendix_confidence_order_spectral_range}
\begin{aligned}
    \left(
        \overline\Lambda_T^{\,k}(b)
    \right)^2
    +
    \log\left(
        \frac{1+T/v_0}{k_{\min}}
    \right)
    =
    \Theta\left(
        \kappa_T(b)g_T
    \right).
\end{aligned}
\end{equation*}
Consequently,
\begin{equation}
\label{eq:appendix_regret_reduced_to_H}
    \left(
        \overline{\mathcal R}_T^{\,k}(b)
    \right)^2
    =
    \Theta\left(
        \frac{
            g_T^2
        }{
            T
        }
        H_T(b)
    \right),
\end{equation}
where
\begin{equation*}
\label{eq:appendix_H_definition}
    H_T(b):=
    \frac{\kappa_T(b)b}
         {\log(1+b/W_T(b))}.
\end{equation*}
Since $h_T(b)\ge1$, $H_T(b)\ge
    \underline H_T(b):=
    \frac{\kappa_T(b)b}{\log(1+b/M)}$.
For $b=\Theta(\log T)$, both $\kappa_T(b)$ and $h_T(b)$
are bounded above by constants and below by one. Thus,
\begin{equation}
\label{eq:appendix_H_log_order}
    H_T(b)=
    \Theta\left(\frac{\log T}{\log\log T}\right).
\end{equation}

We now consider replication numbers smaller than logarithmic order. Suppose
that $b=o(\log T)$.
Recall that $\nu_b
    =
    b-1$,
   and
    $L_b
    =
    \log\frac{
        4N_T(b)
    }{
        \delta
    }$.
Since $\delta$ is fixed and $b=o(\log T)$, $L_b
    =
    \Theta(\log T)$,
    $\frac{L_b}{\nu_b}
    \longrightarrow
    \infty$.
According to Lemma~\ref{lem:plugin_correction_order}, $\kappa_T(b)
\ge
    c_0^2
    \exp\left(
        \frac{2L_b}{\nu_b}
    \right) \longrightarrow \infty$.

Since $b=o(\log T)$, we also have $b\le L_b$ for all sufficiently large $T$, and therefore $\log\left(
        1+\frac{b}{M}
    \right)
    =
    O(\log L_b)$.
Thus,
\[
\begin{aligned}
    \frac{
        \underline H_T(b)
    }{
        L_b/\log L_b
    }
    \ge
    c_1
    \exp(\frac{2L_b}{\nu_b})
    \frac{b}{L_b}
    \frac{
        \log L_b
    }{
        \log(1+b/ M)
    }
    \ge
    c_2
    \frac{
        \exp(\frac{2L_b}{\nu_b})
    }{
        \frac{2L_b}{\nu_b}
    }
    \longrightarrow
    \infty.
\end{aligned}
\]
Since $L_b=\Theta(\log T)$, this gives
\begin{equation}
\label{eq:appendix_small_b_H_larger}
    \underline H_T(b)
    \gg
    \frac{
        \log T
    }{
        \log\log T
    }.
\end{equation}
Thus, a replication number smaller than logarithmic order gives a larger
bound than \eqref{eq:appendix_H_log_order}.

Finally, suppose that $\frac{b}{\log T}
    \longrightarrow
    \infty$.
Since $\kappa_T(b)\ge1$, $\underline H_T(b)
    \ge
    \frac{
        b
    }{
        \log(1+b/ M)
    }$.
Let $s_T
    :=
    \frac{b}{\log T}$.
Then $s_T\to\infty$, and
\[
\begin{aligned}
    \frac{
        \underline H_T(b)
    }{
        \log T/\log\log T
    }
    &\ge
    \frac{
        s_T\log\log T
    }{
        \log\left(
            1+
            s_T\log T/ M
        \right)
    }.
\end{aligned}
\]
The denominator satisfies
\[
    \log\left(
        1+
        s_T\log T/ M
    \right)
    \le
    C+
    \log s_T+
    \log\log T
\]
for a constant $C$. Therefore,
\[
    \frac{
        s_T\log\log T
    }{
        C+\log s_T+\log\log T
    }
    \longrightarrow
    \infty,
\]
and hence
\begin{equation}
\label{eq:appendix_large_b_H_larger}
    \underline H_T(b)
    \gg
    \frac{
        \log T
    }{
        \log\log T
    }.
\end{equation}
Thus, a replication number larger than logarithmic order also gives a larger
bound than \eqref{eq:appendix_H_log_order}. Since $H_T(b)\ge\underline H_T(b)$, these lower bounds
also apply to $H_T(b)$. Therefore, neither sublogarithmic
nor superlogarithmic replication can minimize the bound.

To summarize, we have shown that a minimizing sequence cannot satisfy $b_T^{\mathrm{bd}}
    =
    o(\log T)$
or $\frac{
        b_T^{\mathrm{bd}}
    }{
        \log T
    }
    \longrightarrow
    \infty$.
We finally verify that the preceding comparisons cover arbitrary minimizing
sequences, including those for which the ratio to $\log T$ has no limit. Let
$r_T:=\frac{b_T^{\mathrm{bd}}}{\log T}$. If $b_T^{\mathrm{bd}}$ were not of order $\log T$, then either $\liminf_{T\to\infty}r_T=0$ or $\limsup_{T\to\infty}r_T=\infty$.
In the first case, there exists a subsequence $T_j\to\infty$ such that
$r_{T_j}\to0$. Equations
\eqref{eq:appendix_regret_reduced_to_H},
\eqref{eq:appendix_H_log_order}, and
\eqref{eq:appendix_small_b_H_larger} then imply that the logarithmic
candidate $b_0(T_j)$ gives a strictly smaller bound than
$b_{T_j}^{\mathrm{bd}}$ for all sufficiently large $j$. This contradicts
the definition of $b_{T_j}^{\mathrm{bd}}$. The same contradiction follows
from \eqref{eq:appendix_large_b_H_larger} if a subsequence satisfies
$r_{T_j}\to\infty$. Therefore, we have
\[
    0<
    \liminf_{T\to\infty}r_T
    \le
    \limsup_{T\to\infty}r_T
    <\infty,
\]
which is equivalent to $b_T^{\mathrm{bd}}=\Theta(\log T)$.
\end{proof}

\section{Kernel Classes Satisfying the Conditions of
Theorem~\ref{thm:general_kernel_replication}}
\label{app:kernels_satisfying_replication_conditions}

This section verifies the three conditions in
Theorem~\ref{thm:general_kernel_replication} for several commonly used kernel
classes. Recall that $g_T=
    G_k\left(
        \frac{T}{v_0}
    \right)$,
where $G_k$ is a deterministic spectral upper bound on the maximum
information gain. We verify the conditions below.

\subsection{Finite-rank kernels}

Suppose that the kernel has finite rank $r$. Taking $D=r$ in
Theorem~3 of \cite{vakili2021information} removes the spectral-tail term.
Under the normalization $k_\theta(x,x)\le1$, we may use
\begin{equation*}
\label{eq:finite_rank_G}
    G_k(u)
    =
    \frac{r}{2}
    \log\left(
        1+\frac{u}{r}
    \right).
\end{equation*}
For fixed $r$, this function is nondecreasing and
\[
\begin{aligned}
    \frac{G_k(2u)}{G_k(u)}
    &=
    \frac{
        \log(1+2u/r)
    }{
        \log(1+u/r)
    }
    =
    O(1).
\end{aligned}
\]
Thus, the first condition holds.

Moreover, $g_T
    =
    \Theta(\log T)$,
because $v_0$ is fixed. Therefore, $\log T
    =
    O(g_T)$,
so the second condition holds. Finally,
\[
    \frac{
        g_T
    }{
        \sqrt{
            T\log\log T
        }
    }
    =
    O\left(
        \frac{
            \log T
        }{
            \sqrt{
                T\log\log T
            }
        }
    \right)
    \longrightarrow
    0.
\]
Hence, the third condition also holds.

This class includes linear kernels and fixed-degree polynomial kernels on
bounded finite-dimensional domains.

\subsection{Kernels with exponential eigenvalue decay}

The eigenvalues of this type of kernel satisfy $\lambda_j
    \le
    C_1
    \exp\left(
        -C_2j^{\beta_e}
    \right)$
for some constants $C_1,C_2>0$ and $\beta_e>0$. Under the regularity
conditions of \citet{vakili2021information}, the corresponding
information-gain bound can be written as
\begin{equation}
\label{eq:appendix_exponential_G}
    G_k(u)=
    C_k\log(1+u)
    [1+\log(1+u)]^{1/\beta_e}.
\end{equation}

The function in \eqref{eq:appendix_exponential_G} is nondecreasing. Also,
for sufficiently large $u$,
\[
\begin{aligned}
    \frac{
        G_k(2u)
    }{
        G_k(u)
    }
    =
    \left[
        \frac{
            \log(1+2u)
        }{
            \log(1+u)
        }
    \right]^{1+1/\beta_e}
    \longrightarrow
    1.
\end{aligned}
\]
Thus, the first condition of Theorem~\ref{thm:general_kernel_replication} holds.

For fixed $v_0$,
\[
    g_T = G_k\left(\frac{T}{v_0}\right)
    =
    \Theta\left(
        (\log T)^{1+1/\beta_e}
    \right).
\]
Since $1+1/\beta_e>1$, $\log T
    =
    O(g_T)$,
which verifies the second condition.

For the third condition,
\[
\begin{aligned}
    \frac{
        g_T
    }{
        \sqrt{
            T\log\log T
        }
    }
    &=
    O\left(
        \frac{
            (\log T)^{1+1/\beta_e}
        }{
            \sqrt{
                T\log\log T
            }
        }
    \right)
    \longrightarrow
    0.
\end{aligned}
\]
Hence, this class of kernel satisfies all three conditions of
Theorem~\ref{thm:general_kernel_replication}.

The SE kernel on a compact subset of $\mathbb R^d$ is an
important example. Its eigenvalue decay leads to
\[
    G_{\mathrm{SE}}(u)
    =
    O\left(
        [\log(1+u)]^{d+1}
    \right).
\]

\subsection{Kernels with polynomial eigenvalue decay}

The eigenvalues of this type of kernel satisfy
$\lambda_j
    \le
    C_1j^{-\beta_p}$
for some $\beta_p>1$. Under the regularity conditions in
\citet{vakili2021information}, 
\begin{equation}
\label{eq:appendix_polynomial_G}
    G_k(u)
    =
    C_k
    u^{1/\beta_p}
    \left[
        \log(1+u)
    \right]^{1-1/\beta_p}.
\end{equation}

The function in \eqref{eq:appendix_polynomial_G} is nondecreasing. Moreover,
\[
\begin{aligned}
    \frac{
        G_k(2u)
    }{
        G_k(u)
    }
    =
    2^{1/\beta_p}
    \left[
        \frac{
            \log(1+2u)
        }{
            \log(1+u)
        }
    \right]^{1-1/\beta_p}
    \longrightarrow
    2^{1/\beta_p}.
\end{aligned}
\]
Thus, the first condition holds.

For fixed $v_0$,
\[
    g_T
    =
    \Theta\left(
        T^{1/\beta_p}
        (\log T)^{1-1/\beta_p}
    \right).
\]
Since $\beta_p>1$, the polynomial term dominates $\log T$, and hence $\log T
    =
    O(g_T)$.
Therefore, the second condition holds.

For the third condition,
\[
\begin{aligned}
    \frac{
        g_T
    }{
        \sqrt{
            T\log\log T
        }
    }
    &=
    \Theta\left(
        T^{1/\beta_p-1/2}
        \frac{
            (\log T)^{1-1/\beta_p}
        }{
            \sqrt{\log\log T}
        }
    \right).
\end{aligned}
\]
This quantity converges to zero whenever $\frac{1}{\beta_p}
    <
    \frac12$,
or equivalently,
$\beta_p>2$.
Thus, kernels with polynomial eigenvalue decay satisfying
$\beta_p>2$ meet all three conditions of
Theorem~\ref{thm:general_kernel_replication}.

A Mat\'ern kernel with smoothness parameter $\nu$ in dimension $d$ has
polynomial eigenvalue decay with $\beta_p
    =
    \frac{2\nu+d}{d}$.
Therefore, $\beta_p>2$ is equivalent to $\nu>\frac d2$.
In this case,
\[
    G_{\mathrm{Mat}}(u)
    =
    O\left(
        u^{\frac{d}{2\nu+d}}
        [\log(1+u)]^{\frac{2\nu}{2\nu+d}}
    \right),
\]
and all three conditions of
Theorem~\ref{thm:general_kernel_replication} are satisfied.

\section{Derivation of the Rate after Choosing the Replication Number and Kernels}
\label{app:kernel_specific_rates}

Under the conditions of Theorem~\ref{thm:general_kernel_replication}, choose $b_T
    =
    \Theta(\log T)$. Then $N_T
    =
    \Theta\left(
        \frac{T}{\log T}
    \right)$
and, by Lemma \ref{lem:plugin_correction_order}, 
$\kappa_T=O(1)$ and $h_T=O(1)$.
Since $M$ is fixed and finite, $W_T=\Theta(1)$, and $\log(1+b_T/W_T)=\Theta(\log\log T)$.
The proof of
Theorem~\ref{thm:general_kernel_replication} shows that $\overline\gamma_T^{\,k}(b_T)
    =
    \Theta(g_T)$.
Since $\log T
    =
    O(g_T)$,
the definition of the confidence coefficient gives
\[
    \left(
        \overline\Lambda_T^{\,k}(b_T)
    \right)^2
    +
    \log\left(
        \frac{1+T/v_0}{k_{\min}}
    \right)
    =
    O(g_T).
\]
Substitution into Corollary~\ref{thm:observed_simple_regret} yields
\[
\begin{aligned}
    f(\hat x_T)-f^\star
    =
    O\left(
        \sqrt{
            \frac{
                g_T
                \log T
                g_T
            }{
                T\log\log T
            }
        }
    \right)
    =
    O\left(
        g_T
        \sqrt{
            \frac{
                \log T
            }{
                T\log\log T
            }
        }
    \right).
\end{aligned}
\]

For a SE kernel,
\[
    g_T
    =
    O\left(
        (\log T)^{d+1}
    \right).
\]
Therefore,
\[
\begin{aligned}
    f(\hat x_T)-f^\star
    =
    O\left(
        (\log T)^{d+1}
        \sqrt{
            \frac{
                \log T
            }{
                T\log\log T
            }
        }
    \right)
    =
    O\left(
        \frac{
            (\log T)^{d+3/2}
        }{
            \sqrt{
                T\log\log T
            }
        }
    \right).
\end{aligned}
\]

For a Mat\'ern kernel with smoothness parameter $\nu$,
\[
    g_T
    =
    O\left(
        T^{\frac{d}{2\nu+d}}
        (\log T)^{\frac{2\nu}{2\nu+d}}
    \right).
\]
Consequently,
\[
\begin{aligned}
    f(\hat x_T)-f^\star
    =
    \widetilde O\left(
        T^{\frac{d}{2\nu+d}}
        T^{-1/2}
    \right)
    =
    \widetilde O\left(
        T^{
            -\frac{
                2\nu-d
            }{
                2(2\nu+d)
            }
        }
    \right).
\end{aligned}
\]
This rate decreases polynomially when $\nu>\frac d2$.

\section{Experiment Settings and Additional Results}
\subsection{Objective Functions and Experimental Settings}
\label{app:rkhs-protocol}
This subsection provides the complete construction and implementation details for the controlled RKHS experiments in Section 7.1. 

For each dimension $d\in\{2,4\}$, the decision space is
$\mathcal X_d=[0,1]^d$. We consider a SE kernel and a
Mat\'ern-$5/2$ kernel, both with isotropic lengthscale $\ell=0.2$.
For each combination of dimension and kernel, we construct one fixed
objective as the finite kernel expansion $f_{d,k}(x)
    =
    \sum_{j=1}^{100}w_{d,j}k(x,z_{d,j})$.
Consequently, $f_{d,k}$ belongs exactly to the RKHS induced by $k$.
Within each dimension, the two kernels use the same 100 basis points
and standard-normal weights. Separate basis points and weights are
generated for the two- and four-dimensional experiments. Each
objective is fixed and used in every macro-replication.
The true extrema used to compute the simple regret are obtained
from a $262{,}144$-point Sobol screen followed by multistart local
optimization. Tables~\ref{tab:app-functions-2d}
and~\ref{tab:app-functions-4d} report the resulting minimizers,
objective ranges, and RKHS norms (Here $R_f=f_{\max}-f^\star$ and
    $B_f=\lVert f\rVert_{\mathcal H_k}$).
\begin{table}[htbp]
    \centering
    \small
    \caption{Properties of the two-dimensional RKHS objectives}
    \label{tab:app-functions-2d}
    \begin{tabular}{@{}lrrrrrr@{}}
        \toprule
        Kernel
        & $x^\star_1$
        & $x^\star_2$
        & $f^\star$
        & $f_{\max}$
        & $R_f$
        & $B_f$ \\
        \midrule
        Mat\'ern-$5/2$
        & 0.058755
        & 0.952746
        & -2.945546
        & 3.072480
        & 6.018026
        & 6.931367 \\
        Squared exponential
        & 0.609984
        & 0.537395
        & -3.081055
        & 3.369256
        & 6.450311
        & 6.935835 \\
        \bottomrule
    \end{tabular}
\end{table}
\begin{table}[htbp]
    \centering
    \small
    \caption{Properties of the fixed four-dimensional RKHS objectives.}
    \label{tab:app-functions-4d}
    \begin{tabular}{@{}lp{58mm}rrrr@{}}
        \toprule
        Kernel
        & $x^\star$
        & $f^\star$
        & $f_{\max}$
        & $R_f$
        & $B_f$ \\
        \midrule
        Mat\'ern-$5/2$
        & $(0.680540,0.617090,0.067218,0.626605)$
        & -5.390637
        & 3.219524
        & 8.610161
        & 11.255655 \\
        Squared exponential
        & $(0.678059,0.621087,0.066322,0.619068)$
        & -6.000295
        & 3.217247
        & 9.217542
        & 11.318412 \\
        \bottomrule
    \end{tabular}
\end{table}

The two-dimensional experiments begin with 20 Latin-hypercube points. Following the work of \cite{jalali2017comparison}, we discretize the space to optimize the EI function. In two dimensional case, the discretized set includes 2,048 scrambled Sobol candidates at each iteration. The four-dimensional experiments begin with 40 Latin-hypercube points and 8,192 discrete scrambled Sobol candidates. In four dimensions, to improve the quality of EI maximization, we add a follow-up gradient search. Specifically, in each iteration, we first select the ten candidates with the largest acquisition values and then adopt L-BFGS-B strting from these ten points.
For the replication-order ablation, every four-dimensional arm uses 20 initial points rather than 40. This is to keep the $O(T^{1/2})$ schedule feasible at the smallest ablation budget.


In the standard CertEI configuration, the replication size is $b_T
    =
    \rho\log T$,
rounded upward to the nearest multiple of five. We use $\rho=8$ for
the Mat\'ern-$5/2$ kernel and $\rho=6$ for the
SE kernel. In addition, the theoretical confidence coefficient is defined through the exact maximum information gain, which is not directly computable. In the numerical experiments, we therefore replace it with a tractable kernel-specific proxy combining its known asymptotic order from \cite{vakili2021information} with the finite-batch bound in Lemma \ref{lem:incumbent_coefficient_optimality}. 

\subsection{Additional EI-incumbent ablation}
\label{app:additional-incumbent}

We first report the three EI-incumbent ablations in the remaining three scenarios.
\begin{figure}[!htb]
    \centering
    \captionsetup[subfigure]{font=scriptsize}
    \begin{subfigure}[t]{0.48\textwidth}
        \centering
        \includegraphics[width=\linewidth]
        {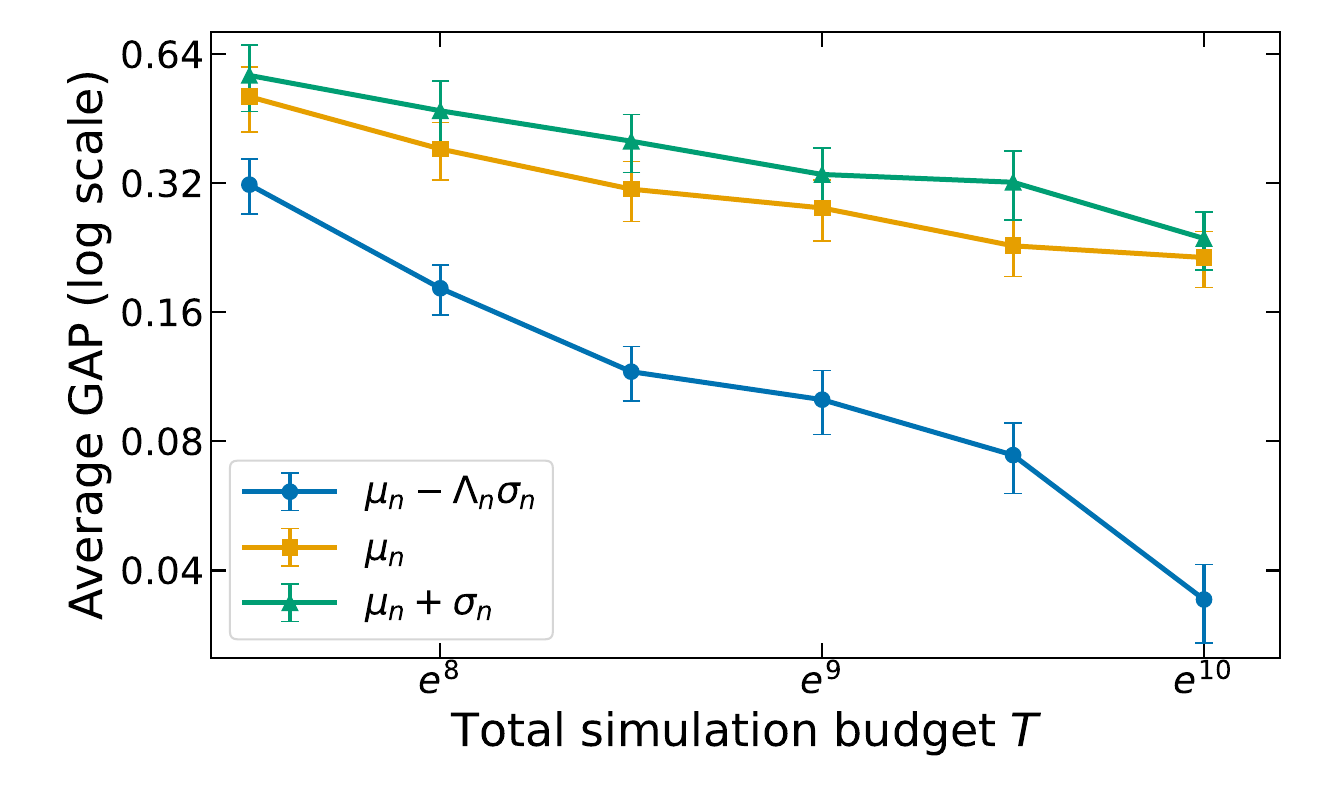}
        \caption{Two-dimensional Mat\'ern-$5/2$.}
        \label{fig:additional-incumbent-matern-2d}
    \end{subfigure}
    \hfill
    \begin{subfigure}[t]{0.48\textwidth}
        \centering
        \includegraphics[width=\linewidth]
        {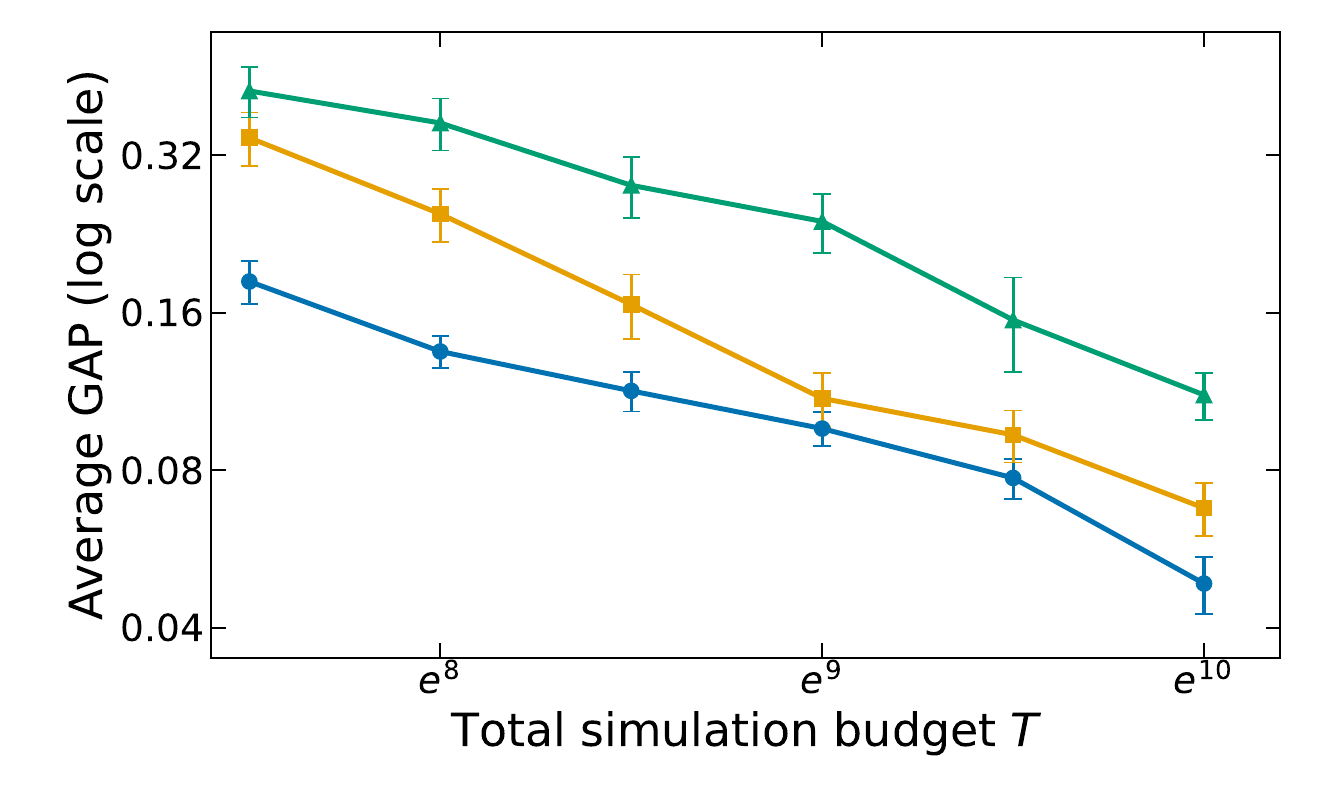}
        \caption{Two-dimensional SE.}
        \label{fig:additional-incumbent-se-2d}
    \end{subfigure}

    \medskip

    \begin{subfigure}[t]{0.48\textwidth}
        \centering
        \includegraphics[width=\linewidth]
        {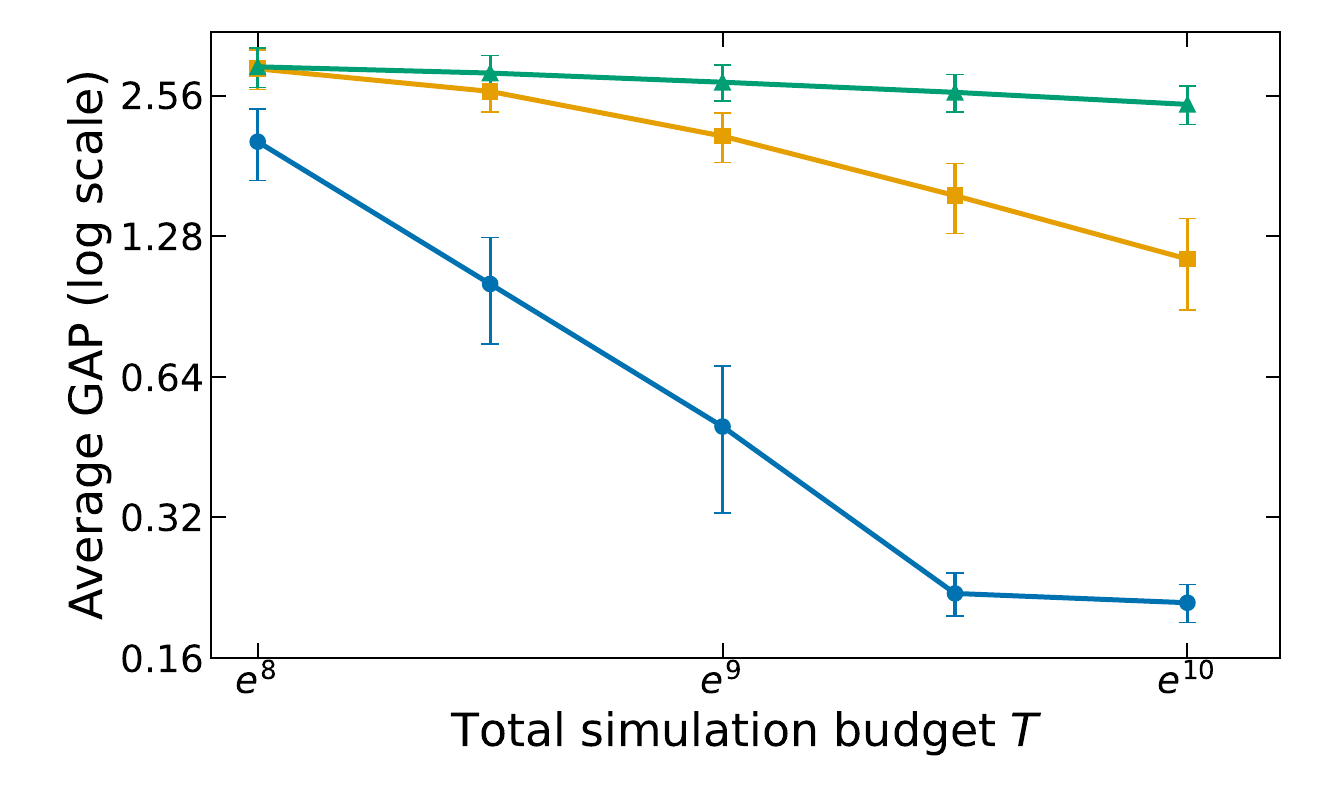}
        \caption{Four-dimensional SE.}
        \label{fig:additional-incumbent-se-4d}
    \end{subfigure}

    \caption{Additional EI-incumbent ablations}
    \label{fig:additional-incumbent}
\end{figure}

Consistent with the four-dimensional Mat\'ern-$5/2$ result in the main text, the CertEI incumbent attains the smallest average GAP at every reported budget in all three additional settings. As the budget increases, the other two incumbent choices converge more slowly, where the upper-quantile incumbent is generally the worst. 

\subsection{Additional Replication-order ablation}
\label{app:additional-replication} 
We report the remaining replication-order experiments.
\begin{figure}[!htb]
    \centering
    \captionsetup[subfigure]{font=scriptsize}
    \begin{subfigure}[t]{0.48\textwidth}
        \centering
        \includegraphics[width=\linewidth]
        {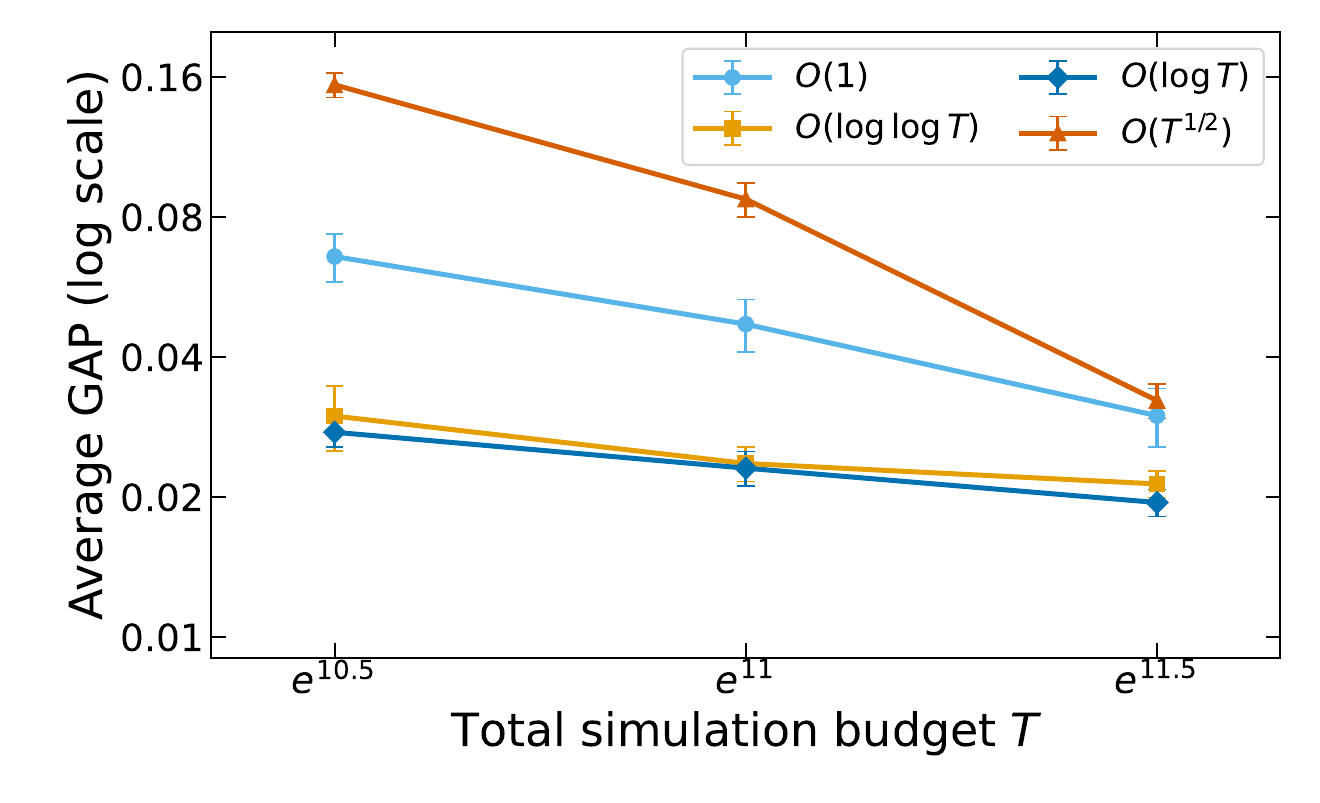}
        \caption{Two-dimensional Mat\'ern-$5/2$.}
        \label{fig:additional-replication-matern-2d}
    \end{subfigure}
    \hfill
    \begin{subfigure}[t]{0.48\textwidth}
        \centering
        \includegraphics[width=\linewidth]
        {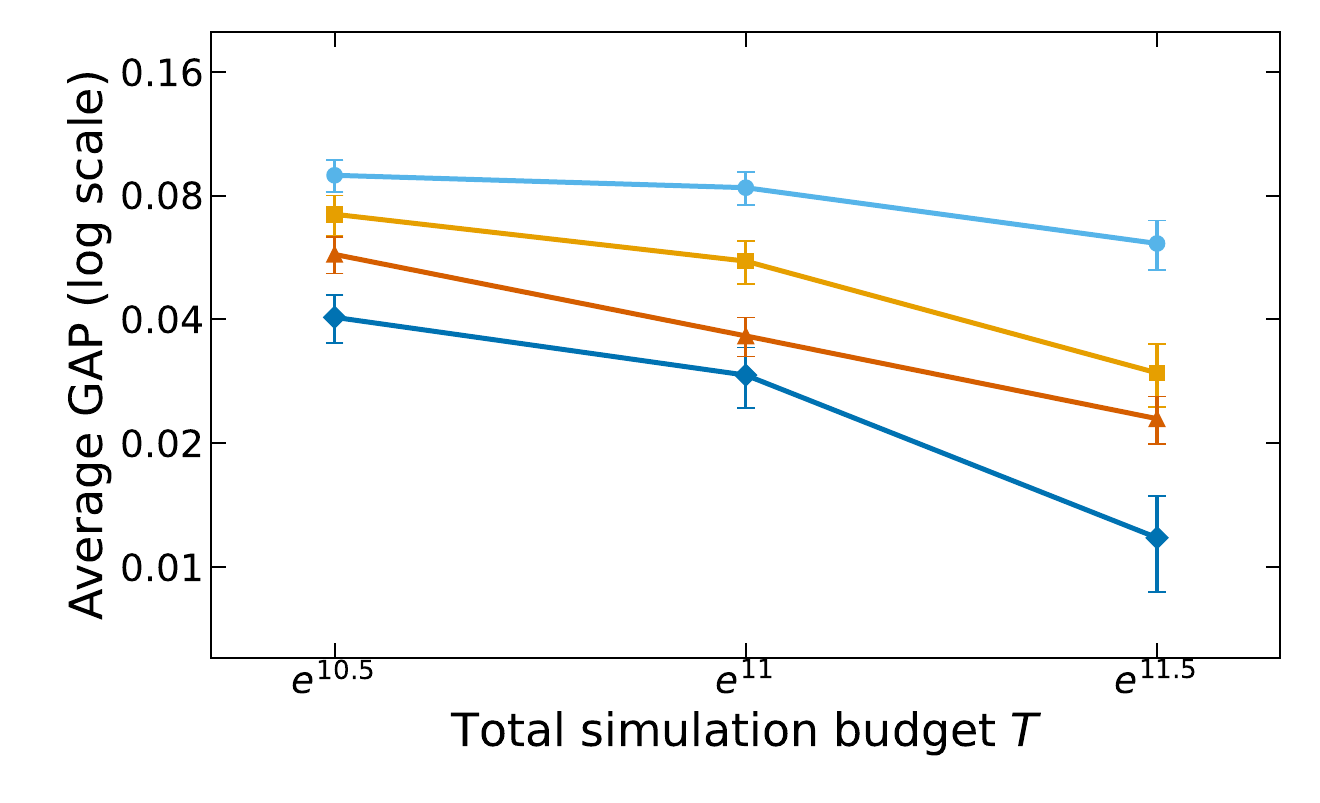}
        \caption{Two-dimensional SE.}
        \label{fig:additional-replication-se-2d}
    \end{subfigure}

    \medskip

    \begin{subfigure}[t]{0.55\textwidth}
        \centering
        \includegraphics[width=\linewidth]
        {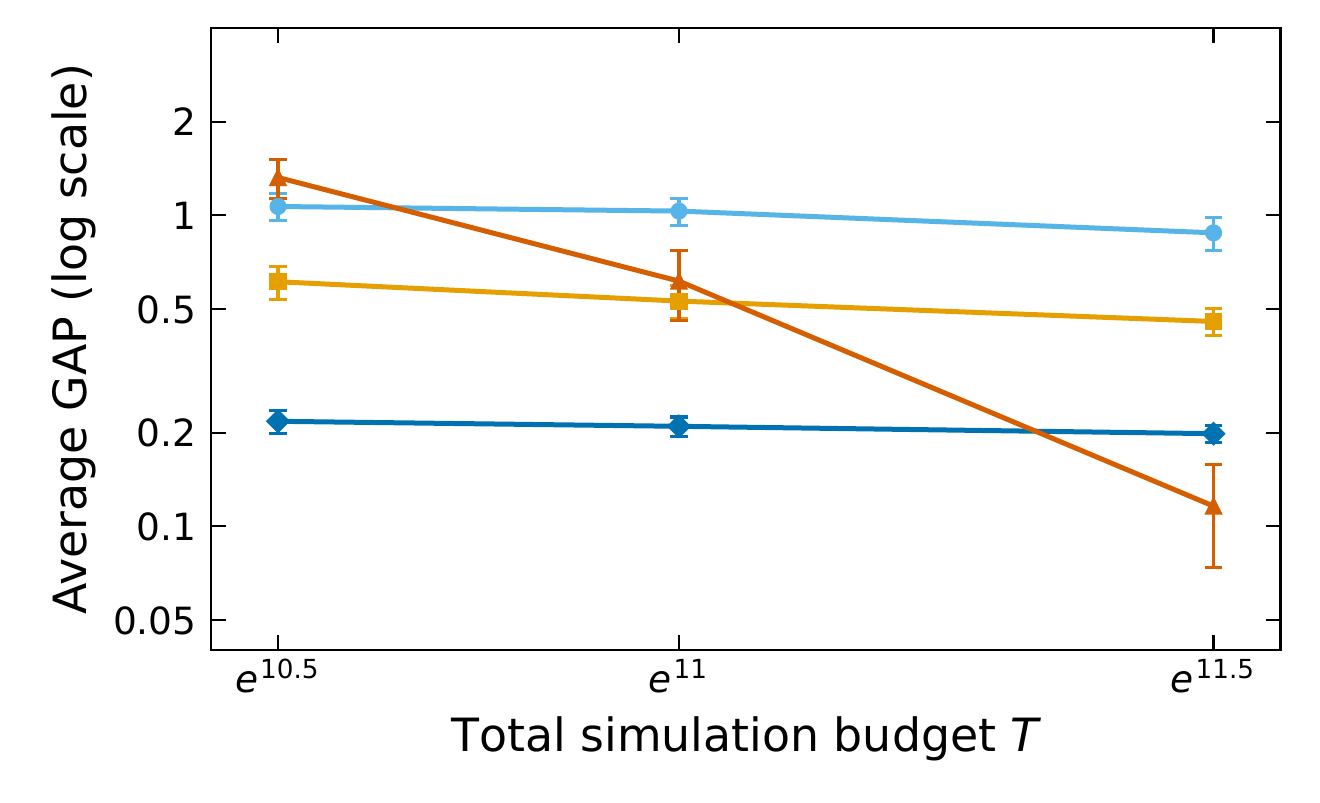}
        \caption{Four-dimensional SE.}
        \label{fig:additional-replication-se-4d}
    \end{subfigure}

    \caption{Average GAP under the four replication orders for the
    additional objective functions}
    \label{fig:additional-replication}
\end{figure}

For both two-dimensional objectives, $O(\log T)$ attains the lowest
average GAP at all three budgets. Its advantage over
$O(\log\log T)$ is relatively small for the Mat\'ern-$5/2$ objective
but more pronounced for the SE objective. By
contrast, $O(T^{1/2})$ performs poorly on the Mat\'ern-$5/2$ objective
at the smallest budget because its large replication number leave only a few
adaptive decisions after initialization. 
For the four-dimensional SE objective,
$O(\log T)$ has the lowest average GAP at $T=e^{10.5}$ and
$T=e^{11}$, whereas $O(T^{1/2})$ performs best at $T=e^{11.5}$.
Thus, a larger replication order can become competitive when the
available budget is sufficiently large, but it is less effective at
the two smaller budgets. Together with the four-dimensional
Mat\'ern-$5/2$ result in the main text, these experiments support
$O(\log T)$ as a stable compromise between the number of searched
locations and the precision obtained at each location.

\subsection{Additional Final-recommendation ablation}
\label{app:additional-final}

We further examine the final-recommendation rule on the three additional instances.
\begin{figure}[!htb]
    \centering
    \captionsetup[subfigure]{font=scriptsize}
    \begin{subfigure}[t]{0.48\textwidth}
        \centering
        \includegraphics[width=\linewidth]
        {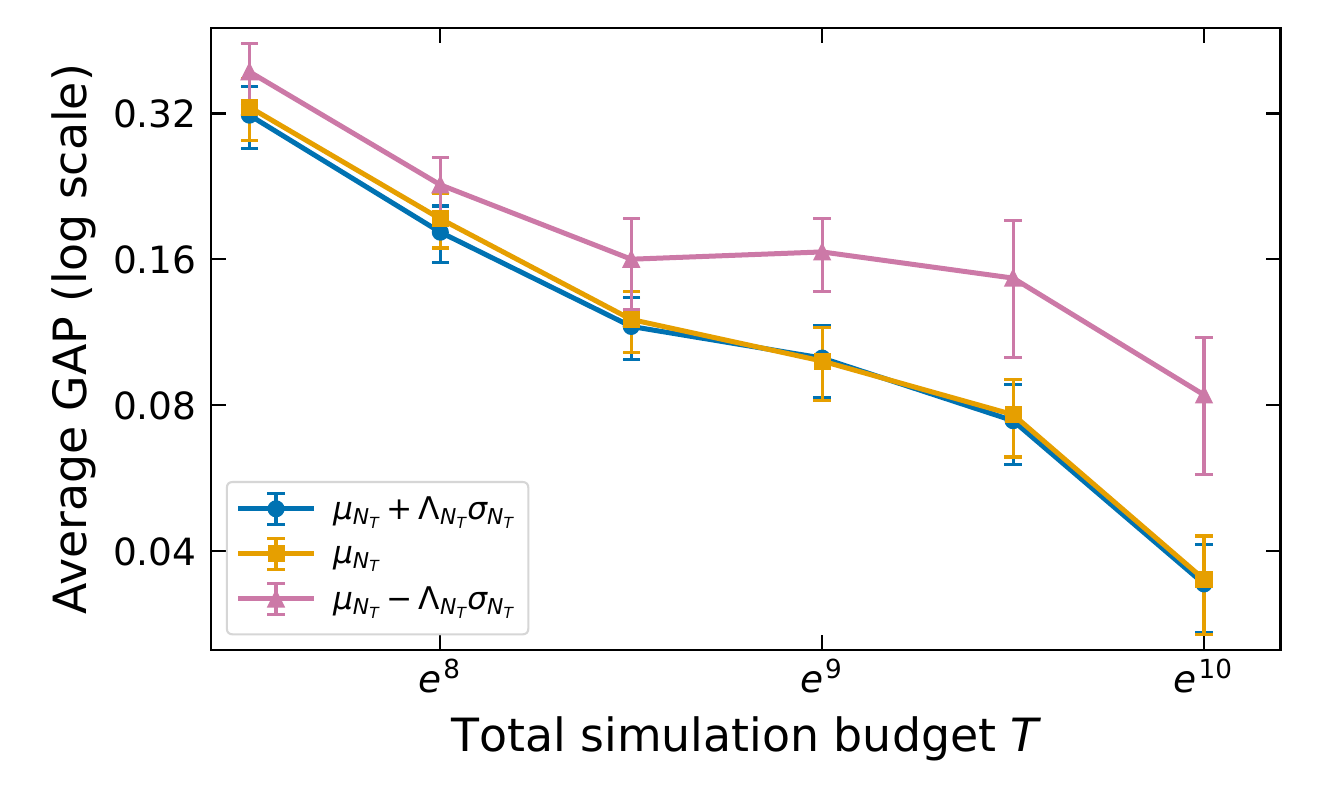}
        \caption{Two-dimensional Mat\'ern-$5/2$.}
        \label{fig:additional-final-matern-2d}
    \end{subfigure}
    \hfill
    \begin{subfigure}[t]{0.48\textwidth}
        \centering
        \includegraphics[width=\linewidth]
        {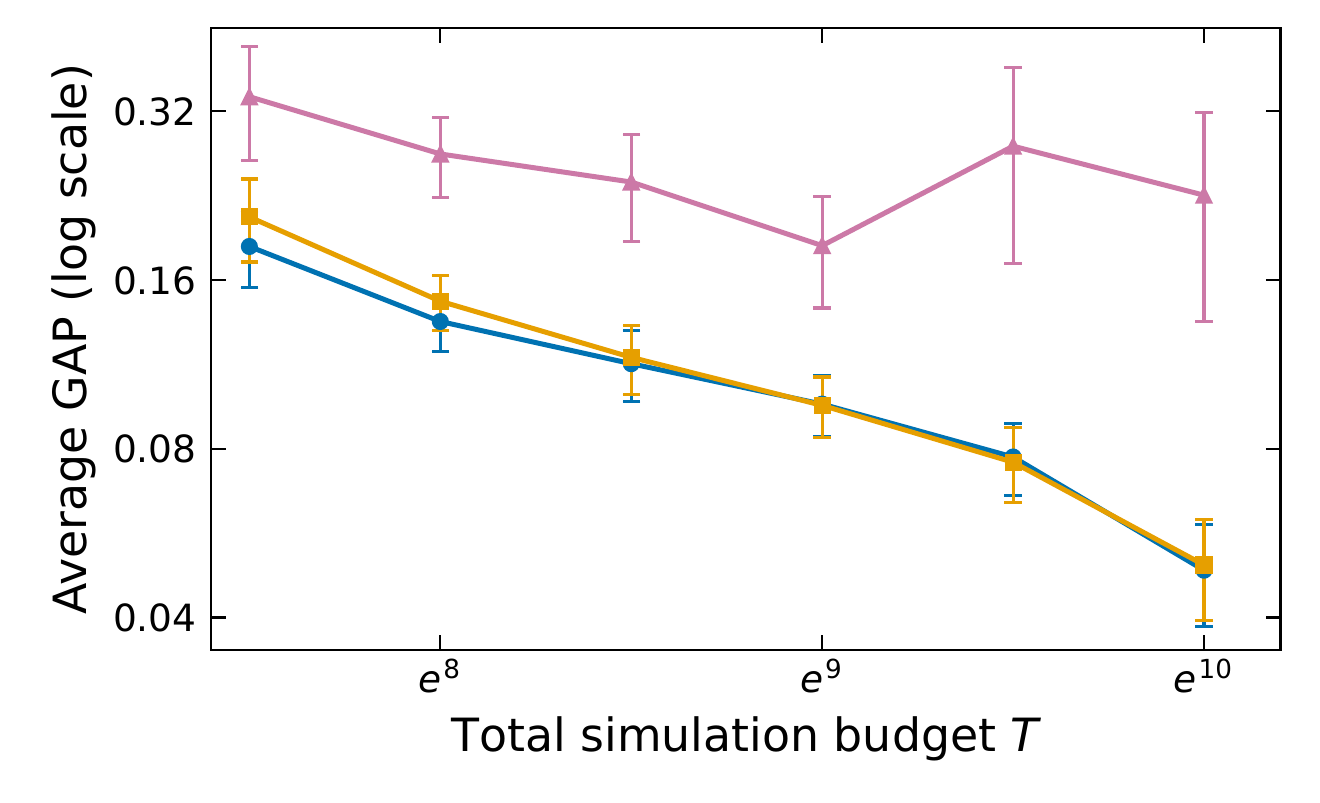}
        \caption{Two-dimensional SE.}
        \label{fig:additional-final-se-2d}
    \end{subfigure}

    \medskip

    \begin{subfigure}[t]{0.48\textwidth}
        \centering
        \includegraphics[width=\linewidth]
        {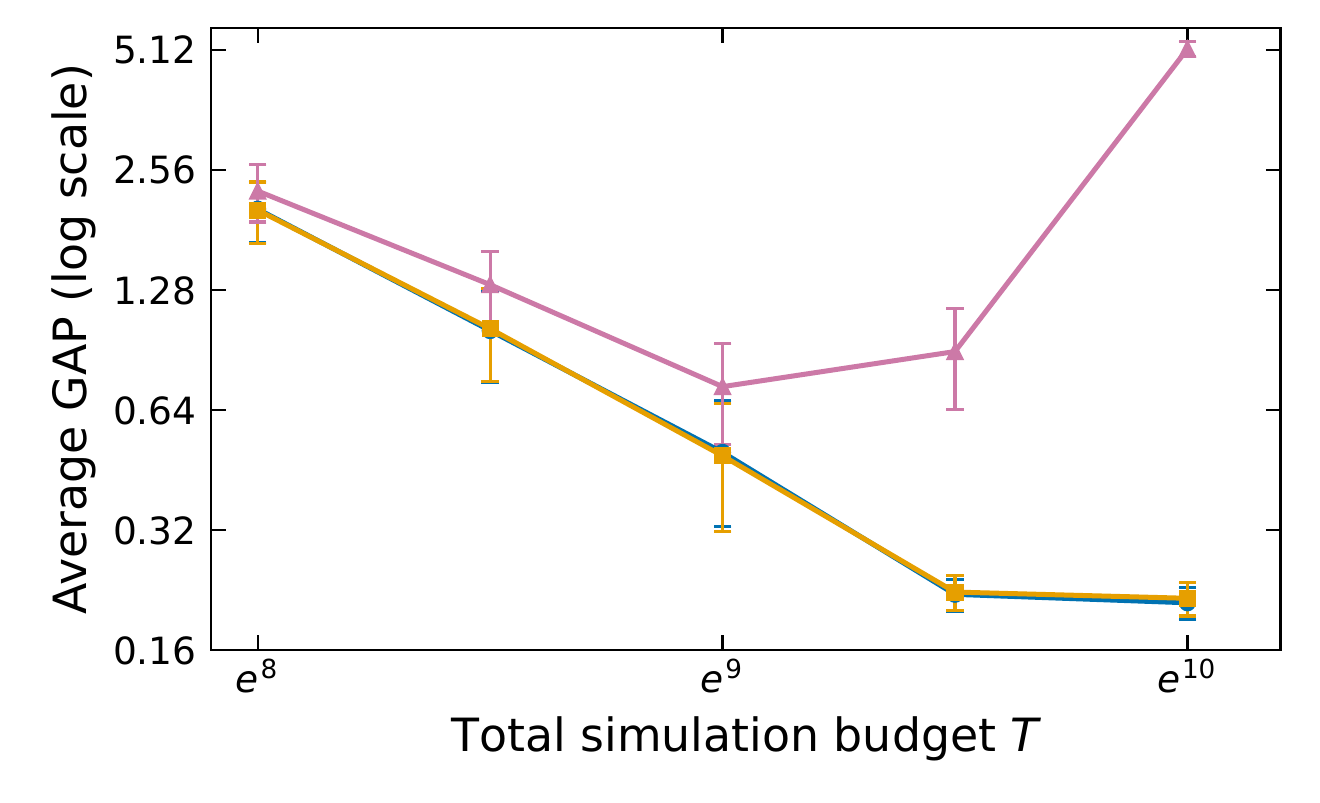}
        \caption{Four-dimensional SE.}
        \label{fig:additional-final-se-4d}
    \end{subfigure}

    \caption{Average GAP under the three final-recommendation rules}
    \label{fig:additional-final}
\end{figure}

Figure~\ref{fig:additional-final} shows that the calibrated upper quantile and posterior-mean rules perform comparably on these instances, like the one reported in the main text. In contrast, the optimistic lower quantile rule has a larger average GAP throughout, with the separation becoming especially pronounced for the four-dimensional SE instance at large budgets. In this case, we see that the GAP of the lower-quantile recommendation become worse when $T$ increases. As $T$ increases, $\Lambda_{N_T}$ also increases, strengthening the uncertainty reward in the optimistic lower-quantile score $\mu_{N_T}(x)-\Lambda_{N_T}\sigma_{N_T}(x)$. In the four-dimensional SE setting, substantial residual uncertainty remains, so the larger coefficient makes this rule increasingly likely to select a highly uncertain but truly poor sampled point, causing the average GAP to rise.


\subsubsection{A diagnostic terminal-selection test}
\label{app:final-stress}

We report an additional experiment designed to distinguish our final recommendation rule from returning the sampled point with the smallest posterior mean. When simulation noise is high, more posterior uncertainty remains at termination under the same budget, making the effect of the uncertainty adjustment easier to observe. This experiment illustrates such a case. 

The minimization domain is $\mathcal X=[0,1]^4$. The objective is a finite
kernel expansion $f(x)=\sum_{j=1}^{14}w_j k_{5/2}(x,c_j)$, where $k_{5/2}$ is the isotropic Mat\'ern-$5/2$ kernel with lengthscale $\ell=0.2$. 
We divide the 14 kernel centers $c_j$ into three groups: safe, global, and decoy. The three groups create regions with different combinations of objective quality and optimization difficulty. The safe basin is broad and supported by several nearby centers, making it relatively easy to locate and estimate, but it is deliberately suboptimal. The global basin is deeper but more localized, while the decoy basins provide isolated suboptimal regions that can appear attractive under noisy posterior estimates. This construction creates a nontrivial terminal-selection problem: the algorithm must distinguish a reliably good sampled point from points whose posterior means appear favorable but remain highly uncertain. 
Let $\mathbf 1_4=(1,1,1,1)^\top$, let
$c_{\mathrm{s}}=0.30\mathbf 1_4$, and let $e_i$ denote the $i$th coordinate
vector. The relative coefficient magnitudes of the safe, global, and decoy components are $1:14:6$.
Table~\ref{tab:final-stress-function} reports the normalized coefficients
used in the experiment.
\begin{table}[htbp]
\centering
\caption{Kernel centers and normalized coefficients of the diagnostic
objective.}
\label{tab:final-stress-function}
\begin{tabular}{@{}p{0.17\linewidth}p{0.59\linewidth}r@{}}
\toprule
Component & Kernel centers & Weight per center \\
\midrule
Safe basin
&
$c_{\mathrm{s}}$ and the eight axis offsets
$c_{\mathrm{s}}\pm0.11e_i$, $i=1,\ldots,4$
&
$-0.06986611$
\\[1mm]

Global basin
&
$(0.72,0.72,0.72,0.72)$
&
$-0.97812556$
\\[1mm]

Decoy basins
&
$(0.76,0.24,0.76,0.24)$,
$(0.24,0.76,0.24,0.76)$,
$(0.78,0.78,0.22,0.22)$, and
$(0.22,0.22,0.78,0.78)$
&
$-0.41919667$
\\
\bottomrule
\end{tabular}
\end{table}
The objective function is normalized to have response range $R_f=f_{\max}-f^\star=1$, and its RKHS norm is $\lVert f\rVert_{\mathcal H}=1.42312669$.
Its global minimizer is $x^\star
=
(0.718338,0.718338,0.718338,0.718338)$.
We add independent homoscedastic Gaussian noise on $f$: $\epsilon\sim\mathcal N\bigl(0,(3R_f)^2\bigr)$. 
Thus, the noise standard deviation is $\sigma_\epsilon=3R_f$. It is worth noting that homoscedasticity is imposed only in the data-generating process. The algorithm is not informed that the noise variance is constant and continues to estimate it at each sampled point using the replication-based procedure. The results are reported in the following figure.

\begin{figure}[!htb]
    \centering
        \includegraphics[width=\linewidth]
        {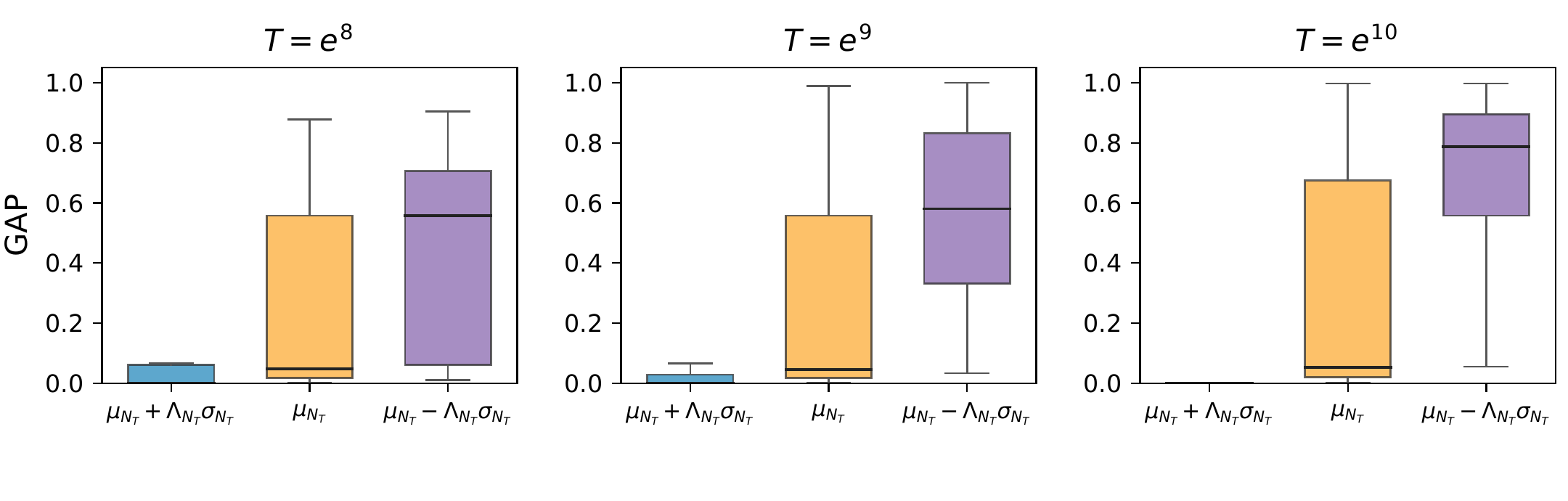}
    \caption{GAP distributions at three representative budgets}
    \label{fig:final-stress}
\end{figure}
Figure~\ref{fig:final-stress} shows a clear separation among the three
rules. The conservative recommendation attains a substantially smaller
GAP than posterior-mean selection throughout the tested budget range,
while the optimistic recommendation performs worst. Under uniformly
high noise, a favorable posterior mean can still be associated with
substantial residual uncertainty; penalizing that uncertainty therefore
helps prevent an insufficiently supported design from being returned.

\subsection{Joint Ablation of the Three Design Choices}
\label{app:joint-ablation}

\begin{figure}[htbp]
    \centering
    \includegraphics[width=\textwidth]
    {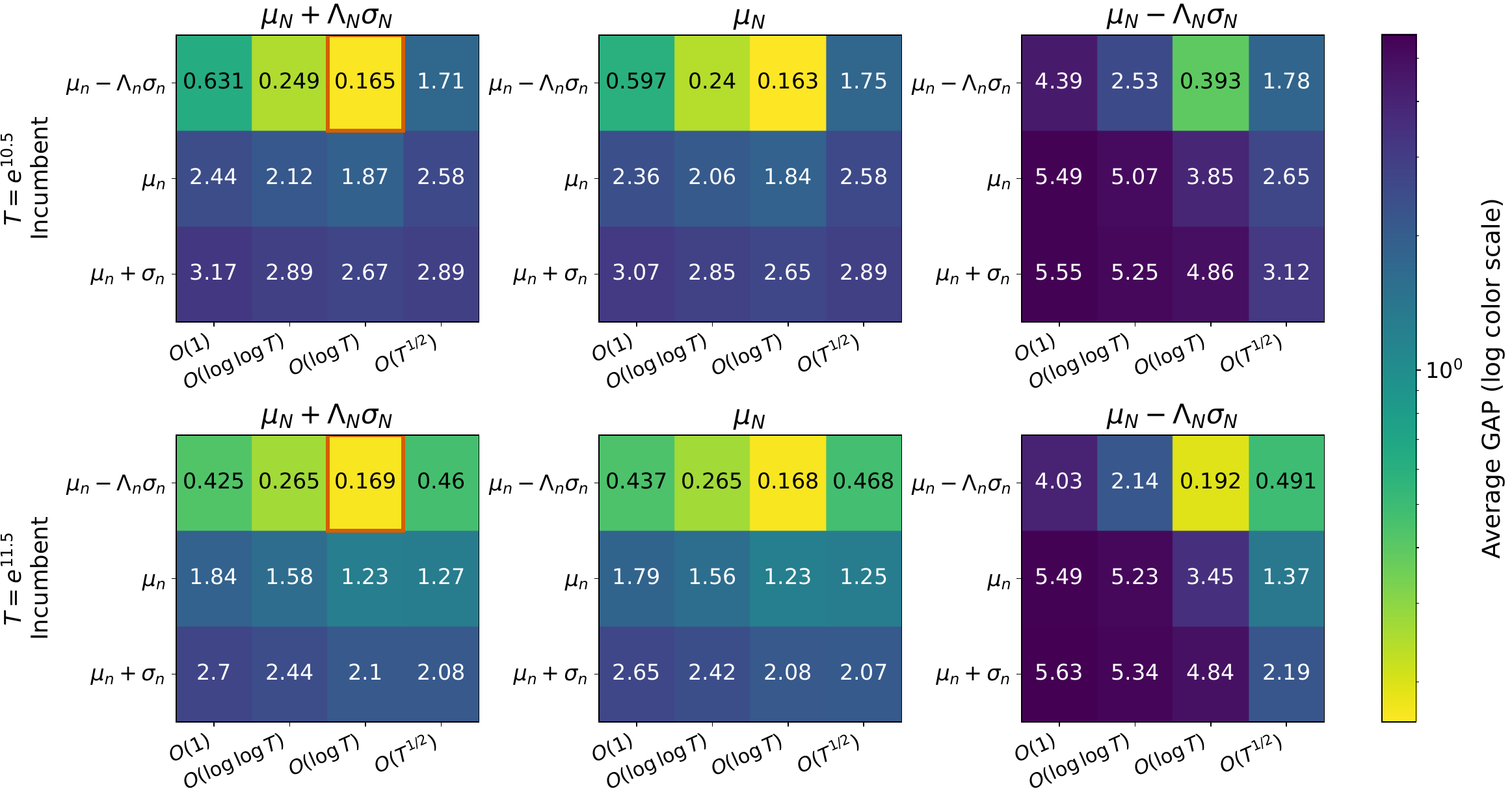}
    \caption{Joint ablation on the four-dimensional Mat\'ern-$5/2$
    objective}
    \label{fig:joint-ablation-matern-4d}
\end{figure}

We present the crossed experiment that varies all three design choices simultaneously. This gives $3\times4\times3=36$ configurations for each function--budget pair. We conduct the crossed experiment on the two-dimensional Mat\'ern-$5/2$, two-dimensional SE, and four-dimensional Mat\'ern-$5/2$ objectives at $T=e^{10.5}$ and $T=e^{11.5}$. The results are shown in the following Figures \ref{fig:joint-ablation-matern-2d}, \ref{fig:joint-ablation-se-2d}, and \ref{fig:joint-ablation-matern-4d}. In these figures, panel rows correspond to the two budgets and panel columns to the three terminal rules. Within each panel, rows correspond to EI incumbents and columns to replication orders. Each displayed cell is the average GAP over 100 macro-replications.

\begin{figure}[htbp]
    \centering
    \includegraphics[width=\textwidth]
    {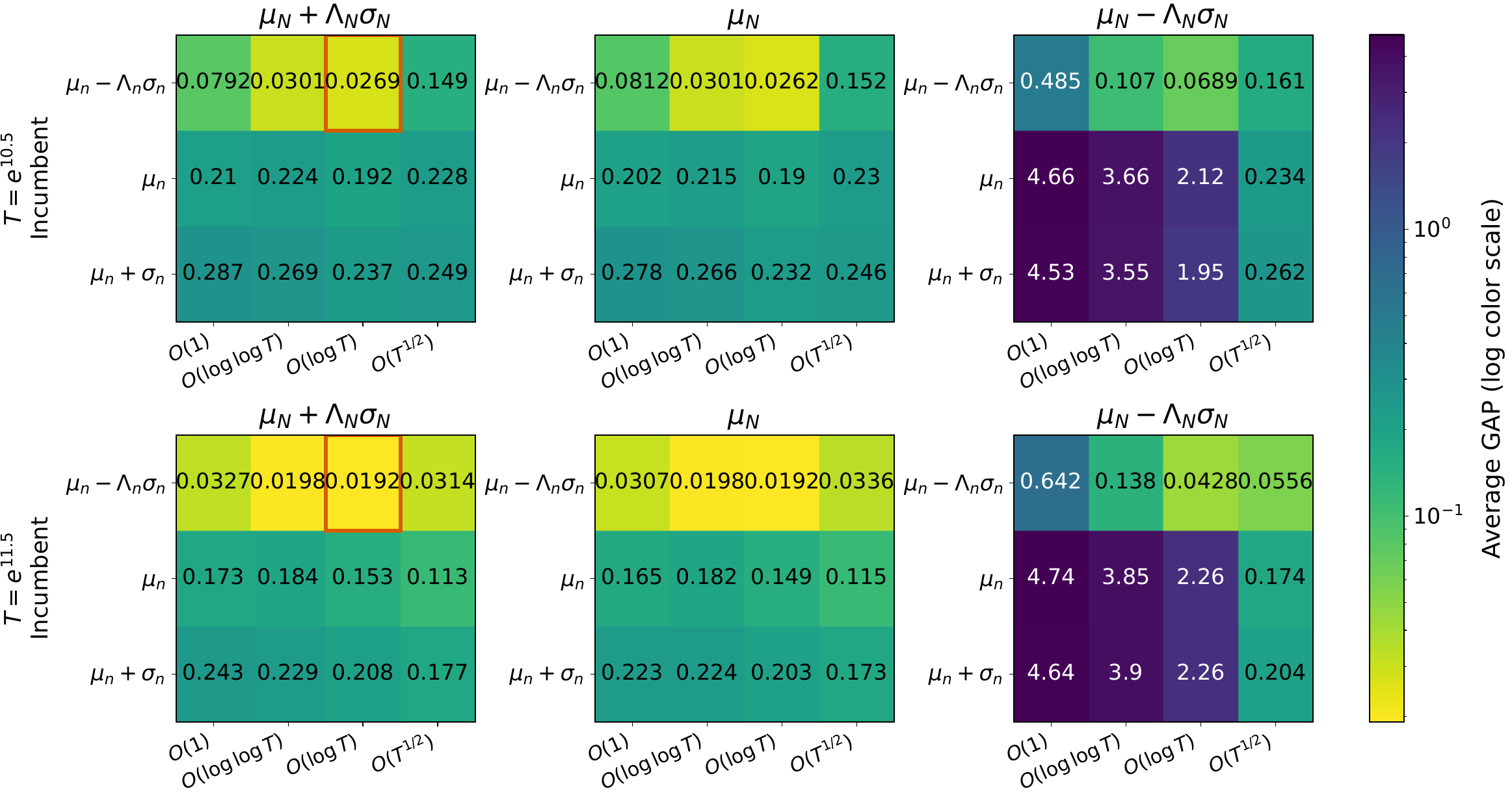}
    \caption{Joint ablation on the two-dimensional Mat\'ern-$5/2$
    objective}
    \label{fig:joint-ablation-matern-2d}
\end{figure}

\begin{figure}[htbp]
    \centering
    \includegraphics[width=\textwidth]
    {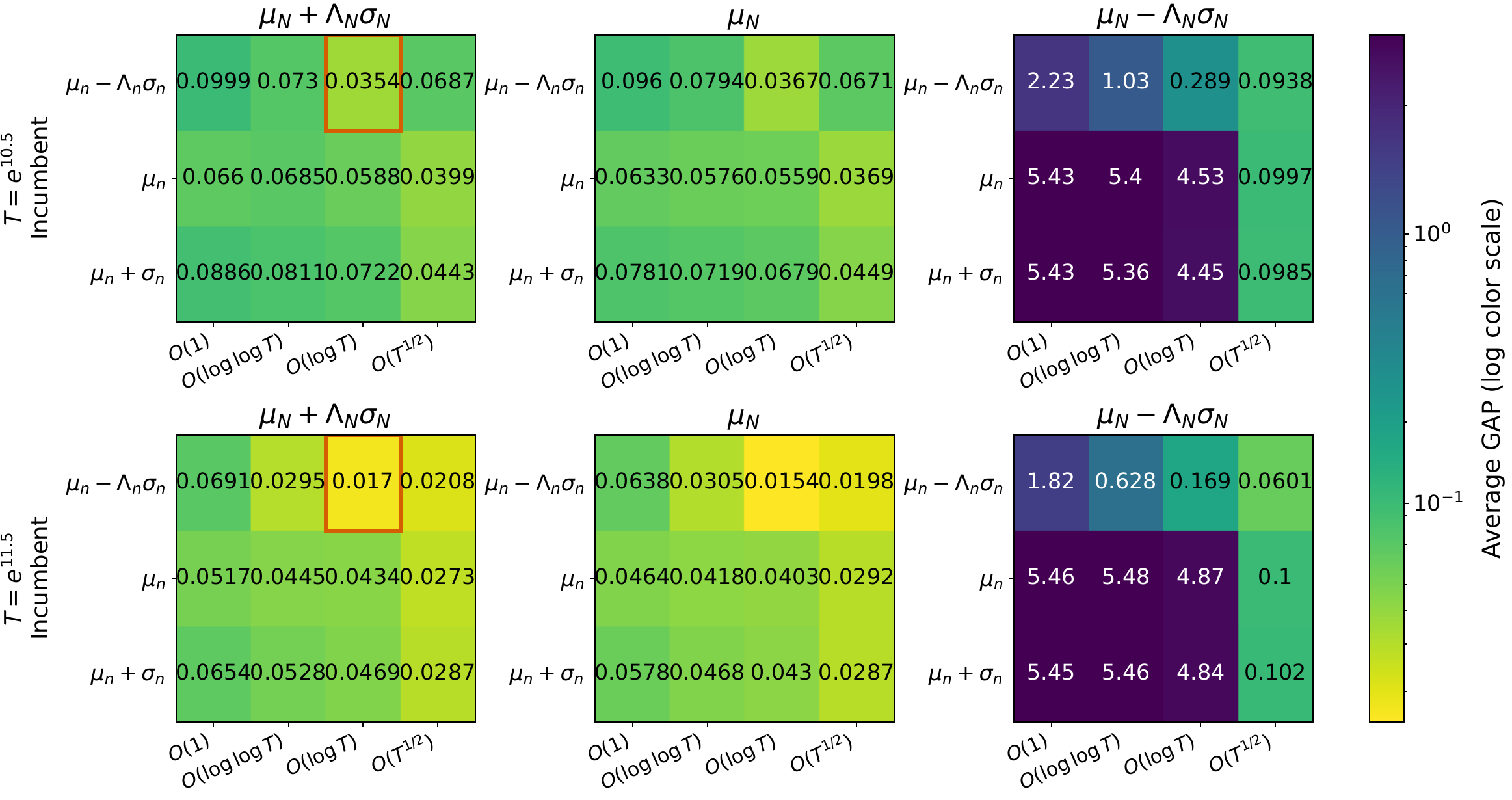}
    \caption{Joint ablation on the two-dimensional
    SE objective}
    \label{fig:joint-ablation-se-2d}
\end{figure}

We observe that the joint ablation yields conclusions consistent with the preceding one-factor-at-a-time experiments. In every tested function--budget setting, the best-performing configurations use $O(\log T)$ replication together with the proposed lower-quantile incumbent. For the final recommendation, the optimistic lower-quantile rule performs substantially worse than the other two alternatives, whereas the posterior-mean rule and our proposed conservative upper-quantile rule yield similar results. More importantly, the complete configuration formed by our three recommended design choices (outlined in red in the figure) consistently ranks among the top two of all 36 configurations in every tested setting. These results show that the three choices also perform robustly when used jointly, rather than being effective only when the other components are held fixed.

\subsection{Settings for the Synthetic Test Functions}
\label{app:synthetic-setting}

We consider the six-hump Camelback, rescaled Branin, Hills, and Hartmann-6 functions. All four algorithms use a Mat\'ern-$5/2$ kernel functions. The GP hyperparameters are initialized by maximum likelihood and re-estimated after every sequential stage using weakly regularized maximum a posteriori estimation.

The standard deviation of the simulation noise is $\tau(x)=R_f\{a+b\,u(x)\}$,
where $a>0$, $b<0$, and $a+b>0$. Therefore, $\tau(x)$ decreases
from $aR_f$ near a global minimizer to $(a+b)R_f$ near the upper end
of the objective range. This deliberately places the largest noise
near the global optimum, producing the same qualitative adverse
noise pattern considered in \cite{jalali2017comparison}, although
the noise magnitudes used here are different. The problem-specific
settings are summarized in Table~\ref{tab:analytical-settings}.

\begin{table}[h]
    \centering
    \small
    \caption{Settings for the analytical test functions. The nominal
    replication size is $\rho\log T$.}
    \label{tab:analytical-settings}
    \begin{tabular}{@{}lcccc@{}}
        \toprule
        Problem & $d$ & $(a,b)$ & $\rho$ & Candidates \\
        \midrule
        Six-hump Camelback & 2 & $(4,-1.75)$  & 8 & 2,048 \\
        Rescaled Branin    & 2 & $(4,-3)$     & 6 & 2,048 \\
        Hills              & 2 & $(8,-1.5)$   & 8 & 2,048 \\
        Hartmann-6         & 6 & $(7.2,-3.5)$ & 8 & 10,000 \\
        \bottomrule
    \end{tabular}
\end{table}

%
%
%

\bibliographystyle{informs2014} 
\bibliography{sample} 


\end{document}